\documentclass{article}

\usepackage{microtype}
\usepackage{graphicx}
\usepackage{subfig}
\usepackage{booktabs}
\usepackage{hyperref}

\usepackage[accepted]{icml2020}

\icmltitlerunning{Landscape Connectivity and Dropout Stability of SGD Solutions for Over-parameterized Neural Networks}

\usepackage{setspace}
\usepackage{cprotect}
\usepackage{amsmath,amssymb,amsthm}
\usepackage{hyperref}
\usepackage{makeidx}
\usepackage{enumerate}
\usepackage{graphicx,float,psfrag,epsfig}
\usepackage{epstopdf}
\usepackage{color}
\usepackage{enumitem}
\usepackage{subfig}
\usepackage{caption}
\usepackage{bigints}
\usepackage{mathtools}
\usepackage[mathscr]{euscript}

\DeclareSymbolFont{rsfs}{U}{rsfs}{m}{n}
\DeclareSymbolFontAlphabet{\mathscrsfs}{rsfs}

\numberwithin{equation}{section}

\newtheoremstyle{myexample}
    {\topsep}       
    {\topsep}
    {\rm }         
    {}               
    {\bf }             
    {.}                    
    {.5em}                  
    {}  

\newtheoremstyle{myremark}
    {\topsep} 
    {\topsep}   
    {\rm}                       
    {}                           
    {\bf}                       
    {.}                  
    {.5em}      
    {}

\newtheorem{claim}{Claim}[section]
\newtheorem{lemma}[claim]{Lemma}

\newtheorem{theorem}{Theorem}

\newtheorem{definition}[claim]{Definition}

\def\bv{{\boldsymbol v}}
\def\bb{{\boldsymbol b}}
\def\ba{{\boldsymbol a}}
\def\bB{{\boldsymbol B}}
\def\bA{{\boldsymbol A}}
\def\bx{{\boldsymbol x}}
\def\bz{{\boldsymbol z}}
\def\bp{{\boldsymbol p}}

\def\by{{\boldsymbol y}}
\def\bw{{\boldsymbol w}}
\def\bM{{\boldsymbol M}}
\def\bI{{\boldsymbol I}}
\def\bW{{\boldsymbol W}}
\def\btheta{{\boldsymbol \theta}}

\def\b0{{\boldsymbol 0}}
\def\sT{{\sf T}}

\definecolor{mydarkblue}{rgb}{0,0.08,0.45}

\begin{document}

\twocolumn[
\icmltitle{Landscape Connectivity and Dropout Stability of SGD Solutions for Over-parameterized Neural Networks}

\icmlkeywords{Machine Learning, ICML}

\icmlsetsymbol{equal}{*}

\begin{icmlauthorlist}
\icmlauthor{Alexander Shevchenko}{to}
\icmlauthor{Marco Mondelli}{to}
\end{icmlauthorlist}

\icmlaffiliation{to}{Institute of Science and Technology, Austria}

\icmlcorrespondingauthor{Alexander Shevchenko}{ashevche@ist.ac.at}

\vskip 0.3in
]

\printAffiliationsAndNotice{}

\begin{abstract}
The optimization of multilayer neural networks typically leads to a solution with zero training error, yet the landscape can exhibit spurious local minima and the minima can be disconnected. In this paper, we shed light on this phenomenon: we show that the combination of stochastic gradient descent (SGD) and over-parameterization makes the landscape of multilayer neural networks approximately connected and thus more favorable to optimization. More specifically, we prove that SGD solutions are connected via a piecewise linear path, and the increase in loss along this path vanishes as the number of neurons grows large. This result is a consequence of the fact that the parameters found by SGD are increasingly dropout stable as the network becomes wider. We show that, if we remove part of the neurons (and suitably rescale the remaining ones), the change in loss is independent of the total number of neurons, and it depends only on how many neurons are left. Our results exhibit a mild dependence on the input dimension: they are dimension-free for two-layer networks and require the number of neurons to scale linearly with the dimension for multilayer networks. We validate our theoretical findings with numerical experiments for different architectures and classification tasks.
\end{abstract}

\section{Introduction}

The recent successes of deep learning have two elements in common: \emph{(i)} a local search algorithm, e.g., stochastic gradient descent (SGD), and \emph{(ii)} an over-parameterized neural network. Even though the training problem can have several local minima \cite{auer1996exponentially} and is NP-hard in the worst case \cite{blum1989training}, the optimization of an over-parameterized network via SGD typically leads to a solution that has small training error and generalizes well. This fact has led to a focus on the theoretical understanding of neural networks' optimization landscape (see, e.g., \cite{livni2014computational, dauphin2014identifying, safran2016quality, pennington2017geometry} and the discussion in Section \ref{sec:rel}). However, most of the existing results either make strong assumptions on the model or do not provide a satisfactory scaling with respect to the parameters of the problem.

From the empirical viewpoint, it has been observed that, if we connect two minima of SGD with a line segment, the loss is large along this path \cite{goodfellow2014qualitatively, keskar2016large}. However, if the path is chosen in a more sophisticated way, one can connect the minima found by SGD via a piecewise linear path where the loss is approximately constant \cite{garipov2018loss, draxler2018essentially}. These findings suggest that the minima of SGD are not isolated points in parameter space, but rather they are approximately connected. In the recent paper \cite{kuditipudi2019explaining}, mode connectivity of multilayer ReLU networks is proved by assuming generic properties of well-trained networks, i.e., dropout stability and noise stability.

In this work, we consider multilayer neural networks trained by one-pass (or online) SGD with the square loss. We show that, as the number of neurons increases, \emph{(i)} the neural network becomes increasingly dropout stable, and \emph{(ii)} the optimization landscape becomes increasingly connected between SGD solutions. We establish quantitative bounds on how much the loss changes after the dropout procedure and along the path connecting two SGD solutions, and we relate this change in loss to the total number of neurons, the size of the dropout pattern, and the input dimension. By doing so, we give a theoretical justification to the empirical observation that the barriers between local minima tend to disappear as the neural network becomes larger \cite{draxler2018essentially}. More specifically, our main contributions can be summarized as follows:

\textbf{Two-layer networks.} We consider the training of a two-layer neural network $\hat{y}(\bx)= \frac{1}{N}\ba^{\sT}\sigma(\bW\bx)$ with $N$ neurons. First, we study the dropout stability of SGD solutions, namely, we bound the change in loss when $N-M$ neurons are removed from the trained network and $M$ remaining neurons are suitably rescaled: \emph{we show that the change in loss scales at most as $\sqrt{\log M/M}$, and therefore it does not depend on the number of neurons $N$ of the original network or on the dimension $d$ of the input}. Then, we characterize the landscape connectivity for the parameters obtained via SGD: \emph{we show that pairs of SGD solutions are connected via a piecewise linear path, and the loss along this path is no larger than the loss at the extremes plus a term that scales as $\sqrt{\log N/N}$}. Let us emphasize that the two solutions of SGD are obtained by running the algorithm on different samples (from the same data distribution), for different initializations, and for the different number of iterations. 

\textbf{Multilayer networks.} We consider the training of a general model of deep neural network with $L+1\ge 4$ layers, where each hidden layer contains $N$ neurons. This model includes as a special case $\hat{\by}(\bx)$ which is equal to 
\begin{equation}\label{eq:multispecial}
\frac{1}{N}\boldsymbol{W}_{L+1} \sigma_{L} \left(\cdots \left(\frac{1}{N}\boldsymbol{W}_2\sigma_1\left(\boldsymbol{W}_1 \bx\right)\right)  \cdots  \right)
\end{equation}
Our results are similar to those for two-layer networks: \emph{(i)} \emph{if we keep at least $M$ neurons in each layer, the change in loss scales at most as $\sqrt{(d+\log M)/M}$}; \emph{(ii)} \emph{pairs of SGD solutions are connected via a piecewise linear path, along which the loss does not increase more than $\sqrt{(d+\log N)/N}$}. In contrast with the two-layer case, these bounds are not dimension-free. However, the dependence on the input dimension $d$ is only linear, since the loss change vanishes as soon as $M, N\gg d$. We assume that, during SGD training, the parameters of the first and last layer are kept fixed, and they are regarded as random features \cite{rahimi2008random}. We believe that this assumption, as well as the requirement of having at least 4 layers, can be removed with an improved analysis. 

The proofs of dropout stability build on recent results concerning the  mean-field description of the SGD dynamics \cite{mei2019mean, araujo2019mean}, see also the discussion in Section \ref{sec:rel}. The proofs of landscape connectivity use ideas from \cite{kuditipudi2019explaining}.

{\bf Organization of the paper.} In Section \ref{sec:rel}, we succinctly review related work. In Section \ref{sec:main2}, we present our rigorous results for two-layer networks: we first assume that the activation function $\sigma$ is bounded, and then we provide an extension to unbounded activations. In Section \ref{sec:mainmul}, we present our results for multilayer networks. In Section \ref{sec:num}, we validate our findings with numerical experiments on fully-connected neural networks trained on MNIST and CIFAR-10 datasets. Finally, in Section \ref{sec:fin} we discuss additional connections to the literature and give directions for future work. All the proofs are deferred to the appendices in the supplementary material, which also contain additional numerical results. 

{\bf Notation.} We use bold symbols for vectors $\ba, \bb$, and capitalized bold symbols for matrices $\bA, \bB$. We denote by~$\|\ba\|_2$~the norm of $\ba$, by $\|\bA\|_{\rm op}$ the operator norm of~$\bA$, by $\langle \ba, \bb\rangle$ the scalar product of $\ba, \bb$, and by $\ba\odot\bb$ the Hadamard (or entrywise) product of $\ba,\bb$. Given an integer~$N$ and a real number $r\ge 1$, we set $[N]=\{1, \ldots, N\}$ and $[r]=\{1, \ldots, \lfloor r\rfloor\}$. Given a discrete set $\mathcal A$, we denote by~$|\mathcal A|$ its cardinality.

\section{Related Work}\label{sec:rel}

The landscape of several non-convex optimization problems has been studied in recent years, including empirical risk minimization \cite{mei2018landscape}, low rank matrix problems \cite{ge2017no}, matrix completion \cite{ge2016matrix}, and semi-definite programs \cite{boumal2016non}. Motivated by the extraordinary success of deep learning, a growing literature is focusing on the loss surfaces of neural networks. Under strong assumptions, in \cite{choromanska2015loss} the loss function is related to a spin glass and it is shown that local minima are located in a well-defined band. It has been shown that local minima are globally optimal in various settings: deep linear networks \cite{kawaguchi2016deep}; fully connected and convolutional neural networks with a wide layer containing more neurons than training samples \cite{nguyen2017loss, nguyen2018optimization};  deep networks with more neurons than training samples and skip connections \cite{nguyen2018loss}. Furthermore, if one of the layers is sufficiently wide, in \cite{nguyen2019connected} it is shown that sublevel sets are connected. Similar results are proved for binary classification in \cite{liang2018adding, liang2018understanding}. In \cite{freeman2016topology}, a two-layer neural networks with ReLU activations is considered, and it is shown that the landscape becomes approximately connected as the number of neurons increases. However, the energy gap scales exponentially with the input dimension. In \cite{venturi2018spurious}, it is shown that there are no spurious valleys when the number of neurons is larger than the intrinsic dimension of the networks. However, for many standard architectures, the intrinsic dimension of the network is infinite.

In this paper, we take a different view and relate the problem to a recent line of work, which shows that the behavior of neural networks trained by SGD tends to a mean field limit, as the number of neurons grows. This phenomenon has been first studied in two-layer neural networks in \cite{mei2018mean,rotskoff2018neural,chizat2018global,sirignano2018mean}. In particular, in \cite{mei2018mean}, it is shown that the SGD dynamics is well approximated by a Wasserstein gradient flow, given that the number of neurons exceeds the data dimension. Improved and dimension-free bounds are provided in \cite{mei2019mean}. Convergence to the global optimum is proved for noisy SGD in \cite{mei2018mean, chizat2018global}, without any explicit rate. A convergence rate which is exponential and dimension-free is proved in \cite{javanmard2019analysis} by exploiting the displacement convexity of the limit dynamics. An argument indicating convergence in a time polynomial in the dimension is provided in \cite{wei2018margin}, but for a different type of continuous flow. Fluctuations around the mean field limit are also studied in \cite{rotskoff2018neural, sirignano2019mean}. The multilayer case is tackled in \cite{nguyen2019mean, sirignano2019meand, araujo2019mean, nguyen2020rigorous}. In \cite{sirignano2019meand}, it is considered a (less natural) model where the number of neurons grows one layer at a time. In \cite{nguyen2019mean}, a formalism is developed to describe the mean field limit, but the results are not rigorous. Rigorous bounds between the SGD dynamics and a limit stochastic process are established in \cite{araujo2019mean}, where it is assumed that the first and last layer are not trained to simplify the analysis. A different approach based on the concept of neuronal embedding is put forward in \cite{nguyen2020rigorous}. In \cite{nguyen2020rigorous}, it is also provided a convergence result for three-layer networks, later generalized in the companion note \cite{nguyen2020rigorousnote}.

In a nutshell, existing mean-field analyses show that the dynamics of SGD is close to a limit stochastic process. However, the consequences of this fact remain largely unexplored, since the limit process is hard to analyze. In this work, we advance the mean-field theory of neural networks, and we provide the first theoretical guarantees on two phenomena widely observed in practice: dropout stability and mode connectivity of SGD solutions.

We remark that the mean-field regime considered in this paper is different from the ``lazy training'' regime that has recently received a lot of attention \cite{allen2019learning,allen2019convergence,chizat2019lazy, du2019gradient, du2019gradient2, jacot2018neural, li2018learning, zou2018stochastic}. In fact, in order to prove convergence of gradient descent in the lazy regime, it is crucially exploited that the parameters stay bounded in a certain region. On the contrary, in the mean field regime, the scaling of the gradient (see Eqs. \eqref{eq:SGD} and \eqref{eq:SGDmulti}) ensures that the parameters move away from the initialization. The connection between the mean-field and the lazy regime is investigated in Section 4 of \cite{mei2019mean} and in the recent paper \cite{chen2020mean}. We highlight that neural networks trained in the mean-field regime achieve results comparable to the state of the art for standard datasets, as demonstrated in the numerical results of Section \ref{sec:num}.

\section{Dropout Stability and Connectivity for Two-Layer Networks}\label{sec:main2}

\subsection{Setup}

We consider a two-layer neural network with $N$ neurons:
\begin{equation}\label{eq:2layer}
\textstyle\hat{y}_N(\boldsymbol{x}, \boldsymbol{\theta}) = \frac{1}{N}\sum_{i=1}^N  a_i \sigma( \boldsymbol{x}, \boldsymbol{w}_i),\vspace{-1mm}
\end{equation}
where $\boldsymbol{x}\in \mathbb{R}^d$ is a feature vector, $\hat{y}_N(\boldsymbol{x}, \boldsymbol{\theta})\in\mathbb R$ is the output of the network, $\boldsymbol{\theta} = (\boldsymbol{\theta}_1,\dots,\boldsymbol{\theta}_N)$, with $\boldsymbol{\theta}_i = (a_i, \boldsymbol{w}_i) \in \mathbb{R}^{D+1}$, are the parameters of the network and $\sigma:\mathbb R^d\times \mathbb R^{D}\to \mathbb R$ is an activation function.

A typical example is $\sigma(\bx, \bw)=\sigma(\langle\bx, \bw\rangle)$, for a scalar function $\sigma: \mathbb R \rightarrow \mathbb R$. In order to incorporate a bias term in the hidden layer, one can simply add the feature $1$ to $\boldsymbol{x}$ and adjust the shape of the parameters $\boldsymbol{w}_i$ accordingly. We are interested in minimizing the expected square loss (also known as population risk):
\begin{equation}\label{eq:risk}
L_N(\boldsymbol{\theta}) = \mathbb{E} \left\{\big(y - \hat{y}_N(\boldsymbol{x}, \boldsymbol{\theta})\big)^2\right\},
\end{equation}
where the expectation is taken over $(\bx, y)\sim \mathbb P$. To do so, we are given data $(\bx_k, y_k)_{k\ge 0}\overset{\rm {i.i.d.}}{\sim} \mathbb P$, and we learn the parameters of the network via stochastic gradient descent (SGD) with step size~$s_k$:
\begin{equation}\label{eq:SGD}
\begin{split}
&{\boldsymbol{\theta}}_{i}^{k+1}={\boldsymbol{\theta}}_{i}^{k} - s_k N \cdot \operatorname{Grad}_i(\boldsymbol{\theta}^{k}),\\ 
&\operatorname{Grad}_i(\boldsymbol{\theta}^{k}) = \nabla_{{\boldsymbol{\theta}}_{i}} \big(y_{k}-\hat{y}_N(\boldsymbol{x}_{k} , {\boldsymbol{\theta}}^{k})\big)^2,
\end{split}
\end{equation}
where $\boldsymbol{\theta}^k$ denotes the parameters after $k$ steps of SGD, and the parameters are initialized independently according to the distribution $\rho_0$. We consider a one-pass (or online) model, where each data point is used only once.

Given a neural network with parameters $\btheta$ and a subset~$\mathcal A$ of~$[N]$, the dropout network with parameters $\boldsymbol{\theta}_{\rm S}$ is obtained by setting to $0$ the outputs of the neurons indexed by $[N]\setminus  \mathcal A$ and by suitably rescaling the remaining outputs. Denote  by~$\hat{y}_{|\mathcal A|}(\boldsymbol{x}, \boldsymbol{\theta}_{\rm S})$ and $L_{|\mathcal A|}(\boldsymbol{\theta}_{\rm S})$ the output of the dropout network and its expected square loss, respectively. In formulas,
\begin{equation}\label{eq:dropout}
\begin{split}
\hat{y}_{|\mathcal A|}(\boldsymbol{x}, \boldsymbol{\theta}_{\rm S}) &= \frac{1}{|\mathcal A|}\sum\limits_{i\in \mathcal A}  a_i \sigma( \boldsymbol{x}, \boldsymbol{w}_i),\\
L_{|\mathcal A|}(\boldsymbol{\theta}_{\rm S}) &= \mathbb{E} \left\{\big(y - \hat{y}_{|\mathcal A|}(\boldsymbol{x}, \boldsymbol{\theta}_{\rm S})\big)^2\right\}.
\end{split}
\end{equation}

Let us compare the original network \eqref{eq:2layer} with the dropout network \eqref{eq:dropout}: $\bw_i$ does not change, $a_i$ is rescaled by $|\mathcal A|/|N|$ and in \eqref{eq:dropout} we sum over $|\mathcal A|$ neurons (while in \eqref{eq:2layer} the sum is over $N$ neurons). This is equivalent to setting $|N|-|\mathcal A|$ neurons to zero and rescaling the others by a factor, as in \cite{kuditipudi2019explaining}.

We now define the notions of dropout stability and connectivity for network parameters. 

\begin{definition}[Dropout stability]\label{def:dropoutstab}
Given $\mathcal A\subseteq [N]$, we say that $\boldsymbol{\theta}$ is $\varepsilon_{\rm D}$-dropout stable if 
\begin{equation}\label{eq:dropstab}
|L_N(\boldsymbol{\theta})-L_{|\mathcal A|}(\boldsymbol{\theta}_{\rm S})|\le \varepsilon_{\rm D}.
\end{equation}
\end{definition}

\begin{definition}[Connectivity]\label{def:conn}
We say that two parameters $\boldsymbol{\theta}$ and $\boldsymbol{\theta}'$ are $\varepsilon_{\rm C}$-connected if there exists a continuous path in parameter space $\pi: [0,1] \rightarrow \mathbb{R}^{D\times N}$, such that $\pi(0) = \boldsymbol{\theta}$ and  $\pi(1) = \boldsymbol{\theta}'$ with
\begin{equation}\label{eq:conneq}
L_N(\pi(t)) \leq \max(L_N(\boldsymbol{\theta}), L_N(\boldsymbol{\theta}')) + \varepsilon_{\rm C}.
\end{equation}
\end{definition}

\subsection{Results for Bounded Activations}\label{sec:res2}

We make the following assumptions on the learning rate $s_k$, the data distribution $(\bx, y)\sim \mathbb P$, the activation function $\sigma$, and the initialization $\rho_0$:

\textbf{(A1)} $s_k=\alpha\xi(k\alpha)$, where $\xi:\mathbb R_{\ge 0}\to\mathbb R_{>0}$ is bounded by~$K_1$ and $K_1$-Lipschitz.\\
\textbf{(A2)} The response variables $y$ are bounded by $K_2$ and the gradient $\nabla_{\boldsymbol{w}} \sigma(\boldsymbol{x},\boldsymbol{w})$ is $K_2$ sub-gaussian when $\bx \sim \mathbb P$.\\
\textbf{(A3)} The activation function $\sigma$ is bounded by $K_3$ and differentiable, with gradient bounded by $K_3$ and $K_3$-Lipschitz.\\
\textbf{(A4)} The initialization $\rho_0 $ is supported on $|a_i^0| \leq K_4$.

We are now ready to present our results, which are proved in Appendix \ref{app:th1} in the supplementary material.

\begin{theorem}[Two-layer]\label{th:2layerbdd}
Assume that conditions \textbf{(A1)}-\textbf{(A4)} hold, and fix $T\ge 1$. Let $\boldsymbol{\theta}^k$ be obtained by running $k$ steps of the SGD algorithm \eqref{eq:SGD} with data $\{(\bx_j, y_j)\}_{j= 0}^k\overset{\rm {i.i.d.}}{\sim} \mathbb P$ and initialization $\rho_0$. Then, the following results hold:

\textbf{(A)} Pick $\mathcal A\subseteq [N]$ independent of $\boldsymbol{\theta}^k$. Then, with probability at least $1-e^{-z^2}$, for all $k\in [T/\alpha]$, $\boldsymbol{\theta}^k$ is $\varepsilon_{\rm D}$-dropout stable with $\varepsilon_{\rm D}$ equal to
\begin{align}\label{eq:epsbdd}
   K e^{KT^{3}} \left( \frac{\sqrt{\log |\mathcal A|}+z}{\sqrt{|\mathcal A|}} + \sqrt{\alpha}\big(\sqrt{D+\log N}+z\big)\right),\nonumber\\
\end{align}
where the constant $K$ depends only on the constants $K_i$ of the assumptions.

\textbf{(B)} Fix $T'\ge 1$ and let $(\boldsymbol{\theta}')^{k'}$ be obtained by running $k'$ steps of SGD with data $\{(\bx_j', y_j')\}_{j= 0}^{k'}\overset{\rm {i.i.d.}}{\sim} \mathbb P$ and initialization $\rho_0'$ that satisfies {\bf (A4)}. Then, with probability at least $1-e^{-z^2}$, for all $k\in [T/\alpha]$ and $k'\in [T'/\alpha]$, $\boldsymbol{\theta}^k$ and $(\boldsymbol{\theta}')^{k'}$ are $\varepsilon_{\rm C}$-connected with $\varepsilon_{\rm C}$ equal to
\begin{align}\label{eq:epsbddc}
    K e^{KT_{\rm max}^3} \left(\frac{\sqrt{\log N}+z}{\sqrt{N}} + \sqrt{\alpha}\big(\sqrt{D+\log N}+z\big)\right),\nonumber\\\vspace{-4mm}
\end{align}
where $T_{\rm max} = \max(T, T')$. Furthermore, the path connecting $\boldsymbol{\theta}^k$ with $(\boldsymbol{\theta}')^{k'}$ consists of 7 line segments.

\end{theorem}

The result (A) characterizes the change in loss when only $|\mathcal A|$ neurons remain in the network. In particular, the change in loss scales as $\sqrt{
\log |\mathcal A|/|\mathcal A|}+\sqrt{\alpha(D+\log N)}$, where $N$ is the total number of neurons, $D$ is the dimension of the neurons and $\alpha$ is the step size of SGD. This quantity vanishes as long as $|\mathcal A|\gg 1$ and $\alpha \ll 1/(D+\log N)$. Note that the number of training samples $k$ is such that $k\alpha$ is a constant. Thus, the condition $\alpha \ll 1/(D+\log N)$ implies that $k$ needs to scale only logarithmically with $N$. Furthermore, the condition $|\mathcal A|\gg 1$ implies that $|\mathcal A|$ does not need to scale with $N$, $D$. The proof builds on the machinery developed in \cite{mei2019mean} to provide a mean-field approximation to the dynamics of SGD. In \cite{mei2019mean}, it is shown that, as $N\to\infty$ and $\alpha\to 0$, the parameters $\boldsymbol{\theta}^k$ obtained by running $k$ steps of SGD with step size $\alpha$ are close to $N$ i.i.d. particles that evolve according to a nonlinear dynamics at time $k\alpha$. Here, the idea is to show that \emph{(i)} the parameters $\boldsymbol{\theta}_{\rm S}^k$ are also close to $|\mathcal A|$ such i.i.d. particles, and \emph{(ii)} the quantities $L_N(\boldsymbol{\theta}^k)$ and $L_{|\mathcal A|}(\boldsymbol{\theta}^k_{\rm S})$ concentrate to the same limit value, which represents the limit loss of the nonlinear dynamics.

The result (B) shows that we can connect two different solutions of SGD via a simple path. Note that the two solutions can be obtained by running SGD for the different number of iterations ($k'\neq k$), for different training datasets ($(\bx_j, y_j)\neq (\bx_j', y_j')$) and for different initializations of SGD ($\rho_0\neq\rho_0'$). The proof uses ideas from \cite{kuditipudi2019explaining}. In that work, the authors consider a multilayer neural network with ReLU activations and show how to find a piecewise linear path between two solutions that are dropout stable with $|\mathcal A|=N/2$. In fact, $\varepsilon_{\rm C}$ has a similar scaling to $\varepsilon_{\rm D}$ after setting $|\mathcal A|=N/2$. We are also able to show (and, consequently, exploit) a more general notion of dropout stability for the trained network. In fact, \cite{kuditipudi2019explaining} requires the existence of a single dropout pattern, while here we give a bound for any fixed  dropout pattern (as long as it does not depend on SGD).

The bounds in Theorem \ref{th:2layerbdd} exhibit an exponential dependence on $T$. We remark that, in the mean-field regime, the number of samples $k$ is large, the step size $\alpha$ is small, and $T=k\alpha$ is a constant. In fact, $T$ is the evolution time of the limit stochastic process (which does not depend on $N$, $\alpha$). Empirically, the value of $T$ needed to achieve good accuracy is quite small: $T=1$ gives $<16\%$ error on CIFAR-10, see Section \ref{sec:num}. The exponential dependence on $T$ is common to all existing mean-field analyses, and improving it is an open question. The assumptions on the learning rate, the data distribution and the initialization are mild and only require some regularity. The assumptions on the activation function are fulfilled in several practical settings: $\sigma(\bx, \bw)=\sigma(\langle\bx, \bw\rangle)$, where $\sigma:\mathbb R\to\mathbb R$ is, e.g., the sigmoid  or the hyperbolic tangent.

\subsection{Extension to Unbounded Activations}

Note that Theorem \ref{th:2layerbdd} requires that the activation function is bounded. We can relax this assumption, at the cost of a less tight dependence on the time $T$ of the evolution. In particular, assume further that \emph{(i)} the feature vectors $\bx$ and the initialization $\rho_0$ are bounded, and that \emph{(ii)} the loss at each step of SGD is uniformly bounded, i.e., $\max_{j} |y_j - \hat{y}_N(\boldsymbol{x}_j, \boldsymbol{\theta}^j)| \leq K_5$. This last requirement is reasonable, since the objective of SGD is to minimize such a loss. Then, the results of Theorem \ref{th:2layerbdd} hold also for unbounded $\sigma$, where the term $K e^{KT^3}$ is replaced by a generic $K(T)$, which depends on $T$ and on the constants $K_i$ of the assumptions. The simulation results of Section \ref{sec:num} show that such a dependence on $T$ is mild in practical settings.

The formal statement and the proof of this result is contained in Appendix \ref{app:ubb} in the supplementary material. The idea is to show that, if the parameters of the neural network are initialized with a bounded distribution, then they stay bounded for any finite time $T$ of the SGD evolution. Thus, the SGD evolution does not change if we substitute the unbounded activation function with a bounded one, and we can apply the results for bounded $\sigma$.

\section{Dropout Stability and Connectivity for Multilayer Networks}\label{sec:mainmul}

\subsection{Setup}

We consider a neural network with $L+1\ge 4$ layers, where each hidden layer contains $N$ neurons. Given the input feature vector $\bx\in\mathbb R^{d_0}$, the first layer activations $\bz_{i_{1}}^{(1)}$ for $i_1\in[N]$ have form
$$
\sigma^{(0)}\left(\boldsymbol{x}, \btheta_{i_{1}}^{(0)}\right),\quad \boldsymbol{\theta}_{i_{1}}^{(0)}\in\mathbb R^{D_0}
$$
the intermediate layer $\ell \in [L-1]$ activations $\bz_{i_{\ell+1}}^{(\ell+1)}\left(\boldsymbol{x}, \boldsymbol{\theta}\right)$ for $i_{\ell+1}\in[N]$ are defined as follows
\begin{align*}
&\frac{1}{N} \sum_{i_{\ell}=1}^{N} \ba_{i_\ell, i_{\ell+1}}^{(\ell)}\odot\sigma^{(\ell)}\left(\bz_{i_{\ell}}^{(\ell)}\left(\boldsymbol{x}, \boldsymbol{\theta}\right), \boldsymbol{w}_{i_{\ell}, i_{\ell+1}}^{(\ell)}\right), \\
&\boldsymbol{\theta}^{(\ell)}_{i_\ell, i_{\ell+1}}=(\ba_{i_\ell, i_{\ell+1}}^{(\ell)}, \boldsymbol{w}_{i_{\ell}, i_{\ell+1}}^{(\ell)}) \in\mathbb R^{D_\ell+d_{\ell+1}},
\end{align*}
and the output of network  is given by
\begin{align}\label{eq:mlayer}
&\widehat{\boldsymbol{y}}_{N}\left(\boldsymbol{x}, \boldsymbol{\theta}\right)=\frac{1}{N} \sum_{i_{L}=1}^{N} \ba_{i_L}^{(L)}\odot\sigma^{(L)}\left(\bz_{i_{L}}^{(L)}\left(\boldsymbol{x}, \boldsymbol{\theta}\right), \boldsymbol{w}_{i_L}^{(L)}\right),\nonumber\\
&\boldsymbol{\theta}_{i_L}^{(L)}=(\ba_{i_L}^{(L)}, \boldsymbol{w}_{i_L}^{(L)})\in\mathbb R^{D_L + d_{L+1}}, \quad i_L \in [N].
\end{align}
Here, $\sigma^{(\ell)}:  \mathbb{R}^{d_{\ell}} \times \mathbb{R}^{D_{\ell}} \rightarrow \mathbb{R}^{d_{\ell+1}}$ ($\ell \in\{0, \ldots, L\}$) are the activation functions, and $\boldsymbol{\theta}$ contains the parameters of the network, which are $\boldsymbol{\theta}_{i_{1}}^{(0)}$, $\boldsymbol{\theta}^{(\ell)}_{i_\ell, i_{\ell+1}}$ and $\boldsymbol{\theta}_{i_L}^{(L)}$.

Note that \eqref{eq:mlayer} includes the model \eqref{eq:multispecial} as a special case. To see this, consider the following setting: pick $D_0=d_0$ and stack the parameters $\btheta_{i_1}^{(0)}\in \mathbb R^{d_0}$ into the rows of the matrix $\bW_1\in \mathbb R^{N\times d_0}$; for $i\in [L-1]$, pick $D_\ell=1$ and stack the scalar parameters $\ba_{i_\ell, i_{\ell+1}}^{(\ell)}\in\mathbb R$ into the matrix $\bW_{\ell+1}\in\mathbb R^{N\times N}$; pick $D_L=d_{L+1}$ and stack the parameters $\ba_{i_L}^{(L)}\in\mathbb R^{d_{L+1}}$ into the columns of the matrix $\bW_{L+1}\in\mathbb R^{d_{L+1}\times N}$; finally, assume that the activation function $\sigma^{(\ell)}$ does not depend on $\bw_{i_\ell, i_{\ell+1}}^{(\ell)}$ for $\ell\in [L-1]$ and that $\sigma^{(L)}$ does not depend on $\bw_{i_L}^{(L)}$. Then, in this setting, \eqref{eq:mlayer} can be reduced to \eqref{eq:multispecial}.

 We are interested in minimizing the expected square loss: 
\begin{equation}\label{eq:riskm}
L_N(\boldsymbol{\theta}) = \mathbb{E}\left\{ \big\|\boldsymbol{y} - \widehat{\boldsymbol{y}}_N\left(\boldsymbol{x}, \boldsymbol{\theta}\right)\big\|_2^2\right\},
\end{equation}
where the expectation is taken over $(\bx, \by)\sim \mathbb P$. To do so, we are given data $(\bx_k, \by_k)_{k\ge 0}\overset{\rm {i.i.d.}}{\sim} \mathbb P$, we run SGD with step size $s_k$ for the intermediate layers $\ell \in [L-1]$, and we fix first and last layer:
\begin{align}\label{eq:SGDmulti}
&\boldsymbol{\theta}^{(\ell)}_{i_{\ell}, i_{\ell + 1}}(k+1) = \boldsymbol{\theta}^{(\ell)}_{i_{\ell}, i_{\ell + 1}}(k) - s_k N^2 {\rm Grad}^{(\ell)}_{i_{\ell}, i_{\ell + 1}}
\big(\boldsymbol{\theta}(k)\big),\nonumber\\[0.5em]
&{\rm Grad}^{(\ell)}_{i_{\ell}, i_{\ell + 1}}
\big(\boldsymbol{\theta}(k)\big) = 
\nabla_{\boldsymbol{\theta}^{(\ell)}_{i_{\ell}, i_{\ell + 1}}}
\big\|\boldsymbol{y}_k-\widehat{\boldsymbol{y}}_{N}\left(\boldsymbol{x}_k, \boldsymbol{\theta}(k)\right)\big\|^2_2,\nonumber\\[0.5em]
&\boldsymbol{\theta}_{i_{1}}^{(0)}(k+1) = \boldsymbol{\theta}_{i_{1}}^{(0)}(k),\quad
\boldsymbol{\theta}_{i_L}^{(L)}(k+1) = \boldsymbol{\theta}_{i_L}^{(L)}(k),
\end{align}

where $\boldsymbol{\theta}(k)$ contains the parameters of the network after $k$ steps of SGD. As in the two-layer setting, we consider a one-pass model and the parameters are initialized independently, i.e., $\{\btheta_{i_1}^{(0)}(0)\}_{i_1\in [N]}\overset{\rm {i.i.d.}}{\sim}\rho_0^{(0)}$, $\{\btheta^{(\ell)}_{i_\ell, i_{\ell+1}}(0)\}_{i_\ell, i_{\ell+1}\in [N]} \overset{\rm {i.i.d.}}{\sim}\rho_0^{(\ell)}$,  for $\ell\in [L-1]$, and $\{\btheta_{i_L}^{(L)}(0)\}_{i_L\in [N]}\overset{\rm {i.i.d.}}{\sim}\rho_0^{(L)}$. 

The gradients of $\widehat{\boldsymbol{y}}_{N}$ with respect to the parameters of the network can be computed via backpropagation \cite{goodfellow2016deep}. By doing so (see \citet[Section 3.3]{araujo2019mean}), we obtain that $\boldsymbol{\theta}^{(\ell)}_{i_{\ell}, i_{\ell + 1}}$ evolves at a time scale of $1/N^2$. Thus, we multiply the step size $s_k$ in \eqref{eq:SGDmulti} with the factor $N^2$ in order to avoid falling into the ``lazy training'' regime. In lazy training, the parameters hardly vary but the method still converges to zero training loss, and this regime has received a lot of attention recently \cite{jacot2018neural, li2018learning, zou2018stochastic, du2019gradient, du2019gradient2, allen2019convergence, allen2019learning, chizat2019lazy}. Let us emphasize that the SGD scalings in \eqref{eq:SGD} and \eqref{eq:SGDmulti} imply that the parameters move as long as the product of the number of iterations with the step size is non-vanishing.

Note also that the parameters of layers $\ell=0$ and $\ell=L$, i.e., $\{\btheta_{i_1}^{(0)}\}_{i_1\in [N]}$ and $\{\btheta_{i_L}^{(L)}\}_{i_L\in [N]}$, stay fixed to their initial values. This is done for technical reasons. In fact, by computing the backpropagation equations, one obtains that $\btheta_{i_1}^{(0)}$ and $\btheta_{i_L}^{(L)}$ evolve at a time scale of $1/N$, which makes it challenging to analyze their trajectories. We regard the parameters $\btheta_{i_1}^{(0)}$ and $\btheta_{i_L}^{(L)}$ as random features \cite{rahimi2008random} close to the input and the output.

Given a neural network with parameters $\btheta$ and subsets $\mathcal A_1, \ldots, \mathcal A_L$ of $[N]$, the dropout network with parameters $\boldsymbol{\theta}_{\rm S}$ is obtained by setting to $0$ the outputs of the neurons indexed by $[N]\setminus\mathcal A_i$ at layer $i$ and by suitably rescaling the remaining outputs. With an abuse of notation, denote by $\widehat{\boldsymbol{y}}_{|\mathcal A|}\left(\boldsymbol{x}, \boldsymbol{\theta}_{\rm S}\right)$ and $L_{|\mathcal A|}(\boldsymbol{\theta}_{\rm S})$ the output of the dropout network and its expected square loss, respectively. In formulas, the dropout version of activations $\bz_{i_{\ell+1}}^{(\ell+1)}\left(\boldsymbol{x}, \boldsymbol{\theta}_{\rm S}\right)$ of layer $\ell \in [L-1]$ for $i_{\ell+1}\in\mathcal A_{\ell+1}$ are given by
$$
\frac{1}{|\mathcal{A}_{\ell}|} \sum_{i_{\ell}\in \mathcal{A}_{\ell}} \ba_{i_{\ell}, i_{\ell+1}}^{(\ell)}\odot\sigma^{(\ell)}\left(\bz_{i_{\ell}}^{(\ell)}\left(\boldsymbol{x}, \boldsymbol{\theta}_{\rm S}\right), \bw_{i_{\ell}, i_{\ell+1}}^{(\ell)}\right),
$$
the output of dropout network $\widehat{\boldsymbol{y}}_{|\mathcal A|}\left(\boldsymbol{x}, \boldsymbol{\theta}_{\rm S}\right)$ takes the form
$$
\frac{1}{|\mathcal A_L|} \sum_{i_{L}\in\mathcal A_L}  \ba_{i_L}^{(L)}\odot\sigma^{(L)}\left(\bz_{i_{L}}^{(L)}\left(\boldsymbol{x}, \boldsymbol{\theta}_{\rm S}\right), \bw_{i_{L}}^{(L)}\right),
$$
and, consequently, the expected square loss is defined by
$$
L_{|\mathcal A|}(\boldsymbol{\theta}_{\rm S}) = \mathbb{E}\left\{ \big\|\boldsymbol{y} - \widehat{\boldsymbol{y}}_{|\mathcal A|}\left(\boldsymbol{x}, \boldsymbol{\theta}_{\rm S}\right)\big\|_2^2\right\},
$$
where $\bz_{i_{1}}^{(1)}\left(\boldsymbol{x}, \boldsymbol{\theta}_{\rm S}\right)=\bz_{i_{1}}^{(1)}\left(\boldsymbol{x}, \boldsymbol{\theta}\right)$ for $i_1\in\mathcal A_1$. The definitions of dropout stability and connectivity are analogous to those for two-layer networks: \emph{(i)} $\boldsymbol{\theta}$ is $\varepsilon_{\rm D}$-dropout stable if \eqref{eq:dropstab} holds; and \emph{(ii)} $\boldsymbol{\theta}$ and $\boldsymbol{\theta}'$ are $\varepsilon_{\rm C}$-connected if they are connected by a continuous path in parameter space such that \eqref{eq:conneq} holds. 

\begin{figure*}[t!]
    \centering
    \subfloat[MNIST, two-layer]{\includegraphics[width=1.04\columnwidth]{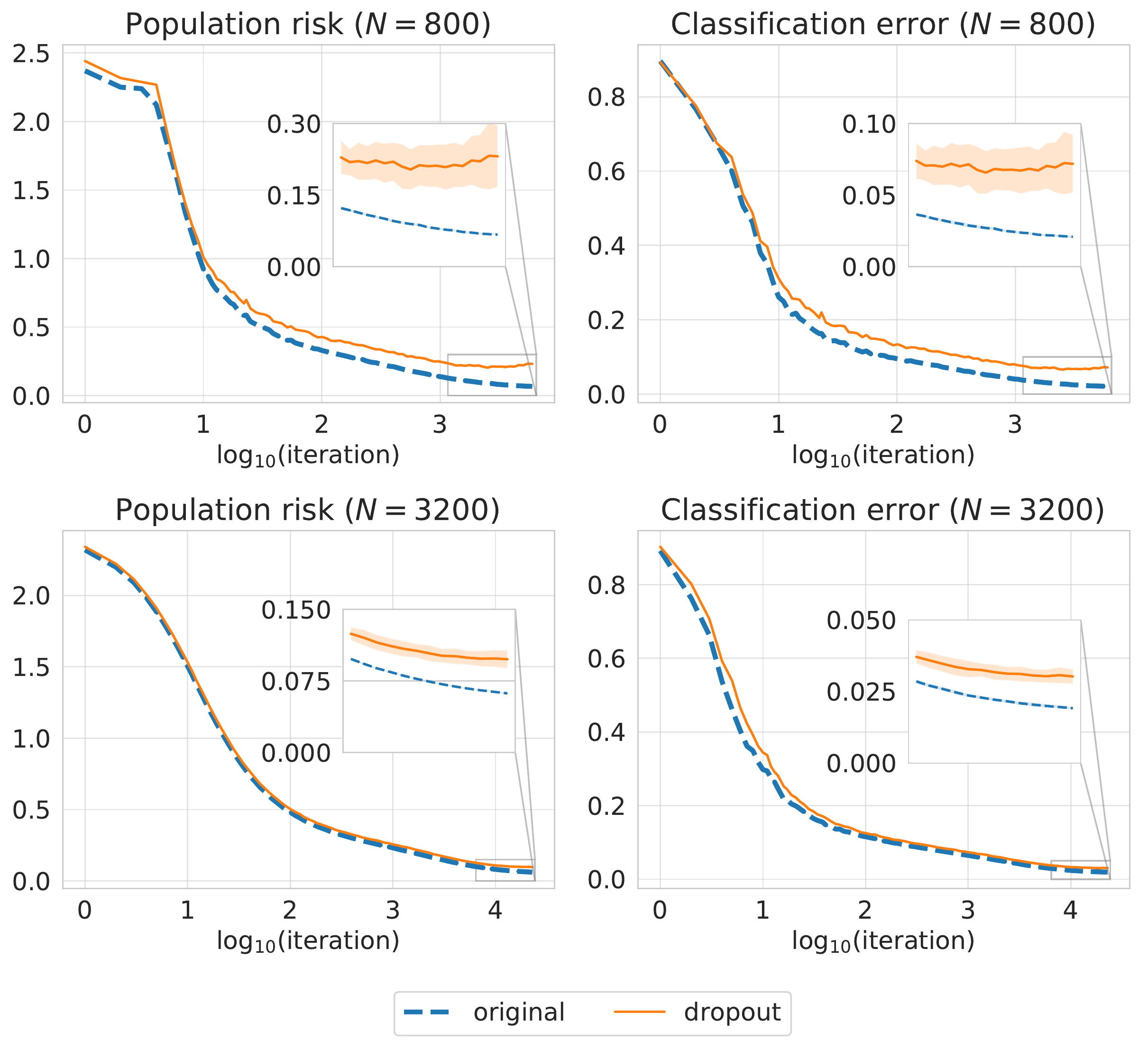}}
    \subfloat[CIFAR-10, three-layer]{\includegraphics[width=1.04\columnwidth]{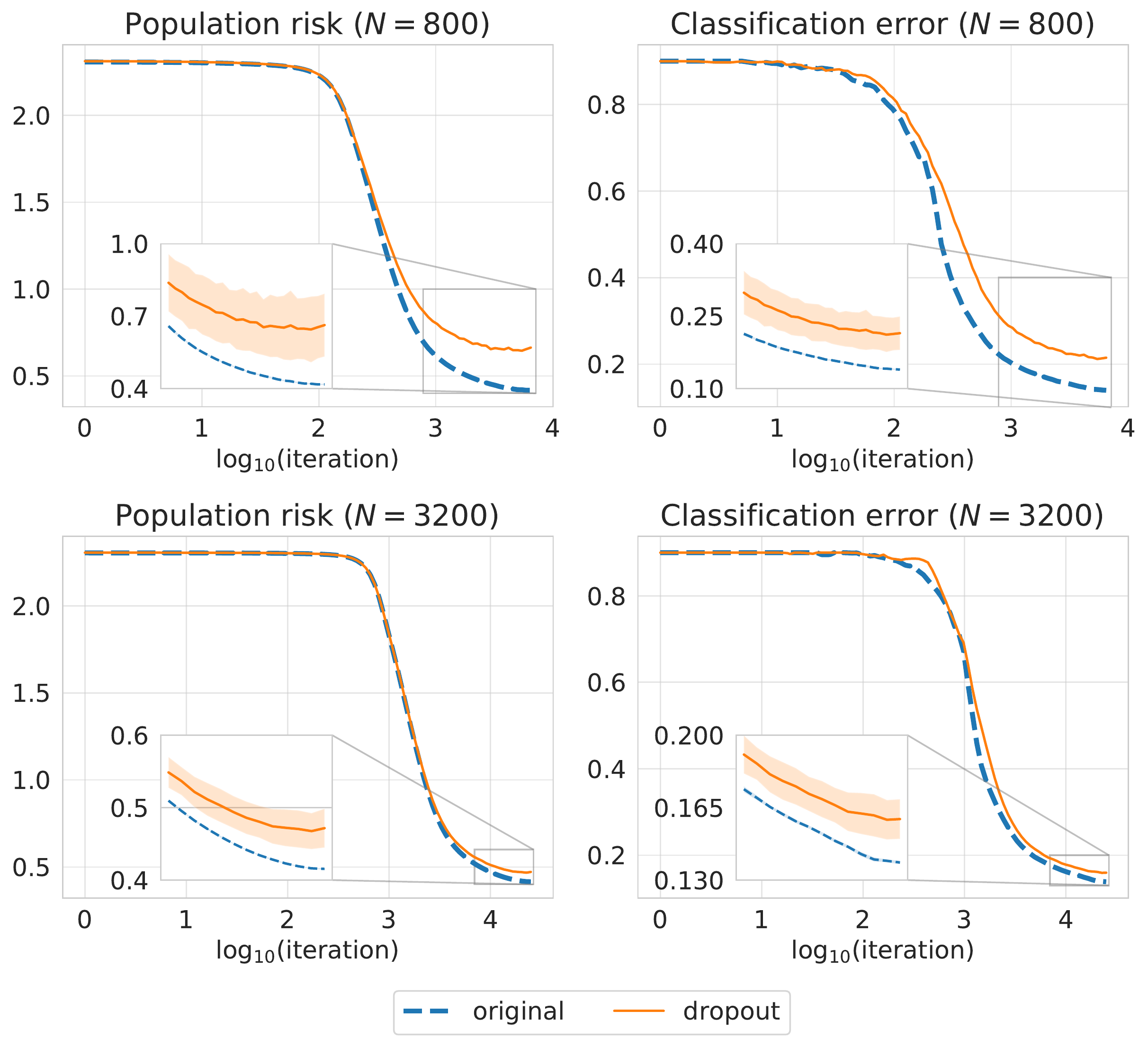}}
\caption{Comparison of population risk and classification error between the trained network (blue dashed curve) and the dropout network (orange curve). In the full scale plot, we show the average values, and in the zoomed version we also provide the error bar.}\label{fig:1}
\end{figure*}

\subsection{Results}

We make the following assumptions on the learning rate $s_k$, the data distribution $(\bx, \by)\sim \mathbb P$, the activation functions $\sigma^{(\ell)}$, and the initializations $\rho_0^{(\ell)}$:

\textbf{(B1)} $s_k=\alpha\xi(k\alpha)$, where $\xi:\mathbb R_{\ge 0}\to\mathbb R_{>0}$ is bounded by $K_1$ and $K_1$-Lipschitz.\\
\textbf{(B2)} The response variables $\by$ are bounded by $K_2$.\\
\textbf{(B3)} For $\ell\in\{0, \ldots, L\}$, the activation function $\sigma^{(\ell)}$ is bounded by $K_3$, with Fr\'echet derivative bounded by $K_3$ and $K_3$-Lipschitz.\\
\textbf{(B4)} The initializations $\{\rho_0^{(\ell)}\}_{\ell=0}^L$ have finite first moment and they are supported on $\|\ba_{i_\ell, i_{\ell+1}}^{(\ell)}(0)\|_2\le K_4$ for $\ell\in [L-1]$, and $\|\ba_{i_L}^{(L)}(0)\|_2\le K_4$.

We are now ready to present our results, which are proved in Appendix \ref{app:thmulti} in the supplementary material.

\begin{figure*}[t!]
    \centering
    \subfloat[MNIST, two-layer]{\includegraphics[width=\columnwidth]{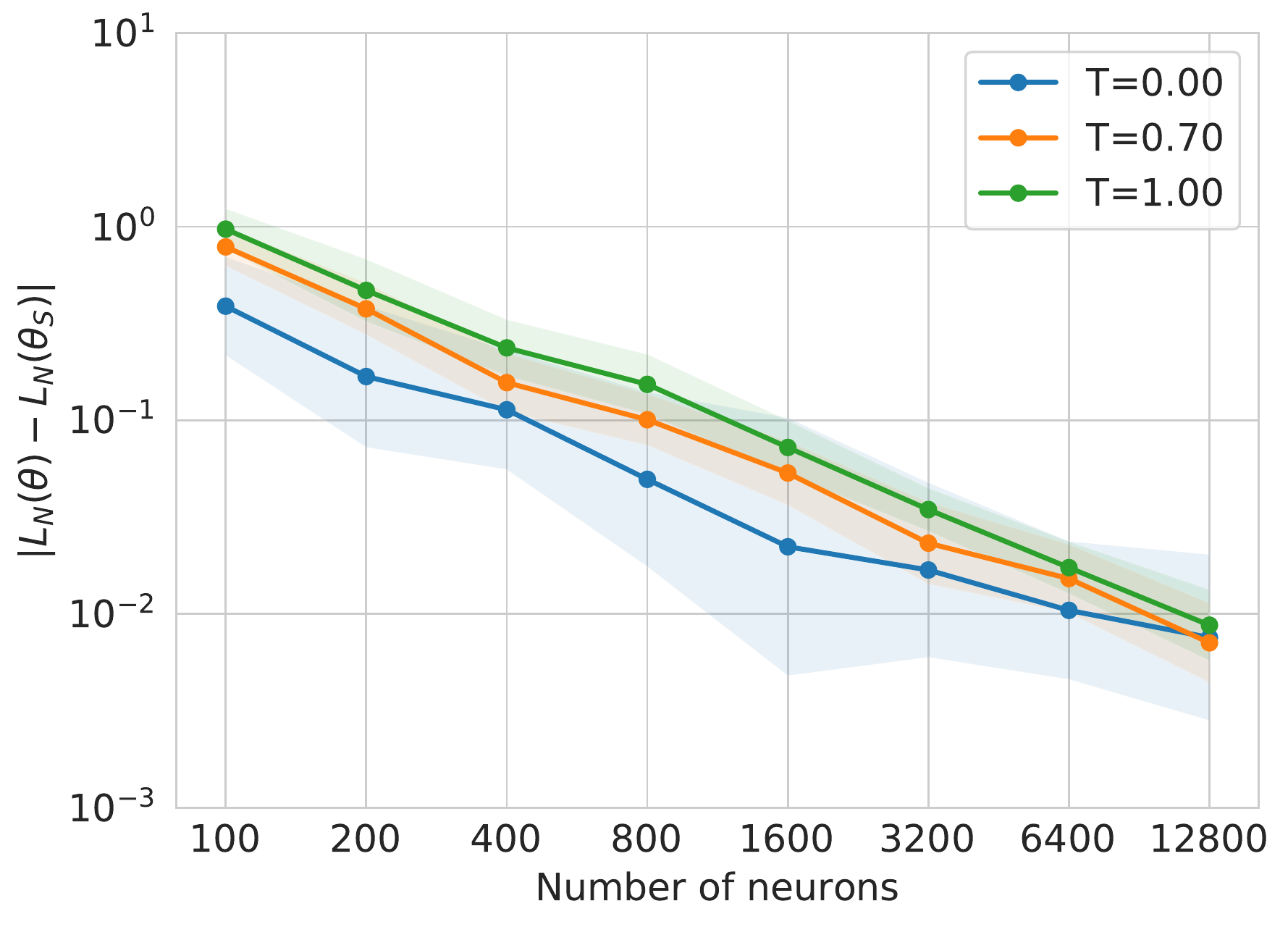}}
    \subfloat[CIFAR-10, three-layer]{\includegraphics[width=\columnwidth]{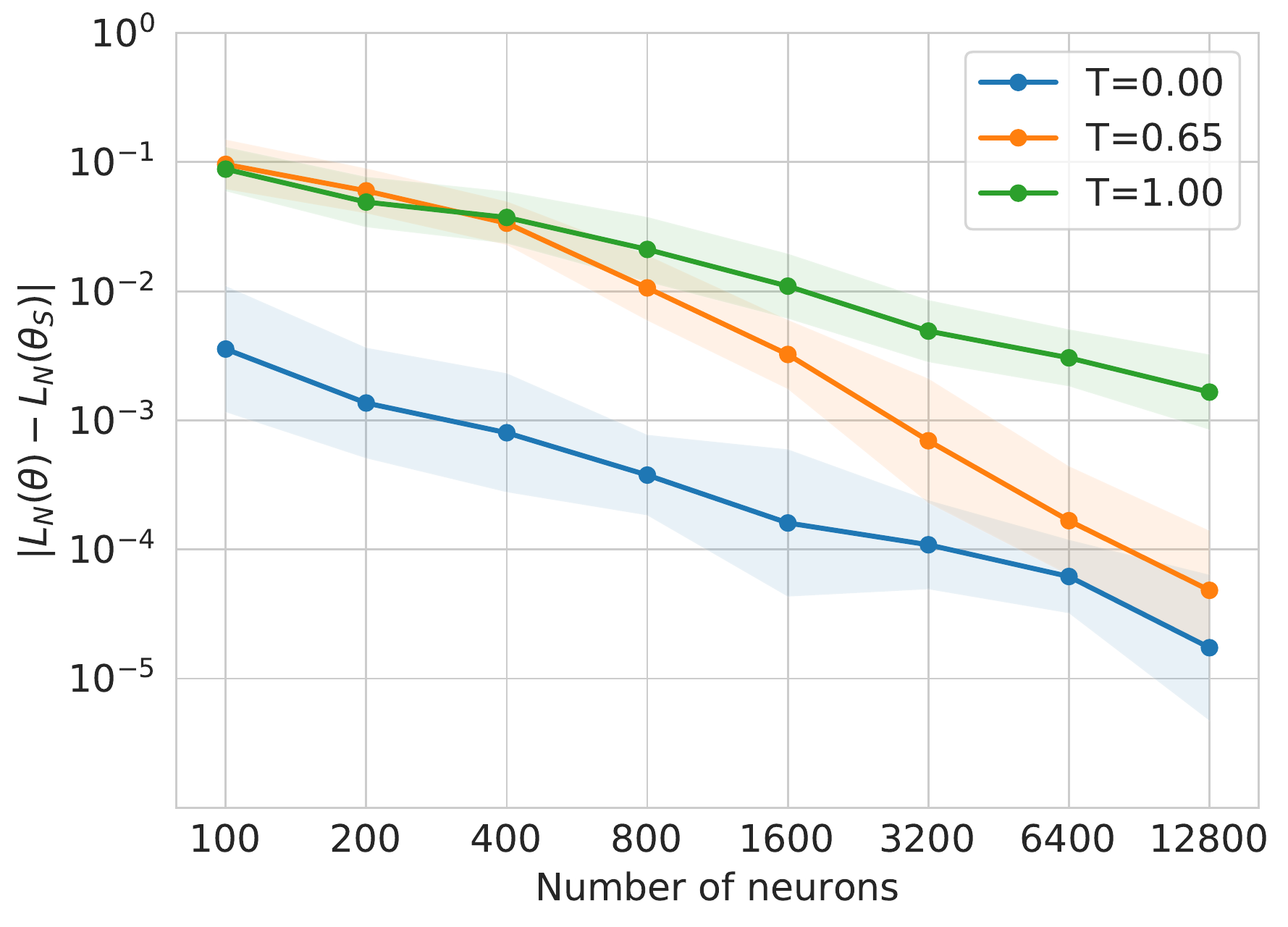}}
    \caption{Change in loss after removing half of the neurons from each layer, as a function of the number of neurons $N$ of the full network.}\label{fig:2}
\end{figure*}

\begin{theorem}[Multilayer]\label{th:2layerbddmulti}
Assume that conditions \textbf{(B1)}-\textbf{(B4)} hold, let $\boldsymbol{\theta}(k)$ be obtained by running $k$ steps of the SGD algorithm \eqref{eq:SGDmulti} with data $\{(\bx_j, \by_j)\}_{j= 0}^k\overset{\rm {i.i.d.}}{\sim} \mathbb P$ and initializations $\{\rho_0^{(\ell)}\}_{\ell=0}^L$, and define $T=k\alpha>0$. Then, the following results hold:

\textbf{(A)} Pick $\mathcal A_1, \ldots, \mathcal A_L\subseteq [N]$ independent of $\btheta(k)$. Then, with probability at least $1-e^{-z^2}$, $\boldsymbol{\theta}(k)$ is $\varepsilon_{\rm D}$-dropout stable with $\varepsilon_{\rm D}$ equal to
\begin{align}\label{eq:epsbddmulti}
    K(T, L) \left(\frac{\sqrt{d}+z}{\sqrt{A_{\rm min}}} +\sqrt{\frac{\log N}{N}}+ \sqrt{\alpha}\big(\sqrt{d+\log N}+z\big)\right)\nonumber\\
\end{align}
where $A_{\rm min}=\min_{i\in [L]}|\mathcal A_i|$, $d=\max_{\ell\in\{0, \ldots, L+1\}}d_\ell$ and the constant $K(T, L)$ depends on $T, L$ and on the constants $K_i$ of the assumptions.

\textbf{(B)} Let $\boldsymbol{\theta}'(k')$ be obtained by running $k'$ steps of the SGD algorithm \eqref{eq:SGDmulti} with data $\{(\bx_j', \by_j')\}_{j= 0}^{k'}\overset{\rm {i.i.d.}}{\sim} \mathbb P$ and initializations $\{(\rho_0^{(\ell)})'\}_{\ell=0}^L$ that satisfy \textbf{(B4)}, and define $T'=k'\alpha >0$.  Then, with probability at least $1-e^{-z^2}$, $\boldsymbol{\theta}(k)$ and $\boldsymbol{\theta}'(k')$ are $\varepsilon_{\rm C}$-connected with $\varepsilon_{\rm C}$ equal
\begin{align}\label{eq:epsbddmultic}
    K(T_{\rm max}, L)\hspace{-1mm} \left(\frac{\sqrt{d+\log N}+z}{\sqrt{N}} + \sqrt{\alpha}\big(\sqrt{d+\log N}+z\big)\right)\nonumber\\
\end{align}
where $T_{\rm max} = \max(T, T')$.
\end{theorem}

The results are similar in spirit to those of Theorem \ref{th:2layerbdd}, but the analysis is more involved. We remark that, differently from the two-layer case, the ideal particles are not independent, see Remark 5.6 of \cite{araujo2019mean}. We exploit a bound on the norm of the weights during training (see Lemma \ref{lemma:normbdmulti} in Appendix \ref{app:thmultiA}) and a bound on the \emph{maximum} distance between SGD weights and weights of ideal particles. Our analysis improves upon \cite{araujo2019mean}, where the bound is on the \emph{average} distance between SGD and ideal-particle weights (compare \eqref{eq:newnorm} in Appendix \ref{app:thmultiA} and (10.1) in \cite{araujo2019mean}). This improvement is essential to show dropout stability. In fact, dropout stability requires dropping all weights associated to a subnetwork (and not just a given fraction of weights). The stronger guarantee on the distance to ideal particles leads to an extra $\log N$ in our bounds (compare Theorem \ref{th:2layerbddmulti} in this paper and (5.1) in \cite{araujo2019mean}). As concerns the proof of connectivity, we generalize the approach of \cite{kuditipudi2019explaining}, in order to analyze the model \eqref{eq:mlayer}.

The bounds in Theorem \ref{th:2layerbddmulti} are not dimension-free (as in the two-layer case), but the dependence on the dimension $d$ is only linear. In fact, the loss change in \eqref{eq:epsbddmulti} vanishes as long as $A_{\rm min}\gg d$, and $\alpha \ll 1/(d+\log N)$. The condition $A_{\rm min}\gg d$ implies that $A_{\rm min}$ needs to scale at least linearly with $d$, but does not scale with $N$. Furthermore, as in the two-layer case, the condition $\alpha \ll 1/(d+\log N)$ implies that the number of samples $k$ needs to scale only logarithmically with $N$. 

Compared to the two-layer case where there is no assumption on the initialization for $\bw_i$, here we require a mild condition (finite first moment for $\rho_0^{(\ell)}$) in order to simplify the proof.

\section{Numerical Results} \label{sec:num}

We consider two supervised learning tasks: \emph{(a)} MNIST classification with the two-layer neural network \eqref{eq:2layer}; and \emph{(b)} CIFAR-10 classification with the three-layer neural network \eqref{eq:multispecial}. For MNIST, the input dimension is $d=28\times 28=784$ and we normalize pixel values to have zero mean and unit variance. For CIFAR-10, the input is given by VGG-16 features of dimension $d=4\times 4\times 512=8192$. These features are computed by the convolutional layers of the VGG-16 network \cite{simonyan2014very} pre-trained on the ImageNet dataset \cite{russakovsky2015imagenet}. More specifically, we rescale the images to size $128 \times 128$, we rescale pixel values into the range $[-1, 1]$, and we feed them to the pre-trained VGG-16 network to extract the features. Qualitatively similar results (with larger classification error) are obtained by using fully connected networks directly on CIFAR-10 images.

\begin{figure*}[ht!]
    \centering
    \subfloat[MNIST, two-layer]{\includegraphics[width=\columnwidth]{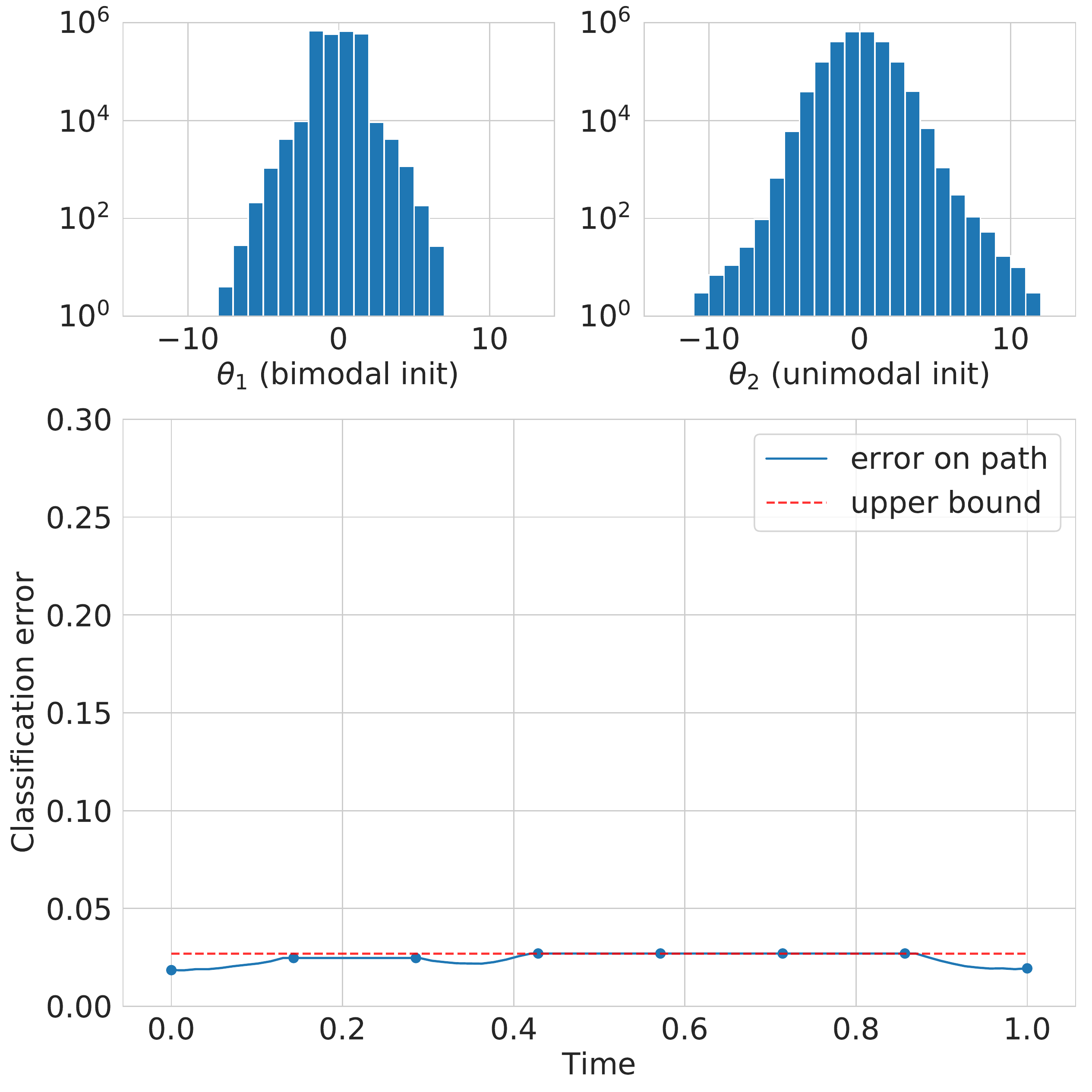}}
    \subfloat[CIFAR-10, three-layer]{\includegraphics[width=0.994\columnwidth]{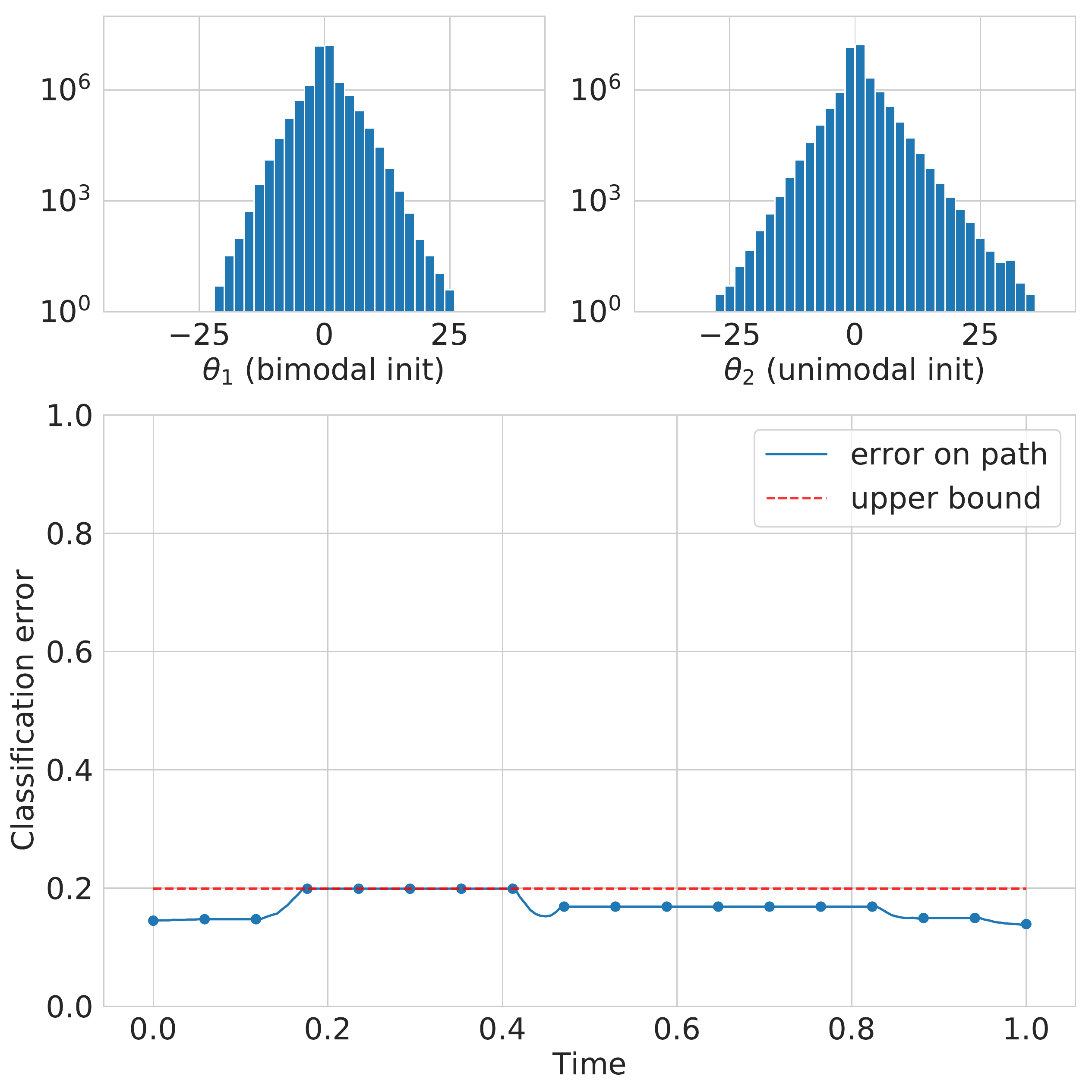}}
\caption{Classification error along a piecewise linear path that connects two SGD solutions $\btheta_1$ and $\btheta_2$, with $N=3200$. As predicted by the theory, the error along the path (blue curve) is no larger than the error of the two SGD solutions plus the change in loss due to the dropout of half of the neurons (red dashed curve).\vspace{-2mm}}\label{fig:3}

\end{figure*}
  
For both tasks, the neural networks have ReLU activation functions, SGD aims at minimizing the cross-entropy loss, and the gradients are averaged over mini-batches of size $100$. In contrast with the setting of Section \ref{sec:mainmul}, all the layers of the neural network are trained. The scaling of the gradient updates follows \eqref{eq:SGD} and \eqref{eq:SGDmulti}: for the first and last layer, the gradient of the loss function is multiplied by a factor of $N$; for the middle layers, the gradient of the loss function is multiplied by a factor of $N^2$. This scaling ensures that the term in front of the learning rate $s_k$ does not depend on $N$, i.e., it is $\Theta(1)$ as $N$ goes large. The learning rate $s_k=\alpha\xi(k\alpha)$ does not depend on the time of the evolution, i.e., $\xi(t)=1$. Furthermore, we set $\alpha=\alpha_0 / N$, where $\alpha_0$ is a constant independent of $N$. We also set the number of training epochs to $k_0\cdot N$, where $k_0$ is a constant independent of $N$. In this way, the product between the learning rate and the number of training epochs is the constant $T=k_0\cdot \alpha_0$, which does not depend on $N$. The initializations of the parameters of the neural network are i.i.d. and do not depend on $N$, as in the setting described for the theoretical results. The population risk and the classification error are obtained by averaging over the test dataset. To measure statistics in the plots, i.e., average value and error bar at 1 standard deviation, we perform 20 independent trials of each experiment.

Figure \ref{fig:1} compares the performance of the trained network (blue dashed curve) and of the dropout network (orange curve), which is obtained by removing the second half of the neurons from each layer (and by suitably rescaling the remaining neurons). On the left, we report the results for MNIST, and on the right for CIFAR-10. For each classification task, we plot the population risk and the classification error for $N=800$ and $N=3200$. The networks are trained until the training loss has reached a plateau ($0.062$ for MNIST and $0.415$ for CIFAR-10 when $N=3200$). As expected, the performance of the dropout network improves with $N$, and it is very close to that of the trained network. For $N=3200$, the classification error of the trained network is $<2\%$ for MNIST and $<14\%$ on CIFAR-10, and the classification error of the dropout network is $\approx3\%$ on MNIST and $<16\%$ on CIFAR-10. 

Figure~\ref{fig:2} plots the change in loss when only half of the neurons remain in the network, as a function of the total number of neurons $N$. For each classification task, we plot the change in loss  at the beginning of training ($0\cdot T$), at an intermediate point where the population risk is still not too small ($\{0.65, 0.7\} \cdot T$), and at the end of training ($1 \cdot T$), where $T$ stands for the product of the learning rate and the total number of training epochs.
The dependence between the change in loss and $N$ is essentially linear in log-log scale, as demonstrated by our theoretical results. Furthermore, the dependence on the time of the dynamics is quite mild. 

Figure \ref{fig:3} shows that the optimization landscape is approximately connected when $N=3200$. We plot the classification error along a piecewise linear path that connects two SGD solutions $\btheta_1$ and $\btheta_2$ initialized with different distributions: the initial distribution of $\btheta_1$ is bimodal, while the initial distribution of $\btheta_2$ is unimodal. We also show the histograms of $\btheta_1$ and $\btheta_2$, in order to highlight that one SGD solution cannot be obtained as a permutation of the other. As expected, the classification error along the path is roughly constant, since the network is dropout stable. More specifically, the error along the path (blue curve) is upper bounded by the error at the extremes plus the change in loss after dropping out half of the neurons of the network (red dashed curve).

\begin{figure}[t!]
  \begin{center}
    \includegraphics[width=0.485\textwidth]{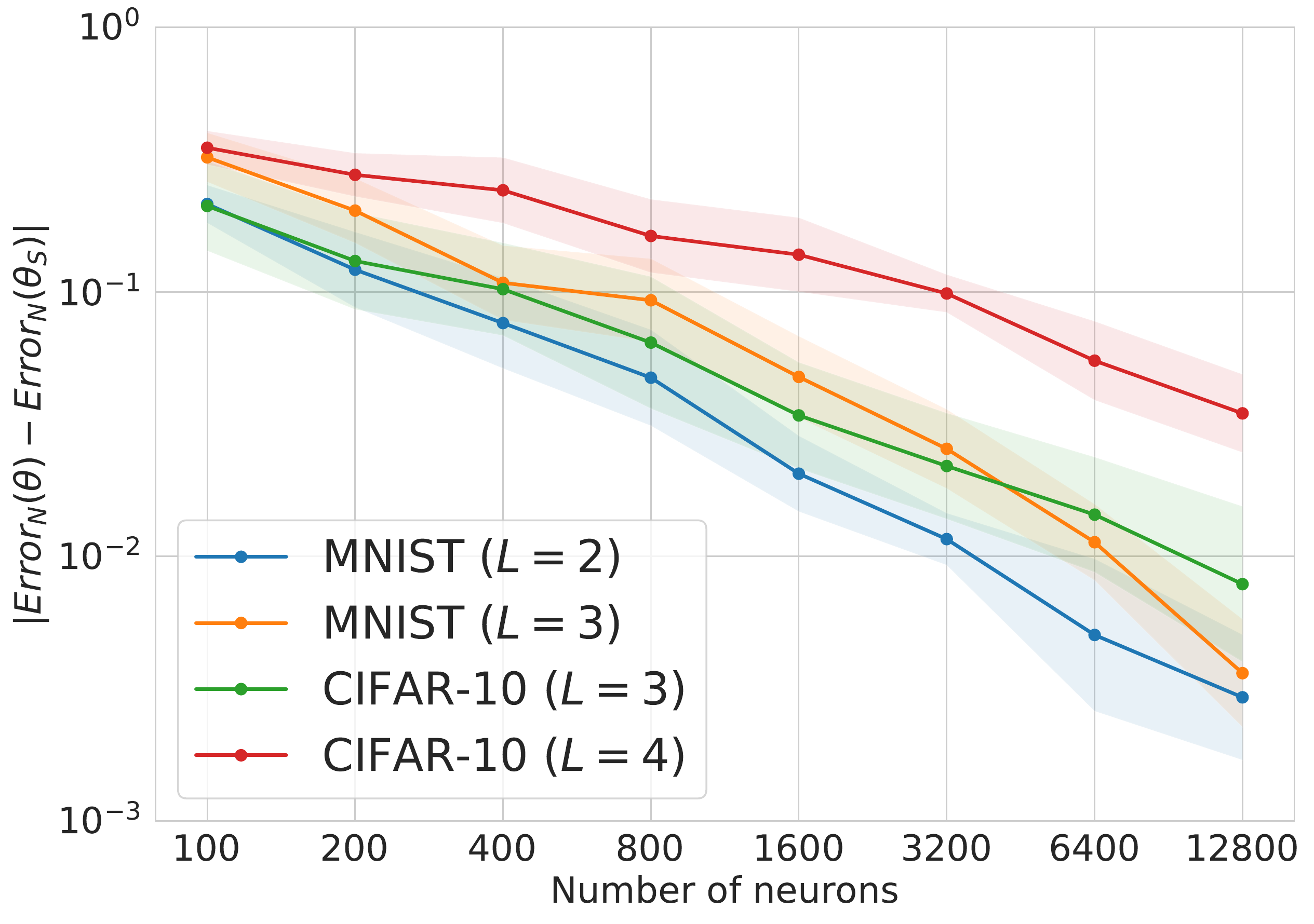}
  \end{center}
  \caption{Change in classification error after removing half of the neurons from each layer, as a function of the number of neurons $N$ of the full network, at the end of training.\vspace{-5mm}}\label{fig:4}\end{figure}

Figure \ref{fig:4} plots the degradation in classification error due to the removal of half of the neurons from each layer. We consider neural networks at the end of training ($1 \cdot T$) and we report the performance degradation as a function of the number of neurons $N$ of the full network. We compare different architectures (two-layer, three-layer and four-layer neural networks) and classification tasks (MNIST and CIFAR-10). In all the cases considered, the performance degradation rapidly decreases, as the width of the network grows. When $N=12800$, the classification error increases only \emph{(i)} by $0.35\%$ for a two-layer network trained on MNIST, \emph{(ii)} by $0.4\%$ for a three-layer network trained on MNIST, \emph{(iii)} by $1\%$ for a three-layer network trained on CIFAR-10, and \emph{(iv)} by $3.6\%$ for a four-layer network trained on CIFAR-10.  

Additional experiments are presented in Appendix \ref{app:addexp} in the supplementary material for the following learning tasks: classification of isotropic Gaussians with the two-layer neural network \eqref{eq:2layer}; MNIST classification with the three-layer neural network \eqref{eq:multispecial}; CIFAR-10 classification with the four-layer neural network \eqref{eq:multispecial}.

\section{Discussion and Future Directions}\label{sec:fin}

The optimization landscape of neural networks can exhibit spurious local minima \cite{yun2018critical, safran2018spurious}, and its minima can be disconnected \cite{freeman2016topology, venturi2018spurious, kuditipudi2019explaining}. In this work, we show that these problematic scenarios are ruled out with SGD training and over-parametrization. In particular, we prove that the optimization landscape of SGD solutions is increasingly connected as the number of neurons grows. The explanation to this phenomenon has been hypothesized by some recent work: the SGD solutions have degrees of freedom to spare \cite{draxler2018essentially} or, equivalently, they are dropout stable \cite{kuditipudi2019explaining}. We give theoretical grounding to this conjecture by proving that SGD solutions are dropout stable, i.e., that the loss does not change much when we remove even a large amount of neurons. In order to have meaningful bounds, the number of neurons does not need to be of the same order of the number of samples (cf. \cite{nguyen2017loss, nguyen2018optimization, nguyen2018loss, nguyen2019connected}). Furthermore, our bounds are dimension-free for two-layer networks and they scale linearly with the dimension for multilayer networks (cf. \cite{freeman2016topology}). Our analysis builds on a recent line of work showing that the dynamics of SGD tends to a mean field limit as the number of neurons increases \cite{mei2018mean, mei2019mean, araujo2019mean}. We believe that with these tools one could prove similar results also for noisy SGD and projected SGD.

The notion of dropout stability is closely related to the fact that neural networks have many redundant connections, and therefore they can be pruned with little performance loss, see, e.g., \cite{guo2016dynamic, molchanov2017variational, frankle2018lottery, liu2018rethinking}. However, it is difficult even to compare the relative merits of the different pruning techniques \cite{gale2019state}, let alone to understand the fundamental reasons leading to sparsity in neural networks. Thus, it would be interesting to investigate whether mean field approaches provide a more principled way of pruning deep neural networks.

\section*{Acknowledgements}
 
M. Mondelli was partially supported by the 2019 Lopez-Loreta Prize. The authors thank Phan-Minh Nguyen for helpful discussions and the IST Distributed Algorithms and Systems Lab for 
providing computational resources. 

\bibliography{example_paper}
\bibliographystyle{icml2020}

\newpage
\onecolumn
\icmltitle{Supplementary Material (Appendix)\\
Landscape Connectivity and Dropout Stability of SGD Solutions for Over-parameterized Neural Networks}
\appendix

\section{Proof of Theorem \ref{th:2layerbdd}}\label{app:th1}

\subsection{Part (A)}\label{app:th1A}

Given $\btheta=(a, \bw)\in\mathbb R^D$, let $\sigma_\star(\bx, \btheta)=a\sigma(\bx, \bw)$. Given $\rho\in \mathscrsfs P(\mathbb{R}^D)$, we define the limit loss as 
\begin{equation}\label{eq:limrisk}
\bar{L}(\rho) = \mathbb{E}\left\{\left(y-\int \sigma_\star(\bx, \btheta)\rho({\rm d}{\boldsymbol{\theta}})\right)^2\right\},
\end{equation}
where the expectation is taken over $(\bx, y)$. For $i\in[N]$ and $t\geq 0$, we consider the following nonlinear dynamics: 
\begin{equation}\label{eq:nd}
   \frac{{\rm d}}{{\rm d}t} \bar{{\boldsymbol{\theta}}}_{i}^{t}=2 \xi(t) \int  \mathbb{E}\left\{\nabla\sigma_\star(\boldsymbol{x},\bar{\boldsymbol{\theta}}_i^t)\,(y-\sigma_\star(\boldsymbol{x},\boldsymbol{\theta}'))\right\}\rho_t({\rm d}\btheta'),
\end{equation}
    where $\nabla$ denotes the gradient with respect to $\bar{{\boldsymbol{\theta}}}_{i}^{t}$ and $\bar{{\boldsymbol{\theta}}}_{i}^{t}\sim \rho_t$. We initialize \eqref{eq:nd} with $\{\bar{{\boldsymbol{\theta}}}_{i}^{0}\}_{i=1}^N\overset{\rm {i.i.d.}}{\sim}\rho_0$.
    
 In \cite{mei2019mean}, it is considered the two-layer neural network \eqref{eq:2layer} with $N$ neurons and bounded activation function $\sigma$, and it is studied the evolution under the SGD algorithm \eqref{eq:SGD} of the parameters $\btheta^k$. In particular, it is shown that, under suitable assumptions, \emph{(i)} the solution of \eqref{eq:nd} exists and it is unique, \emph{(ii)} the $N$ i.i.d. ideal particles $\{\bar{{\boldsymbol{\theta}}}_{i}^{t}\}_{i=1}^N$ are close to the parameters $\btheta^k$ obtained after $k$ steps of SGD with step size $\alpha$, with $t=k\alpha$, and \emph{(iii)} the loss $L_N(\btheta^k)$ concentrates to the limit loss $\bar{L}(\rho_t)$, where $\rho_t$ is the law of $\bar{\btheta}_i^t$.
    
Let us now provide the proof of Theorem \ref{th:2layerbdd}, part (A).

\begin{proof}[Proof of Theorem \ref{th:2layerbdd}, part (A)] Without loss of generality, we can assume that $\btheta^k_{\rm S}$ contains the first $|\mathcal A|$ elements of $\btheta^k$, i.e., $\btheta^k_{\rm S}=(\btheta^k_1, \btheta^k_2, \ldots, \btheta^k_{|\mathcal A|})$. In fact, the subset $\mathcal A$  is independent of the SGD algorithm. Thus, by symmetry, the joint distribution of $\{\btheta_i^k\}_{i\in\mathcal A}$ depends only on $|\mathcal A|$ (and not on the set $\mathcal A$ itself). By Definition \ref{def:dropoutstab}, we need to show that, with probability at least $1-e^{-z^2}$,
\begin{equation}\label{eq:finalbound}
\sup_{k\in [T/\alpha]}|L_N(\boldsymbol{\theta}^k)-L_{|\mathcal A|}(\boldsymbol{\theta}^k_{\rm S})|\le K e^{KT^{3}} \left(\frac{\sqrt{\log |\mathcal A|}+z}{\sqrt{|\mathcal A|}} + \sqrt{\alpha}\big(\sqrt{D+\log N}+z\big)\right).
\end{equation}

Let $\bar{\btheta}^{k\alpha} = (\bar{\btheta}^{k\alpha}_1, \ldots, \bar{\btheta}^{k\alpha}_N)$ be the solution of the nonlinear dynamics \eqref{eq:nd} at time $k\alpha$, with $\bar{{\boldsymbol{\theta}}}_{i}^{k\alpha}\sim \rho_{k\alpha}$. By triangle inequality, we have that
\begin{equation}\label{eq:triangle}
\begin{split}
|L_N(\boldsymbol{\theta}^k)-L_{|\mathcal A|}(\boldsymbol{\theta}^k_{\rm S})| &\le |L_N(\boldsymbol{\theta}^k)-\bar{L}(\rho_{k\alpha})| + |L_{|\mathcal A|}(\boldsymbol{\theta}^k_{\rm S})-\bar{L}(\rho_{k\alpha})|\\
&\le |L_N(\boldsymbol{\theta}^k)-\bar{L}(\rho_{k\alpha})| +|L_{|\mathcal A|}(\boldsymbol{\theta}^k_{\rm S})-L_{|\mathcal A|}(\bar{\boldsymbol{\theta}}^{k\alpha}_{\rm S})|+  |L_{|\mathcal A|}(\bar{\boldsymbol{\theta}}^{k\alpha}_{\rm S})-\bar{L}(\rho_{k\alpha})|,
\end{split}
\end{equation}
where $\bar{L}$ is defined in \eqref{eq:limrisk} and $\bar{\boldsymbol{\theta}}^{{k\alpha}}_{\rm S}=(\bar{\boldsymbol{\theta}}^{{k\alpha}}_1, \bar{\boldsymbol{\theta}}^{{k\alpha}}_2, \ldots, \bar{\boldsymbol{\theta}}^{{k\alpha}}_{|\mathcal A|})$ denotes the vector containing the first $|\mathcal A|$ elements of $\bar{\boldsymbol{\theta}}^{{k\alpha}}$.

Let us consider the first term in the RHS of \eqref{eq:triangle}. Note that, without loss of generality, we can assume that $\alpha\le 1/(C(D+\log N + z^2)e^{CT^3})$, for some constant $C$ depending only on the constants $K_i$ of the assumptions {\bf (A1)}-{\bf (A4)}. Let us explain why this is the case. If $\alpha> 1/(C(D+\log N + z^2)e^{CT^3})$, then the RHS of \eqref{eq:finalbound} is lower bounded by a constant depending only on $K_i$. Furthermore, $y$ and $\sigma$ are bounded, and by Proposition 8 of \cite{mei2019mean}, we have that, with probability at least $1-e^{-z^2}$,
\begin{equation}\label{eq:bdai2}
\begin{split}
\sup_{k \in [T/\alpha]} \max_{i\in [N]}|a_i^k|&\le C_3 (1+T).
\end{split}
\end{equation}
Thus, if $\alpha> 1/(C(D+\log N + z^2)e^{CT^3})$, then the result is trivially true. Consequently, we can apply Theorem 1 of \cite{mei2019mean} and we have that, with probability at least $1-e^{-z^2}$,
\begin{equation}\label{eq:term1}
\sup_{k\in [T/\alpha]}|L_N(\boldsymbol{\theta}^k)-\bar{L}(\rho_{k\alpha})| \le C_1 e^{C_1\, T^{3}} \left(\frac{\sqrt{\log N}+z}{\sqrt{N}} + \sqrt{\alpha}\big(\sqrt{D+\log N}+z\big)\right),
\end{equation}
where $C_1$ depends only on $K_i$. In what follows, the $C_i$ are constants that depend only on $K_i$.

Let us now consider the second term in the RHS of \eqref{eq:triangle}. After some manipulations, we have that
\begin{equation}\label{eq:qq1}
\begin{split}
|L_{|\mathcal A|}(\boldsymbol{\theta}^k_{\rm S})-L_{|\mathcal A|}(\bar{\boldsymbol{\theta}}^{{k\alpha}}_{\rm S})|&\le 2 \max_{i \in \mathcal A}\left|a_i^k\mathbb E\left\{y\sigma(\bx, \bw^k_i)\right\}-\bar{a}_i^{k\alpha}\mathbb E\left\{y\sigma(\bx, \bar{\bw}^{k\alpha}_i)\right\}\right|\\
&\hspace{3em}+\max_{i, j\in\mathcal A}\left|a_i^k a_j^k\mathbb E\{\sigma(\bx, \bw^k_i)\sigma(\bx, \bw^k_j)\}-\bar{a}_i^{k\alpha} \bar{a}_j^{k\alpha}\mathbb E\{\sigma(\bx, \bar{\bw}^{k\alpha}_i)\sigma(\bx, \bar{\bw}^{k\alpha}_j)\}\right|\\
&\le C_2 \left( \max_{i \in \mathcal A} \big(1 + \max(|a_i^k|,  |\bar{a}_i^{{k\alpha}}|)\big)\right)^2 \max_{i \in \mathcal A}\|\boldsymbol{\theta}_i^{k} - \bar{\boldsymbol{\theta}}_i^{{k\alpha}}\|_2\\
&\le C_2 \left( \max_{i \in [N]} \big(1 + \max(|a_i^k|,  |\bar{a}_i^{{k\alpha}}|)\big)\right)^2 \max_{i \in [N]}\|\boldsymbol{\theta}_i^{k} - \bar{\boldsymbol{\theta}}_i^{{k\alpha}}\|_2,\\
\end{split}
\end{equation}
where $\btheta^k_i = (a_i^k, \bw_i^k)$, $\bar{\btheta}^{{k\alpha}}_i = (\bar{a}_i^{{k\alpha}}, \bar{\bw}_i^{{k\alpha}})$, and in the second inequality we use that $y$, $\sigma$ and the gradient of $\sigma$ are bounded. By using Lemma 7 of \cite{mei2019mean}, we have that
\begin{equation}\label{eq:bdai}
\begin{split}
\sup_{t \in [0, T]}\max_{i \in [N]} |\bar{a}_i^{t}|&\le C_3 (1+T).
\end{split}
\end{equation}
Furthermore, by using Propositions 6-7-8 of \cite{mei2019mean}, we have that, with probability at least $1-e^{-z^2}$,
\begin{equation}
\sup_{k \in [T/\alpha]}  \max _{i \in[N]}\|{\boldsymbol{\theta}}_{i}^{k}-\overline{\boldsymbol{\theta}}_{i}^{k\alpha}\|_{2} \leq C_4 e^{C_4T^{3}} \left(\frac{\sqrt{\log N}+z}{\sqrt{N}} + \sqrt{\alpha}\big(\sqrt{D+\log N}+z\big)\right).
\end{equation}
As a result, by combining \eqref{eq:bdai2}, \eqref{eq:bdai} and \eqref{eq:qq1}, we conclude that, with probability at least $1-e^{-z^2}$, 
\begin{equation}\label{eq:term2}
\sup_{k \in [T/\alpha]}|L_{|\mathcal A|}(\boldsymbol{\theta}^k_{\rm S})-L_{|\mathcal A|}(\bar{\boldsymbol{\theta}}^{k\alpha}_{\rm S})|\le C_5 e^{C_5T^{3}} \left(\frac{\sqrt{\log N}+z}{\sqrt{N}} + \sqrt{\alpha}\big(\sqrt{D+\log N}+z\big)\right).
\end{equation}

Finally, let us consider the third term in the RHS of \eqref{eq:triangle}. By triangle inequality, we have that 
\begin{equation}\label{eq:tlast1}
|L_{|\mathcal A|}(\bar{\boldsymbol{\theta}}^{{k\alpha}}_{\rm S})-\bar{L}(\rho_{k\alpha})|\le \left|L_{|\mathcal A|}(\bar{\boldsymbol{\theta}}^{{k\alpha}}_{\rm S})-\mathbb E_{\rho_0}\left\{L_{|\mathcal A|}(\bar{\boldsymbol{\theta}}^{{k\alpha}}_{\rm S})\right\}\right| + \left|\mathbb E_{\rho_0}\left\{L_{|\mathcal A|}(\bar{\boldsymbol{\theta}}^{{k\alpha}}_{\rm S})\right\}-\bar{L}(\rho_{k\alpha})\right|,
\end{equation}
where the notation $\mathbb E_{\rho_0}$ emphasizes that the expectation is taken with respect to $\bar{\btheta}_i^0\sim\rho_0$. Recall that $\bar{L}$ is defined in \eqref{eq:limrisk} and that
\begin{equation}\label{eq:riskhalf}
L_{|\mathcal A|}(\boldsymbol{\theta}_{\rm S}) = \mathbb{E}_{(\bx, y)} \left\{\left(y - \frac{1}{|\mathcal A|}\sum_{i=1}^{|\mathcal A|}\sigma_\star(\bx, \btheta_i)\right)^2\right\},
\end{equation}
where the notation $ \mathbb{E}_{(\bx, y)}$ emphasizes that the expectation is taken with respect to $(\bx, y)\sim \mathbb P$. Furthermore, note that $\{\bar{\boldsymbol{\theta}}^{{k\alpha}}_i\}_{i=1}^{|\mathcal A|}\overset{\rm {i.i.d.}}{\sim}\rho_{k\alpha}$. Thus, after some manipulations, we can rewrite the second term in the RHS of \eqref{eq:tlast1} as 
\begin{equation}
\begin{split}
&\left|L_{|\mathcal A|}(\bar{\boldsymbol{\theta}}^{{k\alpha}}_{\rm S})-\mathbb E_{\rho_0}\left\{L_{|\mathcal A|}(\bar{\boldsymbol{\theta}}^{{k\alpha}}_{\rm S})\right\}\right| \\
&\hspace{2em}=\frac{1}{|\mathcal A|}\left| \int  \mathbb{E}_{(\bx, y)}\left\{\big(\sigma_\star(\bx, \btheta)\big)^2\right\}\rho_{k\alpha}({\rm d}\btheta)-\int  \mathbb{E}_{(\bx, y)}\left\{\sigma_\star(\bx, \btheta_1)\sigma_\star(\bx, \btheta_2)\right\}\rho_{k\alpha}({\rm d}\btheta_1)\rho_{k\alpha}({\rm d}\btheta_2)\right|.
\end{split}
\end{equation}
As $\sigma$ is bounded by assumption \textbf{(A3)} and $\sup_{k \in [T/\alpha]}\max_{i \in [N]}|\bar{a}_i^{{k\alpha}}|$ is bounded by \eqref{eq:bdai}, we deduce that 
\begin{equation}\label{eq:tlast2}
\sup_{k \in [T/\alpha]}\left|L_{|\mathcal A|}(\bar{\boldsymbol{\theta}}^{{k\alpha}}_{\rm S})-\mathbb E_{\rho_0}\left\{L_{|\mathcal A|}(\bar{\boldsymbol{\theta}}^{{k\alpha}}_{\rm S})\right\}\right|\le \frac{C_6\,(1+T)^2}{|\mathcal A|}.
\end{equation}
Let $\btheta$ and $\btheta'$ be two parameters that differ only in one component, i.e., $\btheta=(\btheta_1, \ldots, \btheta_i, \ldots, \btheta_{|\mathcal A|})$ and $\btheta'=(\btheta_1, \ldots, \btheta_i', \ldots, \btheta_{|\mathcal A|})$, and such that $\max_{i \in |\mathcal A|}|a_i|\le C(1+T)$ and $\max_{i \in |\mathcal A|}|a_i'|\le C(1+T)$. Then, 
\begin{equation}
\left|L_{|\mathcal A|}(\btheta)-L_{|\mathcal A|}(\btheta')\right|\le \frac{C_7\,(1+T)^2}{|\mathcal A|}.
\end{equation}
As $\max_{i \in [N]}|\bar{a}_i^{t}|$ is bounded by \eqref{eq:bdai}, by applying McDiarmid's inequality, we obtain that   
\begin{equation}
\mathbb P\left(\left|L_{|\mathcal A|}(\bar{\boldsymbol{\theta}}^{t}_{\rm S})-\mathbb E_{\rho_0}\left\{L_{|\mathcal A|}(\bar{\boldsymbol{\theta}}^{t}_{\rm S})\right\}\right|>\delta\right)\le \exp\left(-\frac{|\mathcal A| \delta^2}{C_8(1+T)^4}\right).
\end{equation}
Furthermore, we have that
\begin{equation}
\begin{split}
\left|L_{|\mathcal A|}(\bar{\boldsymbol{\theta}}^{t}_{\rm S})-L_{|\mathcal A|}(\bar{\boldsymbol{\theta}}^{s}_{\rm S})\right|&\le C_9 \left( \max_{i \in [N]} \big(1 + \max(|\bar{a}_i^t|,  |\bar{a}_i^{s}|)\big)\right)^2 \max_{i \in [N]}\|\bar{\boldsymbol{\theta}}_i^{t} - \bar{\boldsymbol{\theta}}_i^{s}\|_2\\
&\le C_{10}(1+T)^4 |t-s|,
\end{split}
\end{equation}
where in the first inequality we use passages similar to those of \eqref{eq:qq1}, and in the second inequality we use \eqref{eq:bdai} and Lemma 9 of \cite{mei2019mean}. Consequently, 
\begin{equation}
\begin{split}
\left||L_{|\mathcal A|}(\bar{\boldsymbol{\theta}}^{t}_{\rm S})-\mathbb E_{\rho_0}\{L_{|\mathcal A|}(\bar{\boldsymbol{\theta}}^{t}_{\rm S})\}|-|L_{|\mathcal A|}(\bar{\boldsymbol{\theta}}^{s}_{\rm S})-\mathbb E_{\rho_0}\{L_{|\mathcal A|}(\bar{\boldsymbol{\theta}}^{s}_{\rm S})\}|\right|&\le  C_{11}(1+T)^4 |t-s|.
\end{split}
\end{equation}
By taking a union bound over $s\in [T/\nu]$ and bounding the difference between time in the interval grid, we deduce that
\begin{equation}
\mathbb P\left(\sup_{t\in [0, T]}\left|L_{|\mathcal A|}(\bar{\boldsymbol{\theta}}^{t}_{\rm S})-\mathbb E_{\rho_0}\left\{L_{|\mathcal A|}(\bar{\boldsymbol{\theta}}^{t}_{\rm S})\right\}\right|\ge \delta+C_{11}(1+T)^4 \nu\right)\le \frac{T}{\nu}\exp\left(-\frac{|\mathcal A| \delta^2}{C_8(1+T)^4}\right).
\end{equation}
Pick $\nu=1/\sqrt{|\mathcal A|}$ and $\delta = C_8(1+T)^2 (\sqrt{\log(|\mathcal A|T)}+z)/\sqrt{|\mathcal A|}$. Thus, with probability at least $1-e^{-z^2}$, we have that 
\begin{equation}\label{eq:tlast3}
\sup_{k \in [T/\alpha]}\left|L_{|\mathcal A|}(\bar{\boldsymbol{\theta}}^{T}_{\rm S})-\mathbb E_{\rho_0}\left\{L_{|\mathcal A|}(\bar{\boldsymbol{\theta}}^{T}_{\rm S})\right\}\right|\le C_{12}\,(1+T)^3 \frac{\sqrt{\log |\mathcal A|}+z}{\sqrt{|\mathcal A|}}.
\end{equation}
By combining \eqref{eq:tlast2} and \eqref{eq:tlast3}, we conclude that, with probability at least $1-e^{-z^2}$,
\begin{equation}\label{eq:term3}
\sup_{k \in [T/\alpha]}|L_{|\mathcal A|}(\bar{\boldsymbol{\theta}}^{T}_{\rm S})-\bar{L}(\rho_T)| \leq C_{13}\,(1+T)^3 \frac{\sqrt{\log |\mathcal A|}+z}{\sqrt{|\mathcal A|}}.
\end{equation}
Finally, by combining \eqref{eq:triangle}, \eqref{eq:term1}, \eqref{eq:term2} and \eqref{eq:term3}, the result readily follows.
\end{proof}

\subsection{Part (B)}\label{app:th1B}

The proof of part (B) is obtained by combining part (A) with the following lemma. 

\begin{lemma}[Dropout stability implies connectivity -- two-layer]\label{lemma:dc}
Consider a two-layer neural network with $N$ neurons, as in \eqref{eq:2layer}. Given $\mathcal A=[N/2]$, let $\btheta$ and $\btheta'$ be $\varepsilon$-dropout stable as in Definition \ref{def:dropoutstab}. Then, $\btheta$ and $\btheta'$ are $\varepsilon$-connected as in Definition \ref{def:conn}. Furthermore, the path connecting $\btheta$ with $\btheta'$ consists of 7 line segments.
\end{lemma}

\begin{proof}[Proof of Lemma \ref{lemma:dc}]  Let $\btheta=((a_1, \bw_1), (a_2, \bw_2), \ldots, (a_N, \bw_N))$ and $\btheta'=((a_1', \bw_1'), (a_2', \bw_2'), \ldots, (a_N', \bw_N'))$. For the moment, assume that $N$ is even. Consider the piecewise linear path in parameter space that connects $\btheta$ to $\btheta'$ via the following intermediate points:  
\begin{equation}
\begin{split}
\btheta_1&=((2a_1, \bw_1), (2a_2, \bw_2), \ldots, (2a_{N/2}, \bw_{N/2}), (0, \bw_{N/2+1}), (0, \bw_{N/2+2}), \ldots, (0, \bw_N)),\\
\btheta_2&=((2a_1, \bw_1), (2a_2, \bw_2), \ldots, (2a_{N/2}, \bw_{N/2}), (0, \bw_{1}'), (0, \bw_{2}'), \ldots, (0, \bw_{N/2}')),\\
\btheta_3&=((0, \bw_1), (0, \bw_2), \ldots, (0, \bw_{N/2}), (2a_1', \bw_1'), (2a_2', \bw_2'), \ldots, (2a_{N/2}', \bw_{N/2}')),\\
\btheta_4&=((0, \bw_1'), (0, \bw_2'), \ldots, (0, \bw_{N/2}'), (2a_1', \bw_1'), (2a_2', \bw_2'), \ldots, (2a_{N/2}', \bw_{N/2}')),\\
\btheta_5&=((2a_1', \bw_1'), (2a_2', \bw_2'), \ldots, (2a_{N/2}', \bw_{N/2}'), (0, \bw_1'), (0, \bw_2'), \ldots, (0, \bw_{N/2}')),\\
\btheta_6&=((2a_1', \bw_1'), (2a_2', \bw_2'), \ldots, (2a_{N/2}', \bw_{N/2}'), (0, \bw_{N/2+1}'), (0, \bw_{N/2+2}'), \ldots, (0, \bw_N')).
\end{split}
\end{equation}
We will now show that the loss along this path is upper bounded by $\max(L_N(\boldsymbol{\theta}), L_N(\boldsymbol{\theta}'))+\varepsilon$. 

Consider the path that connects $\btheta$ to $\btheta_1$. As $\btheta$ is $\varepsilon$-dropout stable, we have that $L_N(\boldsymbol{\theta}_1) \le L_N(\boldsymbol{\theta})+\varepsilon$. As the loss is convex in the weights of the last layer, the loss along this path is upper bounded by $L_N(\boldsymbol{\theta})+\varepsilon$. Similarly, the loss along the path that connects $\btheta_6$ to $\btheta'$ is upper bounded by $L_N(\boldsymbol{\theta}')+\varepsilon$.

Consider the path that connects $\btheta_1$ to $\btheta_2$. Here, we change $\bw$'s only when the corresponding $a$'s are 0. Thus, the loss does not change along the path. Similarly, the loss does not change along the path that connects $\btheta_3$ to $\btheta_4$ and $\btheta_5$ to $\btheta_6$.

Consider the path that connects $\btheta_2$ to $\btheta_3$. Note that $L_N(\boldsymbol{\theta}_3)=L_N(\boldsymbol{\theta}_5)$. As the loss is convex in the weights of the last layer, the loss along this path is upper bounded by $\max(L_N(\boldsymbol{\theta}), L_N(\boldsymbol{\theta}'))+\varepsilon$.

Finally, consider the path that connects $\btheta_4$ to $\btheta_5$. Here, we are interpolating between two equal subnetworks. Thus, the loss along this path does not change. This concludes the proof for even $N$.

If $N$ is odd, a similar argument can be carried out. The differences are that \emph{(i)} the $\lceil N/2\rceil$-th parameter of $\btheta_1$, $\btheta_2$ and $\btheta_3$ is $(0, \bw_{N/2})$ and  the $\lceil N/2\rceil$-th parameter of $\btheta_4$, $\btheta_5$ and $\btheta_6$ is $(0, \bw_{N/2}')$, and \emph{(ii)} the constant $2$ in front of the $a_i$ is replaced by $N/\lfloor N/2\rfloor$.
\end{proof}

\section{Extension to Unbounded Activation -- Statement and Proof} \label{app:ubb}

We modify the assumptions \textbf{(A2)}, \textbf{(A3)} and \textbf{(A4)} of Section \ref{sec:res2} as follows:

\textbf{(A2')} The feature vectors $\bx$ and the response variables $y$ are bounded by $K_2$, and the gradient $\nabla_{\boldsymbol{w}} \sigma(\boldsymbol{x},\boldsymbol{w})$ is $K_2$ sub-gaussian when $\bx \sim \mathbb P$.

\textbf{(A3')} The activation function $\sigma$ is differentiable, with gradient bounded by $K_3$ and $K_3$-Lipschitz. 

\textbf{(A4')} The initialization $\rho_0$ is supported on $\|\btheta_i^0\|_2\le K_4$.

We are now ready to present our results for unbounded activations in the two-layer setting. 

\begin{theorem}[Two-layer, unbounded activation]\label{th:2layerubdd}
Assume that conditions \textbf{(A1)}, \textbf{(A2')}, \textbf{(A3')} and \textbf{(A4')} hold, and fix $T\ge 1$. Let $\boldsymbol{\theta}^k$ be obtained by running $k$ steps of the SGD algorithm \eqref{eq:SGD} with data $\{(\bx_j, y_j)\}_{j= 0}^k\overset{\rm {i.i.d.}}{\sim} \mathbb P$ and initialization $\rho_0$. Assume further that the loss at each step of SGD is uniformly bounded, i.e., $\max_{j\in \{0, \ldots, k\}} |y_j - \hat{\boldsymbol{y}}_N(\boldsymbol{x}_j, \boldsymbol{\theta}^j)| \leq K_5$. Then, the results of Theorem \ref{th:2layerbdd} hold, with
\begin{equation}\label{eq:parsubb}
\begin{split}
   \varepsilon_{\rm D} &= K(T) \left(  \frac{\sqrt{\log |\mathcal A|}+z}{\sqrt{|\mathcal A|}} + \sqrt{\alpha}\big(\sqrt{D+\log N}+z\big)\right),\\
   \varepsilon_{\rm C} &= K(\max(T, T'))  \left(\frac{\sqrt{\log N}+z}{\sqrt{N}} + \sqrt{\alpha}\big(\sqrt{D+\log N}+z\big)\right).
\end{split}
\end{equation}
where the constant $K(T)$ depends on $T$ and on the constants $K_i$ of the assumptions.
\end{theorem}

To prove the result, we crucially rely on the following bound on the norm of the parameters evolved via SGD.  

\begin{lemma}[Bound on norm of SGD parameters]\label{lemma:normbd}
Under the assumptions of Theorem \ref{th:2layerubdd}, we have that  
\begin{equation}\label{eq:boundnorm}
\sup_{s\in [T/\alpha]}\max_{i\in [N]}\|\boldsymbol{\theta}_i^s\|_2 \leq K e^{K\,T},
\end{equation}
where the constant $K$ depends only on the constants $K_i$ of the assumptions.
\end{lemma}

\begin{proof}[Proof of Lemma \ref{lemma:normbd}]
The SGD update at step $j+1$ gives:
\begin{equation}
\begin{split}
    &a_i^{j+1} = a_i^j +2 \alpha\, \xi(j\alpha)\cdot(y_j - f_N(\boldsymbol{x}_j, \boldsymbol{\theta}^j))\cdot\sigma(\bx_j, \boldsymbol{w}_i^j), \\
    &\boldsymbol{w}_i^{j+1} = \boldsymbol{w}_i^j + 2\alpha\, \xi(j\alpha)\cdot(y_j - f_N(\boldsymbol{x}_j, \boldsymbol{\theta}^j))\cdot a_i^j \nabla_{\bw_i}\sigma(\bx_j, \boldsymbol{w}_i^j).
\end{split}
\end{equation}
We bound the absolute value of the increment $|a_i^{j+1}-a_i^j|$ as
\begin{equation}\label{eq:bounda}
\begin{split}
  |a_i^{j+1} - a_i^{j}| &\leq 2\alpha \xi(j\alpha) \cdot |y_j - f_N(\boldsymbol{x}_j, \boldsymbol{\theta}^j)|\cdot  |\sigma( \bx_j, \boldsymbol{w}_i^j )| \\
  &\stackrel{\mathclap{\mbox{\footnotesize(a)}}}{\le}  \alpha C_1 |\sigma( \bx_j, \boldsymbol{w}_i^j )|  \\
  &\stackrel{\mathclap{\mbox{\footnotesize(b)}}}{\le} \alpha C_2 ( \|\boldsymbol{w}_i^j\|_2 + 1), \\
 \end{split}
\end{equation}
where the constant $C_i$ depends only on $K_i$, in (a) we use that $\xi$ is bounded by $K_1$  and $|y_j - f_N(\boldsymbol{x}_j, \boldsymbol{\theta}^j)|\le K_5$, in (b) we use that $\|\sigma\|_{\rm Lip}\le K_2$ and $\|\bx_j\|_2\le K_2$. Similarly, we bound the absolute value of the increments $\|\boldsymbol{w}_i^{j+1} - \boldsymbol{w}_i^{j}\|_2$ as
\begin{equation}\label{eq:boundw}
\|\boldsymbol{w}_i^{j+1} - \boldsymbol{w}_i^{j}\|_2 \leq \alpha C_3 |a_i^j|.
\end{equation}
By combining \eqref{eq:bounda} and \eqref{eq:boundw}, we get
\begin{equation}\label{eq:c1}
\begin{split}
    \|\boldsymbol{\theta}_i^{j+1} - \boldsymbol{\theta}_i^{j}\|_2 &\leq \|\boldsymbol{w}_i^{j+1} - \boldsymbol{w}_i^{j}\|_2 + |a_i^{j+1} - a_i^{j}| \leq
    \alpha C_4 ( \|\boldsymbol{\theta}_i^j\|_2+1).
\end{split}
\end{equation}
By triangle inequality, we also obtain that
\begin{equation}\label{eq:c2}
\|\boldsymbol{\theta}_i^s\|_2   \leq \sum\limits_{j = 0}^{s-1}  \|\boldsymbol{\theta}_i^{j+1} - \boldsymbol{\theta}_i^{j}\|_2 + \|\boldsymbol{\theta}_i^{0}\|_2.
\end{equation}
As $ \|\boldsymbol{\theta}_i^{0}\|_2$ is bounded, by combining \eqref{eq:c1} and \eqref{eq:c2}, we have that
\begin{equation}
    \|\boldsymbol{\theta}_i^s\|_2  \leq C_5+C_5\,s\alpha+ C_5 \alpha \sum\limits_{j=0}^{s-1} \|\boldsymbol{\theta}_i^j\|_2.
\end{equation}
By using a discrete version of Gronwall's inequality, the result follows.
\end{proof}

Finally, let us present the proof of Theorem \ref{th:2layerubdd}.

\begin{proof}[Proof of Theorem \ref{th:2layerubdd}]
Since the activation function $\sigma$ satisfies assumption \textbf{(A3')}, we can construct $\tilde{\sigma}:\mathbb R^d\times\mathbb R^{D-1}\to\mathbb R$ that satisfies the following two properties:
\begin{description}
\item[(i)] $\tilde{\sigma}(\bx, \bw)$ coincides with $\sigma(\bx,\bw)$ for $\|\bx\|_2\le K_2$ and $\|\bw\|_2 \le Ke^{K\,T}$, where $K_2$ is the constant of assumption \textbf{(A2')} and $Ke^{K\,T}$ is the bound of Lemma \ref{lemma:normbd};
\item[(ii)] $\tilde{\sigma}(\bx, \bw)$ is bounded, differentiable, with bounded and Lipschitz continuous gradient.
\end{description}

Recall that $\boldsymbol{\theta}^k$ is obtained by running $k$ steps of the SGD algorithm \eqref{eq:SGD} with initial condition $\boldsymbol{\theta}^0$, data $\{\bx_j, y_j\}_{j=0}^k$ and activation function $\sigma$. Let $\tilde{\boldsymbol{\theta}}^k$ be obtained by running $k$ steps of SGD with initial condition $\boldsymbol{\theta}^0$, data $\{\bx_j, y_j\}_{j=0}^k$ and activation function $\tilde{\sigma}$. By combining Lemma \ref{lemma:normbd}, assumption \textbf{(A2')} and property \textbf{(i)} of $\tilde{\sigma}$, we immediately deduce that
\begin{equation}\label{eq:same1}
\boldsymbol{\theta}^k=\tilde{\boldsymbol{\theta}}^k.
\end{equation}
Furthermore, we have that
\begin{equation}\label{eq:same2}
\mathbb{E} \left\{\Big(y - \frac{1}{N}\sum\limits_{i=1}^N  a_i \sigma( \boldsymbol{x}, \boldsymbol{w}_i))\Big)^2\right\}=\mathbb{E} \left\{\Big(y - \frac{1}{N}\sum\limits_{i=1}^N  a_i \tilde{\sigma}( \boldsymbol{x}, \boldsymbol{w}_i))\Big)^2\right\},
\end{equation}
namely the loss of $\boldsymbol{\theta}^k$ computed with respect to the activation function $\sigma$ is the same as the loss of $\boldsymbol{\theta}^k$ computed with respect to the activation function $\tilde{\sigma}$.

Note that $\|\tilde{\sigma}\|_{\infty}\le C_1(T)$ for some $C_1(T)$ that depends on $T$ and on the constants $K_i$ of the assumptions. Thus, $\tilde{\sigma}$ satisfies assumptions \textbf{(A2)}  and \textbf{(A3)}, with $K_3$ depending on time $T$ of the evolution. Consequently, by Theorem \ref{th:2layerbdd}, with probability at least $1-e^{-z^2}$, for all $k\in [T/\alpha]$, $\tilde{\boldsymbol{\theta}}^k$ is $\varepsilon_{\rm D}$-dropout stable, with 
\begin{equation}\label{eq:epsbddnew}
 \varepsilon_{\rm D} =K(T) \left(\sqrt{\frac{\log N}{N}} +  \frac{\sqrt{\log |\mathcal A|}+z}{\sqrt{|\mathcal A|}} + \sqrt{\alpha}\big(\sqrt{D+\log N}+z\big)\right).
\end{equation}
By using \eqref{eq:same1} and \eqref{eq:same2}, we conclude that, with probability at least $1-e^{-z^2}$, for all $k\in [T/\alpha]$, $\boldsymbol{\theta}^k$ is $\varepsilon_{\rm D}$-dropout stable. Similarly, with probability at least $1-e^{-z^2}$, for all $k'\in [T'/\alpha]$, $(\boldsymbol{\theta}')^{k'}$ is $\varepsilon_{\rm D}$-dropout stable. Thus, by Lemma \ref{lemma:dc}, the proof is complete. 
\end{proof}

\section{Proof of Theorem \ref{th:2layerbddmulti}}\label{app:thmulti}

\subsection{Part (A)}\label{app:thmultiA}

Let $D=\sum_{i=0}^L D_i$ and let $\rho$ be a probability measure over $\mathbb R^D\cong \mathbb R^{D_0}\times\mathbb R^{D_1}\times\cdots \times\mathbb R^{D_L}$. For $i\in \{0, \ldots, L\}$, we denote by $\rho^{(i)}$ the marginal of $\rho$ over the $i$-th factor $\mathbb R^{D_i}$ of the Cartesian product. For $i\in \{0, \ldots, L-1\}$, we denote by $\rho^{(i, i+1)}$ the marginal of $\rho$ over the $i$-th and the $i+1$-th factors. Furthermore, we denote by $\rho^{(i\mid i+1)}(\cdot\mid \btheta^{(i+1)})$ the conditional distribution of the $i$-th factor given that the $i+1$-th factor is equal to $\btheta^{(i+1)}$. 

Given a feature vector $\bx\in\mathbb R^{d_0}$ and a probability measure $\rho$ over $\mathbb R^D$, we define
\begin{equation}
\begin{split}
\bar{\bz}^{(2)}\left(\boldsymbol{x}, \rho\right)&=\int\sigma^{(1)}\left(\sigma^{(0)}\left(\boldsymbol{x}, \btheta^{(0)}\right), \btheta^{(1)}\right)\,{\rm d}\rho^{(0, 1)}(\btheta^{(0)}, \btheta^{(1)}),\\
\bar{\bz}^{(\ell)}\left(\boldsymbol{x}, \rho\right)&=\int \sigma^{(\ell-1)}\left(\bar{\bz}^{(\ell-1)}\left(\boldsymbol{x}, \rho\right), \btheta^{(\ell-1)}\right)\,{\rm d}\rho^{(\ell-1)}(\btheta^{(\ell-1)}),\qquad \ell\in\{3, \ldots, L-1\},\\
\bar{\bz}^{(L)}\left(\boldsymbol{x}, \rho, \btheta^{(L)}\right)&=\int \sigma^{(L-1)}\left(\bar{\bz}^{(L-1)}\left(\boldsymbol{x}, \rho\right), \btheta^{(L-1)}\right)\,{\rm d}\rho^{(L-1\mid L)}(\btheta^{(L-1)}\mid \btheta^{(L)}),\\
\bar{\by}\left(\boldsymbol{x}, \rho\right)&=\sigma^{(L+1)}\left(\int \sigma^{(L)}\left(\bar{\bz}^{(L)}\left(\boldsymbol{x}, \rho, \btheta^{(L)}\right), \btheta^{(L)}\right)\,{\rm d}\rho^{(L)}(\btheta^{(L)})\right),
\end{split}
\end{equation}
where $\sigma^{(\ell)}:  \mathbb{R}^{d_{\ell}} \times \mathbb{R}^{D_{\ell}} \rightarrow \mathbb{R}^{d_{\ell+1}}$, with $\ell \in\{0, \ldots, L\}$, and $\sigma^{(L+1)}:\mathbb R^{d_{L+1}}\to \mathbb R^{d_{L+1}}$. We remark that $\bar{\bz}^{(L)}$ is defined in terms of the conditional distribution $\rho^{(L-1\mid L)}$. We also define the limit loss as
\begin{equation}\label{eq:lriskm}
\bar{L}(\rho)=\mathbb E\left\{\left\|\by-\bar{\by}\left(\boldsymbol{x}, \rho\right)\right\|_2^2\right\},
\end{equation}
where the expectation is taken over $(\bx, \by)$. Given a probability measure $\rho_0$ over $\mathbb R^D$ and activation functions $\sigma^{(\ell)}$ ($\ell \in\{0, \ldots, L+1\}$), we denote by $\rho^\star_{[0, T]}$ the probability measure over $\mathcal C([0, T], \mathbb R^D)$ which solves the McKean-Vlasov DNN problem with initial condition $\rho_0$, according to Definition 4.4 of \cite{araujo2019mean}. We also denote by $\rho_t^\star$ the marginal of $\rho^\star_{[0, T]}$ at time $t\in[0, T]$.

In \cite{araujo2019mean}, it is considered a model of neural network with $L+1\ge 4$ layers, where each hidden layer contains $N$ neurons. This model can be obtained from \eqref{eq:mlayer} by setting to one the parameters $\{\ba_{i_\ell, i_{\ell+1}}^{\ell}\}_{\ell\in [L-1], i_\ell, i_{\ell+1}\in [N]}$ and $\{\ba_{i_L}^{(L)}\}_{i_L\in [N]}$, and by applying the bounded activation function $\sigma^{(L+1)}$ to the output $\widehat{\boldsymbol{y}}_{N}$. Then, it is studied the evolution under the SGD algorithm \eqref{eq:SGDmulti} of the parameters $\btheta(k)$ of this multilayer neural network. In particular, it is shown that, under suitable assumptions, \emph{(i)} the solution of the McKean-Vlasov DNN problem exists and it is unique, \emph{(ii)} the parameters $\btheta(k)$ obtained after $k$ steps of SGD with step size $\alpha$ are close to particles $\bar{\btheta}(t)$ at time $t=k\alpha$, whose trajectories are distributed according to $\rho_t^\star$, and \emph{(iii)} the loss $L_N(\btheta(k))$ concentrates to the limit loss $\bar{L}(\rho_t^\star)$.

In order to prove Theorem \ref{th:2layerbddmulti}, we will use the following bound on the norm of the parameters $\{\ba_{i_\ell, i_{\ell+1}}^{(\ell)}\}_{\ell\in [L-1], i_\ell, i_{\ell+1}\in [N]}$ evolved via SGD.  

\begin{lemma}[Bound on norm of $\ba_{i_\ell, i_{\ell+1}}^{(\ell)}$]\label{lemma:normbdmulti}
Under the assumptions of Theorem \ref{th:2layerbddmulti}, we have that  
\begin{equation}\label{eq:boundnormm}
\max_{\ell\in [L-1]}\sup_{s\in [T/\alpha]}\max_{i_\ell, i_{\ell+1}\in [N]}\|\ba_{i_\ell, i_{\ell+1}}^{(\ell)}(s)\|_2 \leq K(T, L),
\end{equation}
where the constant $K$ depends only on $T$, $L$ and on the constants $K_i$ of the assumptions.
\end{lemma}

\begin{proof}

For $\ell\in [L-1]$, the SGD update at step $j+1$ gives:
\begin{equation}
\boldsymbol{a}^{(\ell)}_{i_{\ell}, i_{\ell + 1}}(j+1) = \boldsymbol{a}^{(\ell)}_{i_{\ell}, i_{\ell + 1}}(j) + 2\alpha\xi(j\alpha) N^2  \left(\boldsymbol{y}_j-\widehat{\boldsymbol{y}}_{N}\left(\boldsymbol{x}_j, \boldsymbol{\theta}(j)\right)\right)^\sT
{\rm D}_{\boldsymbol{a}^{(\ell)}_{i_{\ell},i_{\ell + 1}}} \widehat{\boldsymbol{y}}_{N}\left(\boldsymbol{x}, \boldsymbol{\theta}(j)\right),
\end{equation}
where ${\rm D}_{\boldsymbol{\theta}^{(\ell)}_{i_{\ell},i_{\ell + 1}}} \widehat{\boldsymbol{y}}_{N}\in \mathbb R^{d_{L+1}}\times \mathbb R^{D_\ell+d_{\ell+1}}$ denotes the Jacobian of $ \widehat{\boldsymbol{y}}_{N}$ with respect to $\boldsymbol{\theta}^{(\ell)}_{i_{\ell},i_{\ell + 1}}$.

Recall that by assumptions \textbf{(B2)-(B3)} the response variables $\boldsymbol{y}_j$ and the activation $\sigma^{(L)}$ are bounded. Moreover, as the final layer of the network is not trained, i.e., $\ba_{i_{L}}^{(L)}(k+1) = \ba_{i_{L}}^{(L)}(k)$ for any $k$, and $\ba_{i_{L}}^{(L)}(0)$ is initialized with a distribution supported on $\|\ba_{i_{L}}^{(L)}(0)\|_2 \leq K_4$, we get that $\ba_{i_{L}}^{(L)}$ is bounded along the whole SGD trajectory. Thus, we are able to conclude that
$
\left\|\boldsymbol{y}_j-\widehat{\boldsymbol{y}}_{N}\left(\boldsymbol{x}_j, \boldsymbol{\theta}(j)\right)\right\|_2 \leq K_5,
$
for some constant $K_5$.

We bound the absolute value of the increment $\|\boldsymbol{a}^{(\ell)}_{i_{\ell}, i_{\ell + 1}}(j+1) - \boldsymbol{a}^{(\ell)}_{i_{\ell}, i_{\ell + 1}}(j)\|_2$ as
\begin{equation}
\begin{split}
\|\boldsymbol{a}^{(\ell)}_{i_{\ell}, i_{\ell + 1}}(j+1) - \boldsymbol{a}^{(\ell)}_{i_{\ell}, i_{\ell + 1}}(j)\|_2&\le 2\,\alpha\,\xi(j\alpha)\, N^2\left\|\boldsymbol{y}_j-\widehat{\boldsymbol{y}}_{N}\left(\boldsymbol{x}_j, \boldsymbol{\theta}(j)\right)\right\|_2\cdot \left\|{\rm D}_{\boldsymbol{a}^{(\ell)}_{i_{\ell},i_{\ell + 1}}} \widehat{\boldsymbol{y}}_{N}\left(\boldsymbol{x}, \boldsymbol{\theta}(j)\right)\right\|_{\rm op}\\
&\le \alpha\, N^2\,C_1\left\|{\rm D}_{\boldsymbol{a}^{(\ell)}_{i_{\ell},i_{\ell + 1}}} \widehat{\boldsymbol{y}}_{N}\left(\boldsymbol{x}, \boldsymbol{\theta}(j)\right)\right\|_{\rm op},
\end{split}
\end{equation}
where we use that $\xi$ is bounded by $K_1$ and $\left\|\boldsymbol{y}_j-\widehat{\boldsymbol{y}}_{N}\left(\boldsymbol{x}_j, \boldsymbol{\theta}(j)\right)\right\|_2\le K_5$. Consequently,
\begin{equation}\label{eq:bdabs}
\max_{i_\ell, i_{\ell+1}\in [N]}\|\boldsymbol{a}^{(\ell)}_{i_{\ell}, i_{\ell + 1}}(j+1) - \boldsymbol{a}^{(\ell)}_{i_{\ell}, i_{\ell + 1}}(j)\|_2 \le \alpha\, N^2\,C_1\max_{i_\ell, i_{\ell+1}\in [N]}\left\|{\rm D}_{\boldsymbol{a}^{(\ell)}_{i_{\ell},i_{\ell + 1}}} \widehat{\boldsymbol{y}}_{N}\left(\boldsymbol{x}, \boldsymbol{\theta}(j)\right)\right\|_{\rm op}.
\end{equation}

Let us now focus on the operator norm of the Jacobian. First, we write
\begin{equation}\label{eq:bdJ0}
\begin{split}
\left\|{\rm D}_{\boldsymbol{a}^{(\ell)}_{i_{\ell},i_{\ell + 1}}} \widehat{\boldsymbol{y}}_{N}\left(\boldsymbol{x}, \boldsymbol{\theta}(j)\right)\right\|_{\rm op}&= \left\|{\rm D}_{\boldsymbol{z}^{(\ell+1)}_{i_{\ell + 1}}} \widehat{\boldsymbol{y}}_{N}\left(\boldsymbol{x}, \boldsymbol{\theta}(j)\right)\cdot {\rm D}_{\boldsymbol{a}^{(\ell)}_{i_{\ell},i_{\ell + 1}}} \boldsymbol{z}^{(\ell+1)}_{i_{\ell + 1}}\left(\boldsymbol{x}, \boldsymbol{\theta}(j)\right)\right\|_{\rm op}\\
&\le\left\|{\rm D}_{\boldsymbol{z}^{(\ell+1)}_{i_{\ell + 1}}} \widehat{\boldsymbol{y}}_{N}\left(\boldsymbol{x}, \boldsymbol{\theta}(j)\right)\right\|_{\rm op}\cdot \left\| {\rm D}_{\boldsymbol{a}^{(\ell)}_{i_{\ell},i_{\ell + 1}}} \boldsymbol{z}^{(\ell+1)}_{i_{\ell + 1}}\left(\boldsymbol{x}, \boldsymbol{\theta}(j)\right)\right\|_{\rm op},
\end{split}
\end{equation}
where the inequality uses the fact that the operator norm is sub-multiplicative. Note that
\begin{equation}
{\rm D}_{\boldsymbol{a}^{(\ell)}_{i_{\ell},i_{\ell + 1}}} \boldsymbol{z}^{(\ell+1)}_{i_{\ell + 1}}\left(\boldsymbol{x}, \boldsymbol{\theta}(j)\right) = {\rm diag}\left(\frac{1}{N}\sigma^{(\ell)}\left(\bz_{i_{\ell}}^{(\ell)}\left(\boldsymbol{x}, \boldsymbol{\theta}\right), \boldsymbol{w}_{i_{\ell}, i_{\ell+1}}^{(\ell)}(j)\right)\right),
\end{equation}
where we denote by ${\rm diag}(\bv)$ the diagonal matrix containing $\bv$ on the diagonal. As $\sigma^{(\ell)}$ is bounded by assumption \textbf{(B3)}, we have that
\begin{equation}\label{eq:bdJ1}
\left\| {\rm D}_{\boldsymbol{a}^{(\ell)}_{i_{\ell},i_{\ell + 1}}} \boldsymbol{z}^{(\ell+1)}_{i_{\ell + 1}}\left(\boldsymbol{x}, \boldsymbol{\theta}(j)\right)\right\|_{\rm op}\le \frac{C_2}{N}.
\end{equation}
Furthermore, the Jacobian ${\rm D}_{\boldsymbol{z}^{(\ell+1)}_{i_{\ell + 1}}} \widehat{\boldsymbol{y}}_{N}\left(\boldsymbol{x}, \boldsymbol{\theta}(j)\right)$ is given by 
\begin{equation}
\begin{split}
{\rm D}_{\boldsymbol{z}^{(L)}_{i_{L}}} \widehat{\boldsymbol{y}}_{N}\left(\boldsymbol{x}, \boldsymbol{\theta}(j)\right) &= \frac{1}{N}\bM^{(L)}_{i_{L}}(\bx, \btheta(j)), \qquad i_L\in [N],\\
{\rm D}_{\boldsymbol{z}^{(\ell)}_{i_{\ell}}} \widehat{\boldsymbol{y}}_{N}\left(\boldsymbol{x}, \boldsymbol{\theta}(j)\right) &= \frac{1}{N^{L-\ell+1}}\sum_{\bp_{\ell+1}^L\in [N]^{L-\ell}}\bM^{(\ell)}_{i_{\ell},\bp_{\ell+1}^L}(\bx, \btheta(j)),\qquad \ell\in [L-1],i_{\ell}\in [N],\\
\end{split}
\end{equation}
where $\bp_{\ell+1}^L$ denotes the multi-index $(p_{\ell+1}, \ldots, p_L)$, $[N]^{L-\ell}$ denotes the $(L-\ell)$-fold Cartesian product of $[N]$ and the matrices $\bM^{(\ell)}_{\bp_{\ell}^L}(\bx, \btheta(j))$ are defined recursively as 
\begin{equation}
\begin{split}
\bM^{(L)}_{p_{L}}(\bx, \btheta(j)) &={\rm D}_{\bz_{p_L}^{(L)}}\left(\ba_{p_L}^{(L)}(j)\odot\sigma^{(L)}\left(\bz_{p_L}^{(L)}(\bx, \btheta(j)), \bw_{p_L}^{(L)}(j)\right)\right)\\
&={\rm diag} (\ba_{p_L}^{(L)}(j))\cdot  {\rm D}_{\bz_{p_L}^{(L)}}\sigma^{(L)}\left(\bz_{p_L}^{(L)}(\bx, \btheta(j)), \bw_{p_L}^{(L)}(j)\right),\\
\bM^{(\ell)}_{\bp_{\ell}^L}(\bx, \btheta(j)) &= \bM^{(\ell+1)}_{\bp_{\ell+1}^L}(\bx, \btheta(j))\cdot {\rm D}_{\bz_{p_\ell}^{(\ell)}}\left(\ba_{p_\ell, p_{\ell+1}}^{(\ell)}(j)\odot\sigma^{(\ell)}\left(\bz_{p_\ell}^{(\ell)}(\bx, \btheta(j)), \bw_{p_\ell, p_{\ell+1}}^{(\ell)}(j)\right)\right)\\
&= \bM^{(\ell+1)}_{\bp_{\ell+1}^L}(\bx, \btheta(j))\cdot {\rm diag}(\ba_{p_\ell, p_{\ell+1}}^{(\ell)}(j))\cdot {\rm D}_{\bz_{p_\ell}^{(\ell)}}\sigma^{(\ell)}\left(\bz_{p_\ell}^{(\ell)}(\bx, \btheta(j)), \bw_{p_\ell, p_{\ell+1}}^{(\ell)}(j)\right).
\end{split}
\end{equation}
Note that $\ba_{p_L}^{(L)}(j)=\ba_{p_L}^{(L)}(0)$ and recall that $\|\ba_{p_L}^{(L)}(0)\|_2$ is bounded by assumption \textbf{(B4)}. Furthermore, $\sigma^{(\ell)}$ has bounded Fr\'echet derivative by assumption \textbf{(B3)}. Thus, we deduce that
\begin{equation}
\left\|\bM^{(L)}_{p_{L}}(\bx, \btheta(j))\right\|_{\rm op}\le C_3,
\end{equation}
and
\begin{equation}
\begin{split}
\left\|\bM^{(\ell)}_{\bp_{\ell}^L}(\bx, \btheta(j))\right\|_{\rm op} &\le C_4(L) \prod_{m=\ell}^{L-1} \|\ba_{p_{m}, p_{m+1}}^{(m)}(j)\|_2\\
&\le C_4(L) \prod_{m=\ell}^{L-1} \max_{i_m, i_{m+1}\in [N]}\|\ba_{i_{m}, i_{m+1}}^{(m)}(j)\|_2.
\end{split}
\end{equation}
Consequently, we have that
\begin{equation}\label{eq:bdJ2}
\begin{split}
\left\|{\rm D}_{\boldsymbol{z}^{(L)}_{i_{L}}} \widehat{\boldsymbol{y}}_{N}\left(\boldsymbol{x}, \boldsymbol{\theta}(j)\right)\right\|_{\rm op}&\le \frac{C_3}{N},\\
\left\|{\rm D}_{\boldsymbol{z}^{(\ell)}_{i_{\ell}}} \widehat{\boldsymbol{y}}_{N}\left(\boldsymbol{x}, \boldsymbol{\theta}(j)\right)\right\|_{\rm op}&\le \frac{C_4(L)}{N} \prod_{m=\ell}^{L-1} \max_{i_m, i_{m+1}\in [N]}\|\ba_{i_{m}, i_{m+1}}^{(m)}(j)\|_2.
\end{split}
\end{equation}
By combining \eqref{eq:bdJ0}, \eqref{eq:bdJ1} and \eqref{eq:bdJ2}, we obtain that
\begin{equation}
\begin{split}
\left\|{\rm D}_{\boldsymbol{a}^{(L-1)}_{i_{L-1},i_{L}}} \widehat{\boldsymbol{y}}_{N}\left(\boldsymbol{x}, \boldsymbol{\theta}(j)\right)\right\|_{\rm op}&\le \frac{C_5}{N^2},\label{eq:bdfinJ1}\\
\left\|{\rm D}_{\boldsymbol{a}^{(\ell)}_{i_{\ell},i_{\ell + 1}}} \widehat{\boldsymbol{y}}_{N}\left(\boldsymbol{x}, \boldsymbol{\theta}(j)\right)\right\|_{\rm op}&\le \frac{C_6(L)}{N^2} \prod_{m=\ell+1}^{L-1} \max_{i_m, i_{m+1}\in [N]}\|\ba_{i_{m}, i_{m+1}}^{(m)}(j)\|_2, \qquad \ell\in [L-2].
\end{split}
\end{equation}
By using also \eqref{eq:bdabs}, we have that
\begin{equation}\label{eq:bdabs2}
\begin{split}
\max_{i_{L-1}, i_{L}\in [N]}&\|\boldsymbol{a}^{(L-1)}_{i_{L-1}, i_{L}}(j+1) - \boldsymbol{a}^{(L-1)}_{i_{L-1}, i_{L}}(j)\|_2\le \alpha\, C_7,\\
\max_{i_\ell, i_{\ell+1}\in [N]}\|\boldsymbol{a}^{(\ell)}_{i_{\ell}, i_{\ell + 1}}(j+1) - \boldsymbol{a}^{(\ell)}_{i_{\ell}, i_{\ell + 1}}(j)\|_2& \le \alpha \, C_8(L)\prod_{m=\ell+1}^{L-1} \max_{i_m, i_{m+1}\in [N]}\|\ba_{i_{m}, i_{m+1}}^{(m)}(j)\|_2, \qquad \ell\in [L-2].
\end{split}
\end{equation}
By triangle inequality, we also obtain that, for $\ell\in [L-1]$ and $i_\ell, i_{\ell+1}\in [N]$,
\begin{equation}\label{eq:tf1}
\|\boldsymbol{a}^{(\ell)}_{i_{\ell}, i_{\ell + 1}}(s)\|_2\le \sum_{j=0}^{s-1}\|\boldsymbol{a}^{(\ell)}_{i_{\ell}, i_{\ell + 1}}(j+1) - \boldsymbol{a}^{(\ell)}_{i_{\ell}, i_{\ell + 1}}(j)\|_2 + \|\boldsymbol{a}^{(\ell)}_{i_{\ell}, i_{\ell + 1}}(0)\|_2.
\end{equation}
As $\|\ba_{i_\ell, i_{\ell+1}}^{(\ell)}(0)\|_2$ and $\|\ba_{i_L}^{(L)}(0)\|_2$ are bounded, by combining \eqref{eq:bdabs2} and \eqref{eq:tf1}, we have that
\begin{equation}
\begin{split}
\max_{s\in [k]}\max_{i_{L-1}, i_{L}\in [N]}\|\boldsymbol{a}^{(L-1)}_{i_{L-1}, i_{L}}(s)\|_2&\le C + C_7 \,T,\\
\max_{s\in [k]}\max_{i_\ell, i_{\ell+1}\in [N]}\|\boldsymbol{a}^{(\ell)}_{i_{\ell}, i_{\ell + 1}}(s)\|_2&\le C + C_8(L)\, T\prod_{m=\ell+1}^{L-1} \max_{s\in [k]}\max_{i_m, i_{m+1}\in [N]}\|\ba_{i_{m}, i_{m+1}}^{(m)}(s)\|_2, \qquad \ell\in [L-2].
\end{split}
\end{equation}
where we have used that $T=k\alpha$. By doing a step of induction on $\ell \in \{ L-2, L-3, \ldots, 1\}$, the proof is complete. 
\end{proof}

We are now ready to provide the proof of Theorem \ref{th:2layerbddmulti}, part (A).

\begin{proof}[Proof of Theorem \ref{th:2layerbddmulti}, part (A)] For $\ell\in [L]$, we construct $\tilde{\sigma}^{(\ell)}:\mathbb R^{d_{\ell}}\times \mathbb R^{D_\ell+d_{\ell+1}}\to \mathbb R^{d_{\ell+1}}$ that satisfies the following two properties:
\begin{description}
\item[(i)] $\tilde{\sigma}^{(\ell)}(\bz, (\bw, \ba))$ coincides with $\ba\odot \sigma^{(\ell)}(\bz, \bw)$ for all $(\bz, \bw)\in \mathbb R^{d_{\ell}}\times \mathbb R^{D_\ell}$ and for $\|\ba\|_2\le K(T, L)$, where $K(T, L)$ is the bound of Lemma \ref{lemma:normbdmulti};
\item[(ii)] $\tilde{\sigma}^{(\ell)}$ is bounded, with Fr\'echet derivatives bounded and Lipschitz.
\end{description} 
Similarly, we construct $\tilde{\sigma}^{(L+1)}:\mathbb R^{d_{L+1}}\to \mathbb R^{d_{L+1}}$ that satisfies the following two properties:
\begin{description}
\item[(i)] $\tilde{\sigma}^{(L+1)}(\bz)=\bz$ for $\|\bz\|_2\le K_3\,K_4$, where $K_3$ is the bound on $\sigma^{(L)}$ and $K_4$ is the bound on $\|\ba_{i_L}^{(L)}(0)\|_2$ (see assumptions \textbf{(B3)}-\textbf{(B4)});
\item[(ii)] $\tilde{\sigma}^{(L+1)}$ is bounded, with Fr\'echet derivatives bounded and Lipschitz.
\end{description} 

Define
\begin{equation}\label{eq:mlayerbis}
\begin{split}
(\bz_{i_{1}}^{(1)})'\left(\boldsymbol{x}, \boldsymbol{\theta}\right)&=\sigma^{(0)}\left(\boldsymbol{x}, \boldsymbol{\theta}_{i_{1}}^{(0)}\right),\qquad i_1\in[N],\\
(\bz_{i_{\ell+1}}^{(\ell+1)})'\left(\boldsymbol{x}, \boldsymbol{\theta}\right)&=\frac{1}{N} \sum_{i_{\ell}=1}^{N} \tilde{\sigma}^{(\ell)}\left((\bz_{i_{\ell}}^{(\ell)})'\left(\boldsymbol{x}, \boldsymbol{\theta}\right), \boldsymbol{\theta}_{i_{\ell}, i_{\ell+1}}^{(\ell)}\right),\qquad \ell\in[L-1], \,i_{\ell+1}\in[N],\\
\widehat{\boldsymbol{y}}_{N}'\left(\boldsymbol{x}, \boldsymbol{\theta}\right)&=\tilde{\sigma}^{(L+1)}\left(\frac{1}{N} \sum_{i_{L}=1}^{N} \tilde{\sigma}^{(L)}\left((\bz_{i_{L}}^{(L)})'\left(\boldsymbol{x}, \boldsymbol{\theta}\right), \boldsymbol{\theta}_{i_{L}}^{(L)}\right)\right),
\end{split}
\end{equation}
and
\begin{equation}\label{eq:riskmbis}
L_N'(\boldsymbol{\theta}) = \mathbb{E}\left\{ \big\|\boldsymbol{y} - \widehat{\boldsymbol{y}}_N'\left(\boldsymbol{x}, \boldsymbol{\theta}\right)\big\|_2^2\right\}.
\end{equation}
Let $\boldsymbol{\theta}'(k)$ be obtained by running $k$ steps of the SGD algorithm \eqref{eq:SGDmulti} with $\widehat{\boldsymbol{y}}_{N}\left(\boldsymbol{x}, \boldsymbol{\theta}\right)$ replaced by $\widehat{\boldsymbol{y}}_{N}'\left(\boldsymbol{x}, \boldsymbol{\theta}\right)$. Recall that $\ba_{i_\ell, i_{\ell+1}}^{(\ell)}(s)$ is bounded by Lemma \ref{lemma:normbdmulti}, $\ba_{i_L}^{(L)}(s)$ is bounded by assumption \textbf{(B4)} and $\sigma^{(\ell)}$ is bounded by assumption \textbf{(B3)}. Thus, we have that  $\boldsymbol{\theta}'(k)=\boldsymbol{\theta}(k)$ and $L_N'(\boldsymbol{\theta}'(k))=L_N(\boldsymbol{\theta}(k))$. To simplify notation, in the rest of the proof we will drop the symbol $'$ from $\btheta$ and $L_N$. By definition of dropout stability, the proof is completed by showing that, with probability at least $1-e^{-z^2}$,
\begin{equation}
|L_N(\boldsymbol{\theta}(k))-L_{|\mathcal A|}(\boldsymbol{\theta}_{\rm S}(k))|\le K(T, L) \left(\frac{\sqrt{d}+z}{\sqrt{N}} + \sqrt{\alpha}\big(\sqrt{d}+z\big)\right).
\end{equation}

By construction, the activation functions $\{\tilde{\sigma}^{(\ell)}\}_{\ell \in[L+1]}$ are bounded, with Fr\'echet derivatives that are bounded and Lipschitz. Thus, the technical assumptions of \cite{araujo2019mean} are fulfilled. Let $\rho^\star_{[0, T]}$ denote the unique solution to the McKean-Vlasov DNN problem with initial condition $\rho_0$ and activation functions $\sigma^{(0)}$ and $\tilde{\sigma}^{(\ell)}$, with $\ell \in\{0, \ldots, L+1\}$. Furthermore, let $\bar{\btheta}(t)$, with $t\in [0, T]$, be the associated ideal particles. Furthermore, let $\bar{\boldsymbol{\theta}}_{\rm S}(t)$ be obtained from $\bar{\boldsymbol{\theta}}(t)$ in the same way in which $\boldsymbol{\theta}_{\rm S}(k)$ is obtained from $\boldsymbol{\theta}(k)$. By triangle inequality, we have that
\begin{equation}\label{eq:trianglemulti}
\begin{split}
|L_N(\boldsymbol{\theta}(k))-L_{|\mathcal A|}(\boldsymbol{\theta}_{\rm S}(k))| &\le |L_N(\boldsymbol{\theta}(k))-\bar{L}(\rho^\star_T)| + |L_{|\mathcal A|}(\boldsymbol{\theta}_{\rm S}(k))-\bar{L}(\rho^\star_T)|\\
&\le  |L_N(\boldsymbol{\theta}(k))-L_N(\bar{\boldsymbol{\theta}}(T))|  +|L_{|\mathcal A|}(\boldsymbol{\theta}_{\rm S}(k))-L_{|\mathcal A|}(\bar{\boldsymbol{\theta}}_{\rm S}(T))|\\
&\hspace{6em}|L_N(\bar{\boldsymbol{\theta}}(T))-\bar{L}(\rho^\star_T)|+  |L_{|\mathcal A|}(\bar{\boldsymbol{\theta}}_{\rm S}(T))-\bar{L}(\rho^\star_T)|,
\end{split}
\end{equation}
where $\rho^\star_T$ denotes the marginal of $\rho^\star_{[0, T]}$ at time $T$ and $\bar{L}$ is defined in \eqref{eq:lriskm}.

Given a vector of parameters $\btheta$ containing $N_\ell$ neurons in layer $\ell$ ($\ell\in [L]$), we define the norm
\begin{equation}\label{eq:newnorm}
\|\boldsymbol{\theta}\|_{\infty} = \max\left( \sup_{i_{1}\in [N_{1}]} \left\|\boldsymbol{\theta}_{i_{1}}^{(0)}\right\|_2, \sup _{\ell \in[L-1], i_{\ell}\in [N_\ell], i_{\ell+1}\in [N_{\ell+1}]} \left\|\boldsymbol{\theta}_{i_{\ell}, i_{\ell+1}}^{(\ell)}\right\|_2, \sup_{i_{L}\in [N_{L}]}\left\|\boldsymbol{\theta}_{i_{L}}^{(L)}\right\|_2\right).
\end{equation}
As a preliminary result, we provide a bound on $\|\boldsymbol{\theta}(k)-\bar{\boldsymbol{\theta}}(T)\|_{\infty}$.

Consider the continuous time gradient descent process $\tilde{\btheta}(t)$, defined as
\begin{equation}\label{eq:SGDmultic}
\begin{split}
\tilde{\boldsymbol{\theta}}_{i_{1}}^{(0)}(t) &= \tilde{\boldsymbol{\theta}}_{i_{1}}^{(0)}(0),\\
\tilde{\boldsymbol{\theta}}^{(\ell)}_{i_{\ell}, i_{\ell + 1}}(t) = \tilde{\boldsymbol{\theta}}^{(\ell)}_{i_{\ell}, i_{\ell + 1}}(0) + 2 \int_0^t \alpha \xi(s) N^2 &\mathbb E\left\{ \left(\boldsymbol{y}-\widehat{\boldsymbol{y}}_{N}\left(\boldsymbol{x}, \tilde{\boldsymbol{\theta}}(s)\right)\right)^\sT
{\rm D}_{\tilde{\boldsymbol{\theta}}^{(\ell)}_{i_{\ell},i_{\ell + 1}}} \widehat{\boldsymbol{y}}_{N}\left(\boldsymbol{x}, \tilde{\boldsymbol{\theta}}(s)\right)\right\}{\rm d}s,\\
\tilde{\boldsymbol{\theta}}_{i_L}^{(L)}(t) &= \tilde{\boldsymbol{\theta}}_{i_L}^{(L)}(0),
\end{split}
\end{equation}
with the initialization $\tilde{\boldsymbol{\theta}}_{i_{1}}^{(0)}(0)=\boldsymbol{\theta}_{i_{1}}^{(0)}(0)$, $\tilde{\boldsymbol{\theta}}^{(\ell)}_{i_{\ell}, i_{\ell + 1}}(0) =\boldsymbol{\theta}^{(\ell)}_{i_{\ell}, i_{\ell + 1}}(0)$ and $\tilde{\boldsymbol{\theta}}_{i_L}^{(L)}(0)=\boldsymbol{\theta}_{i_L}^{(L)}(0)$. By triangle inequality, we have that
\begin{equation}\label{eq:triaux}
\|\boldsymbol{\theta}(k)-\bar{\boldsymbol{\theta}}(T)\|_{\infty}\le \|\boldsymbol{\theta}(k)-\tilde{\boldsymbol{\theta}}(T)\|_{\infty}+\|\tilde{\boldsymbol{\theta}}(T)-\bar{\boldsymbol{\theta}}(T)\|_{\infty}.
\end{equation}

In order to bound the first term in the RHS of \eqref{eq:triaux}, we follow a strategy similar to that of Proposition 10.1 in \cite{araujo2019mean}. From formula (10.8) of \cite{araujo2019mean}, we have that
\begin{equation}
\begin{split}
    \left\| \boldsymbol{\theta}_{i_{\ell}, i_{\ell+1}}^{(\ell)}(m)- \tilde{\boldsymbol{\theta}}_{i_{\ell}, i_{\ell+1}}^{(\ell)}(m\alpha)\right\|_2 &\leq 
    \alpha \left\|\operatorname{Mrt}_{i_{\ell},i_{\ell + 1}}^{(\ell)}(m)\right\|_2 + \\
   &\hspace{-3em} \sum\limits_{r=1}^m \int_{(r-1)\alpha}^{r\alpha} \mathbb{E}\Bigg\{ \Big\|\alpha\xi((r-1)\alpha) \left(\boldsymbol{y}-\widehat{\boldsymbol{y}}_{N}\left(\boldsymbol{x}, \boldsymbol{\theta}(r-1)\right)\right)^\sT
{\rm D}_{\boldsymbol{\theta}^{(\ell)}_{i_{\ell},i_{\ell + 1}}} \widehat{\boldsymbol{y}}_{N}\left(\boldsymbol{x}, \boldsymbol{\theta}(r-1)\right)\\
&\hspace{3em}- \alpha\xi(s)\left(\boldsymbol{y}-\widehat{\boldsymbol{y}}_{N}\left(\boldsymbol{x},\tilde{ \boldsymbol{\theta}}(s)\right)\right)^\sT
{\rm D}_{\tilde{\boldsymbol{\theta}}^{(\ell)}_{i_{\ell},i_{\ell + 1}}} \widehat{\boldsymbol{y}}_{N}\left(\boldsymbol{x}, \tilde{\boldsymbol{\theta}}(s)\right)\Big\|_2\Bigg\}{\rm d}s,
\end{split}
\end{equation}
where
\begin{equation}
\begin{split}
    \operatorname{Mrt}_{i_{\ell},i_{\ell + 1}}^{(\ell)}(m) = \sum\limits_{r=1}^m \alpha\xi((r-1)\alpha)\bigg(&\left(\boldsymbol{y}_{r-1}-\widehat{\boldsymbol{y}}_{N}\left(\boldsymbol{x}_{r-1}, \boldsymbol{\theta}(r-1)\right)\right)^\sT
{\rm D}_{\boldsymbol{\theta}^{(\ell)}_{i_{\ell},i_{\ell + 1}}} \widehat{\boldsymbol{y}}_{N}\left(\boldsymbol{x}_{r-1}, \boldsymbol{\theta}(r-1)\right) \\
&- \mathbb{E}\Big\{ \left(\boldsymbol{y}-\widehat{\boldsymbol{y}}_{N}\left(\boldsymbol{x}, \boldsymbol{\theta}(r-1)\right)\right)^\sT
{\rm D}_{\boldsymbol{\theta}^{(\ell)}_{i_{\ell},i_{\ell + 1}}} \widehat{\boldsymbol{y}}_{N}\left(\boldsymbol{x}, \boldsymbol{\theta}(r-1)\right)\Big\}\bigg)
\end{split}
\end{equation}
is a martingale with respect to the filtration $\left\{\mathcal{F}_m,\ m \in \mathbb{N}\right\}$ with $\mathcal{F}_m = \sigma\Big(\boldsymbol{\theta}(0), (\boldsymbol{x}_0, \boldsymbol{y}_0), \dots, (\boldsymbol{x}_{m-1}, \boldsymbol{y}_{m-1})\Big)$. By taking the $\sup$ on both sides, we have that
\begin{equation}\label{eq:I-II}
\begin{split}
  \sup _{\ell \in[L-1], i_{\ell}, i_{\ell+1}\in [N]}   \left\| \boldsymbol{\theta}_{i_{\ell}, i_{\ell+1}}^{(\ell)}(m)- \tilde{\boldsymbol{\theta}}_{i_{\ell}, i_{\ell+1}}^{(\ell)}(T)\right\|_2 &\leq 
   \overbrace{ \alpha \sup _{\ell \in[L-1], i_{\ell}, i_{\ell+1}\in [N]}\left\|\operatorname{Mrt}_{i_{\ell},i_{\ell + 1}}^{(\ell)}(m)\right\|_2}^{\mbox{\it (I)}} + \\
   &\hspace{-17em} \overbrace{\sum\limits_{r=1}^m \int_{(r-1)\alpha}^{r\alpha} \mathbb{E}\Bigg\{ \sup _{\ell \in[L-1], i_{\ell}, i_{\ell+1}\in [N]}\Big\|\alpha\xi((r-1)\alpha) \left(\boldsymbol{y}-\widehat{\boldsymbol{y}}_{N}\left(\boldsymbol{x}, \boldsymbol{\theta}(r-1)\right)\right)^\sT
{\rm D}_{\boldsymbol{\theta}^{(\ell)}_{i_{\ell},i_{\ell + 1}}} \widehat{\boldsymbol{y}}_{N}\left(\boldsymbol{x}, \boldsymbol{\theta}(r-1)\right)}^{\mbox{\it (II)}}\\
&\hspace{-2em}- \alpha\xi(s)\left(\boldsymbol{y}-\widehat{\boldsymbol{y}}_{N}\left(\boldsymbol{x},\tilde{ \boldsymbol{\theta}}(s)\right)\right)^\sT
{\rm D}_{\tilde{\boldsymbol{\theta}}^{(\ell)}_{i_{\ell},i_{\ell + 1}}} \widehat{\boldsymbol{y}}_{N}\left(\boldsymbol{x}, \tilde{\boldsymbol{\theta}}(s)\right)\Big\|_2\Bigg\}{\rm d}s.
\end{split}
\end{equation}
Given two parameters $\btheta_1$ and $\btheta_2$, by following the argument of Lemma B.17 of \cite{araujo2019mean}, we have that
\begin{equation}
\begin{split}
&\left\|\left(\boldsymbol{y}-\widehat{\boldsymbol{y}}_{N}\left(\boldsymbol{x}, \boldsymbol{\theta}_1\right)\right)^\sT
{\rm D}_{\boldsymbol{\theta}^{(\ell)}_{i_{\ell},i_{\ell + 1}}} \widehat{\boldsymbol{y}}_{N}\left(\boldsymbol{x}, \boldsymbol{\theta}_1\right)-\left(\boldsymbol{y}-\widehat{\boldsymbol{y}}_{N}\left(\boldsymbol{x}, \boldsymbol{\theta}_2\right)\right)^\sT
{\rm D}_{\boldsymbol{\theta}^{(\ell)}_{i_{\ell},i_{\ell + 1}}} \widehat{\boldsymbol{y}}_{N}\left(\boldsymbol{x}, \boldsymbol{\theta}_2\right)\right\|_2\\
&\hspace{30em}\le C_1 \|\btheta_1-\btheta_2\|_{\infty}.
\end{split}
\end{equation}
In what follows, the $C_i$ are constants that depend on $L$, $T$, and on the constants $K_i$ of the assumptions. 

Consequently, we can bound the second term in the RHS of \eqref{eq:I-II} as 
\begin{equation}
\mbox{\it (II)} \leq C_2\sum\limits_{r=1}^m \int_{(r-1)\varepsilon}^{r\varepsilon} \Big(
|(r-1)\varepsilon - s| + \|\boldsymbol{\theta}(r-1) - \tilde{\boldsymbol{\theta}}(s)\|_{\infty}
\Big) {\rm d}s,
\end{equation}
where we have used that the quantity
\begin{equation}
\left(\boldsymbol{y}-\widehat{\boldsymbol{y}}_{N}\left(\boldsymbol{x}, \boldsymbol{\theta}\right)\right)^\sT
{\rm D}_{\boldsymbol{\theta}^{(\ell)}_{i_{\ell},i_{\ell + 1}}} \widehat{\boldsymbol{y}}_{N}\left(\boldsymbol{x}, \boldsymbol{\theta}\right)
\end{equation}
is bounded for all $\btheta$. By using also that the process $t\to \tilde{\btheta}(t)$ is Lipschitz in time, we obtain the bound
\begin{equation}\label{eq:IIfin}
\mbox{\it (II)} \leq C_3\, \alpha \,T + C_3\, \alpha \sum\limits_{r=0}^{m-1} \left\|\btheta(r)- \tilde{\boldsymbol{\theta}}(r)\right\|_{\infty}.
\end{equation}
By combining \eqref{eq:IIfin} with \eqref{eq:I-II} and by applying a discrete Gronwall inequality, we have that
\begin{equation}\label{eq:dbg}
\left\|\boldsymbol{\theta}(k)-\tilde{\boldsymbol{\theta}}(T)\right\|_{\infty} \leq \alpha \, e^{C_{3}\,T} \Bigg(\sup _{m\in [k]}\left\|\operatorname{Mrt}(m)\right\|_\infty + C_3\,T \Bigg),
\end{equation}
where we have defined
\begin{equation}
\left\|\operatorname{Mrt}(m)\right\|_{\infty}=\sup _{\ell \in[L-1], i_{\ell}, i_{\ell+1}\in [N]}\left\|\operatorname{Mrt}_{i_{\ell},i_{\ell + 1}}^{(\ell)}(m)\right\|_2.
\end{equation}
Note that $e^{\zeta\left\|\operatorname{Mrt}(m)\right\|_{\infty}}$ is a submartingale. By using a Cram\'er-Chernoff argument, we have that
\begin{equation}\label{eq:ccb4}
\mathbb P\left(\sup _{m \in [k]}\left\|\operatorname{Mrt}_{N}(m)\right\|_{\infty}>u\right) \leq \inf_{\zeta\in \mathbb R^+}e^{-\zeta\cdot u} \mathbb{E}\left\{e^{\zeta\left\|\operatorname{Mrt}(\tau)\right\|_{\infty}}\right\}\le \inf_{\zeta\in \mathbb R^+}e^{-\zeta\cdot u} \mathbb{E}\left\{e^{\zeta\left\|\operatorname{Mrt}(k)\right\|_{\infty}}\right\},
\end{equation}
where $\tau=\inf\{m\le k, \left\|\operatorname{Mrt}_{N}(m)\right\|_{\infty}>u\}\wedge k$ is a stopping time, and in the second inequality  we have applied the optional stopping theorem to the submartingale $e^{\zeta\left\|\operatorname{Mrt}(m)\right\|_{\infty}}$. Furthermore, for any $\zeta>0$, we have that  
\begin{equation}\label{eq:ccb1}
\begin{split}
   \mathbb{E}\left\{e^{\zeta\left\|\operatorname{Mrt}(k)\right\|_{\infty}}\right\}&\leq \sum\limits_{\ell = 1}^{L} \sum\limits_{i_\ell, i_{\ell + 1} = 1}^{N}
   \mathbb{E}\left\{e^{\zeta\left\|\operatorname{Mrt}_{i_\ell, i_{\ell+1}}^{(\ell)}(k)\right\|_2}\right\}.
\end{split}
\end{equation}
Note that the martingale $\operatorname{Mrt}_{i_\ell, i_{\ell+1}}^{(\ell)}(k)$ has bounded increments. Thus, by using a modification of Hoeffding's Lemma and an $\varepsilon$-net argument (cf. Lemma A.3 of \cite{araujo2019mean}), we obtain that
\begin{equation}\label{eq:ccb2}
\mathbb{E}\left\{e^{\zeta\left\|\operatorname{Mrt}_{i_{\ell} i_{\ell+1}}^{(\ell)}(k)\right\|_2}\right\} \leq 5^{d} \cdot e^{C_4 k\zeta^2},
\end{equation}
with $d = \max\limits_{ i \in [L - 1]}d_i$. By combining \eqref{eq:ccb4}, \eqref{eq:ccb1} and \eqref{eq:ccb2}, we deduce that
\begin{equation}\label{eq:ccb3}
\mathbb P\left(\sup _{m \in [k]}\left\|\operatorname{Mrt}_{N}(m)\right\|_{\infty}>u\right) \leq L N^2 5^{d} \inf_{\zeta\in \mathbb R^+}e^{-\zeta\, u+C_4 k\zeta^2}.
\end{equation}
By optimizing over $\zeta$, we have that, with probability at least $1 - e^{-z^2}$,
\begin{equation}\label{eq:ccb5}
\sup _{m \in [k]}\left\|\operatorname{Mrt}_{N}(m)\right\|_{\infty}\le C_{5} \sqrt{\frac{1}{\alpha}}\left(\sqrt{d+\log N}+z\right).
\end{equation}
Finally, by combining \eqref{eq:ccb5} with \eqref{eq:dbg}, we conclude that, with probability at least $1 - e^{-z^2}$,
\begin{equation}\label{eq:strong1}
\left\|\boldsymbol{\theta}(k)-\tilde{\boldsymbol{\theta}}(T)\right\|_{\infty} \leq C_{6}\,\sqrt{\alpha} ( \sqrt{ d + \log N}+z).
\end{equation}

Let us bound the second term in the RHS of \eqref{eq:triaux}. By following the strategy of Lemma 12.2 in \cite{araujo2019mean}, we have that, with probability at least $1-e^{-u^2}$,
\begin{equation}
\left\|\tilde{\boldsymbol{\theta}}^{(\ell)}_{i_{\ell}, i_{\ell + 1}}(t) - \overline{\boldsymbol{\theta}}^{(\ell)}_{i_{\ell}, i_{\ell + 1}}(t)\right\|_{2} \leq C_7 \int_{0}^t \left\|\tilde{\boldsymbol{\theta}}(s) - \overline{\boldsymbol{\theta}}(s)\right\|_{\infty} {\rm d}s + C_7 \frac{u + \sqrt{d}}{\sqrt{N}}.
\end{equation}
By doing a union bound over $i_\ell, i_{\ell+1}\in [N]$ and $\ell\in [L-1]$, we deduce that, with probability at least $1-e^{-z^2}$,
\begin{equation}
\left\|\tilde{\boldsymbol{\theta}}(t) - \overline{\boldsymbol{\theta}}(t)\right\|_{\infty} \leq C_7 \int_{0}^t \left\|\tilde{\boldsymbol{\theta}}(s) - \overline{\boldsymbol{\theta}}(s)\right\|_{\infty} {\rm d}s + C_8 \frac{z + \sqrt{d+\log N}}{\sqrt{N}}.
\end{equation}
By Gronwall lemma, we conclude that, with probability at least $1-e^{-z^2}$,
\begin{equation}\label{eq:strong2}
\left\|\tilde{\boldsymbol{\theta}}(T) - \overline{\boldsymbol{\theta}}(T)\right\|_{\infty} \leq C_8 e^{C_7 T} \,\frac{z + \sqrt{d+\log N}}{\sqrt{N}}.
\end{equation}
By combining \eqref{eq:strong1} and \eqref{eq:strong2}, we have that, with probability at least $1-e^{-z^2}$, 
\begin{equation}\label{eq:strong3}
\|\boldsymbol{\theta}(k)-\bar{\boldsymbol{\theta}}(T)\|_{\infty}\le C_9\left(\frac{z + \sqrt{d+\log N}}{\sqrt{N}} +\sqrt{\alpha} ( \sqrt{ d + \log N}+z)\right) .
\end{equation}

At this point, we are ready to bound the various terms in the RHS of \eqref{eq:trianglemulti}. In order to bound the first term, note that $L_N$ is Lipschitz with $\|\cdot\|_{\infty}$. Thus, we obtain that, with probability at least $1-e^{-z^2}$,
\begin{equation}\label{eq:bdm1}
|L_N(\boldsymbol{\theta}(k))-L_N(\bar{\boldsymbol{\theta}}(T))|\le C_{10}\left(\frac{z + \sqrt{d+\log N}}{\sqrt{N}} +\sqrt{\alpha} ( \sqrt{ d + \log N}+z)\right) .
\end{equation}

In order to bound the second term in the RHS of \eqref{eq:trianglemulti}, note that 
\begin{equation}\label{eq:Lb}
\|\boldsymbol{\theta}_{\rm S}(k)-\bar{\boldsymbol{\theta}}_{\rm S}(T)\|_{\infty}\le \|\boldsymbol{\theta}(k)-\bar{\boldsymbol{\theta}}(T)\|_{\infty}.
\end{equation}
As $L_{|\mathcal A|}$ is Lipschitz with $\|\cdot\|_{\infty}$, by combining \eqref{eq:strong3} and \eqref{eq:Lb}, we obtain the bound
\begin{equation}\label{eq:bdm2}
|L_{|\mathcal A|}(\boldsymbol{\theta}_{\rm S}(k))-L_{|\mathcal A|}(\bar{\boldsymbol{\theta}}_{\rm S}(T))|\le C_{11}\left(\frac{z + \sqrt{d+\log N}}{\sqrt{N}} +\sqrt{\alpha} ( \sqrt{ d + \log N}+z)\right) ,
\end{equation}
with probability at least $1-e^{-z^2}$.

Finally, let us consider the remaining two terms in the RHS of \eqref{eq:trianglemulti}. Fix $\bx\in\mathbb R^{d_0}$. Then, by Lemma 11.4 of \cite{araujo2019mean}, we have that, for $\zeta>0$,
\begin{equation}
\log \mathbb E\left\{e^{\zeta\|\widehat{\boldsymbol{y}}_{N}\left(\boldsymbol{x}, \bar{\boldsymbol{\theta}}(T)\right)-\bar{\by}(\bx, \rho^\star_T)\|_2}\right\}\le C_{12}\left(d+\frac{\zeta^2}{N}\right).
\end{equation}
By using similar arguments, we also have that, for $\zeta>0$,
\begin{equation}
\log \mathbb E\left\{e^{\zeta\|\widehat{\boldsymbol{y}}_{|\mathcal A|}\left(\boldsymbol{x}, \bar{\boldsymbol{\theta}}_{\rm S}(T)\right)-\bar{\by}(\bx, \rho^\star_T)\|_2}\right\}\le C_{13}\left(d+\frac{\zeta^2}{A_{\rm min}}\right).
\end{equation}

Thus, by applying Markov inequality and optimizing over $\zeta$, we deduce that 
\begin{equation}
\begin{split}
\|\widehat{\boldsymbol{y}}_{N}\left(\boldsymbol{x}, \bar{\boldsymbol{\theta}}(T)\right)-\bar{\by}(\bx, \rho^\star_T)\|_2&\le C_{14}\frac{\sqrt{d}+z}{\sqrt{N}},\\
\|\widehat{\boldsymbol{y}}_{|\mathcal A|}\left(\boldsymbol{x}, \bar{\boldsymbol{\theta}}_{\rm S}(T)\right)-\bar{\by}(\bx, \rho^\star_T)\|_2&\le C_{14}\frac{\sqrt{d}+z}{\sqrt{A_{\rm min}}},
\end{split}
\end{equation}
with probability at least $1-e^{-z^2}$. By using that $\by$, $\widehat{\boldsymbol{y}}_{N}\left(\boldsymbol{x}, \bar{\boldsymbol{\theta}}(T)\right)$, $\widehat{\boldsymbol{y}}_{|\mathcal A|}\left(\boldsymbol{x}, \bar{\boldsymbol{\theta}}_{\rm S}(T)\right)$ and $\bar{\by}(\bx, \rho^\star_T)$ are bounded, we conclude that 
\begin{equation}\label{eq:bdm3}
\begin{split}
|L_N(\bar{\boldsymbol{\theta}}(T))-\bar{L}(\rho^\star_T)|&\le  C_{15}\frac{\sqrt{d}+z}{\sqrt{N}},\\
|L_{|\mathcal A|}(\bar{\boldsymbol{\theta}}_{\rm S}(T))-\bar{L}(\rho^\star_T)|&\le  C_{15}\frac{\sqrt{d}+z}{\sqrt{A_{\rm min}}},
\end{split}
\end{equation}
with probability at least $1-e^{-z^2}$. By combining \eqref{eq:bdm1}, \eqref{eq:bdm2} and \eqref{eq:bdm3}, the proof is complete. 
\end{proof}

\subsection{Part (B)}

The proof of part (B) is obtained by combining part (A) with the following result, which extends Lemma \ref{lemma:dc} to the multilayer case.  

\begin{lemma}[Dropout stability implies connectivity -- multilayer]\label{lemma:dcm}
Consider a neural network with $L+1\ge 4$ layers, where each hidden layer contains $N$ neurons, as in \eqref{eq:mlayer}. For any $k\in [L]$, assume that $\boldsymbol{\theta}$ and $\bar{\boldsymbol{\theta}}$
are $\varepsilon$-dropout stable given $\mathcal A_i=[N/2]$ for $i\in \{k, \ldots, L\}$. Then, $\boldsymbol{\theta}$
and $\bar{\boldsymbol{\theta}}$
are $\varepsilon$-connected. 
\end{lemma}

Given a vector of parameters $\btheta$, it is helpful to write it as
\begin{equation}
\begin{split}
\boldsymbol{\theta}^{(L)} &= \left\{\left[\boldsymbol{a}_{i_L}^{(L)}\right]_{i_L\in [N]}, \left[\boldsymbol{w}_{i_L}^{(L)}\right]_{i_L\in [N]}\right\},\\
\boldsymbol{\theta}^{(\ell)} &= \left\{\left[\boldsymbol{a}_{i_{\ell+1},i_{\ell}}^{(\ell)}\right]_{i_{\ell+1},i_\ell\in [N]}, \left[\boldsymbol{w}_{i_{\ell+1},i_{\ell}}^{(\ell)}\right]_{i_{\ell+1},i_\ell\in [N]} \right\}, \qquad\ell\in [L-1],\\
\boldsymbol{\theta}^{(0)} &= \left[\boldsymbol{\theta}_{i_0}^{(0)}\right]_{i_0 \in [N]}.
\end{split}
\end{equation}
In words, we stack the parameters $\boldsymbol{\theta}^{(\ell)}$ of layer $\ell$ into a matrix, and the $(i, j)$-th element of this matrix contains the parameter $\boldsymbol{\theta}^{(\ell)}_{j, i}=(\boldsymbol{a}^{(\ell)}_{j, i}, \boldsymbol{w}^{(\ell)}_{j, i})$ connecting the $j$-th neuron of layer $\ell$ with the $i$-th neuron of layer $\ell+1$. Furthermore, let us partition the parameters $\boldsymbol{\theta}$ as
\begin{equation}
\begin{split}
    \boldsymbol{\theta}^{(L)} &= \left\{\left[\begin{array}{c|c}
 \boldsymbol{a}^{(L)}_{\rm t} & \boldsymbol{a}^{(L)}_{\rm b}
\end{array}\right],\left[\begin{array}{c|c}
 \boldsymbol{w}^{(L)}_{\rm t} & \boldsymbol{w}^{(L)}_{\rm b}
\end{array}\right]\right\},\\
    \boldsymbol{\theta}^{(\ell)} &= \left\{\left[\begin{array}{c|c}
 \boldsymbol{a}_{\rm t, t}^{(\ell)} &  \boldsymbol{a}_{\rm t, b}^{(\ell)} \\ \hline
  \boldsymbol{a}_{\rm b, t}^{(\ell)} &  \boldsymbol{a}_{\rm b, b}^{(\ell)}
\end{array}\right],\ \left[\begin{array}{c|c}
 \boldsymbol{w}_{\rm t, t}^{(\ell)} &  \boldsymbol{w}_{\rm t, b}^{(\ell)} \\ \hline
  \boldsymbol{w}_{\rm b, t}^{(\ell)} &  \boldsymbol{w}_{\rm b, b}^{(\ell)}
\end{array}\right]\right\},\qquad \ell \in [L-1],\\ 
\boldsymbol{\theta}^{(0)} &= \left[\begin{array}{cc}
     \boldsymbol{\theta}_{\rm t}^{(0)}  \\ \hline
     \boldsymbol{\theta}_{\rm b}^{(0)}
\end{array}\right].
\end{split}
\end{equation}
In words, $\boldsymbol{\theta}_{\rm t, t}^{(\ell)}=(\boldsymbol{a}_{\rm t, t}^{(\ell)}, \boldsymbol{w}_{\rm t, t}^{(\ell)})$ contains the parameters connecting the top half neurons of layer $\ell$ with the top half neurons of layer $\ell+1$; $\boldsymbol{\theta}_{\rm t, b}^{(\ell)}=(\boldsymbol{a}_{\rm t, b}^{(\ell)}, \boldsymbol{w}_{\rm t, b}^{(\ell)})$ contains the parameters connecting the bottom half neurons of layer $\ell$ with the top half neurons of layer $\ell+1$; $\boldsymbol{\theta}_{\rm b, t}^{(\ell)}=(\boldsymbol{a}_{\rm b, t}^{(\ell)}, \boldsymbol{w}_{\rm b, t}^{(\ell)})$ contains the parameters connecting the top half neurons of layer $\ell$ with the bottom half neurons of layer $\ell+1$; and $\boldsymbol{\theta}_{\rm b, b}^{(\ell)}=(\boldsymbol{a}_{\rm b, b}^{(\ell)}, \boldsymbol{w}_{\rm b, b}^{(\ell)})$ contains the parameters connecting the bottom half neurons of layer $\ell$ with the bottom half neurons of layer $\ell+1$. The partition for the first and the last layer is similarly defined. 

At this point, we are ready to present the proof of Lemma \ref{lemma:dcm}.

\begin{proof}[Proof of Lemma \ref{lemma:dcm}] For the moment, assume that $N$ is even. Let $\btheta_{{\rm S}, k}$ be obtained from $\btheta$ by keeping only the top half neurons at layer $\ell\in \{k, \ldots, L\}$. With an abuse of notation, we can partition the parameters $\boldsymbol{\theta}_{{\rm S}, k}$ as
\begin{equation}
\begin{split}
    \boldsymbol{\theta}^{(L)}_{{\rm S}, k} &= \left\{\left[\begin{array}{c|c}
 2\boldsymbol{a}^{(L)}_{\rm t} & \boldsymbol{0}
\end{array}\right],\left[\begin{array}{c|c}
 \boldsymbol{w}^{(L)}_{\rm t} & \boldsymbol{0}
\end{array}\right]\right\},\\
    \boldsymbol{\theta}^{(\ell)}_{{\rm S}, k} &= \left\{\left[\begin{array}{c|c}
 2\boldsymbol{a}_{\rm t, t}^{(\ell)} &  \boldsymbol{0} \\ \hline
  \boldsymbol{0} &  \boldsymbol{0}
\end{array}\right],\ \left[\begin{array}{c|c}
 \boldsymbol{w}_{\rm t, t}^{(\ell)} &  \boldsymbol{0} \\ \hline
  \boldsymbol{0} &  \boldsymbol{0}
\end{array}\right]\right\},\qquad \ell \in \{k, \ldots, L-1\},\\ 
    \boldsymbol{\theta}^{(\ell)}_{{\rm S}, k} &= \left\{\left[\begin{array}{c|c}
 \boldsymbol{a}_{\rm t, t}^{(\ell)} &  \boldsymbol{a}_{\rm t, b}^{(\ell)} \\ \hline
  \boldsymbol{a}_{\rm b, t}^{(\ell)} &  \boldsymbol{a}_{\rm b, b}^{(\ell)}
\end{array}\right],\ \left[\begin{array}{c|c}
 \boldsymbol{w}_{\rm t, t}^{(\ell)} &  \boldsymbol{w}_{\rm t, b}^{(\ell)} \\ \hline
  \boldsymbol{w}_{\rm b, t}^{(\ell)} &  \boldsymbol{w}_{\rm b, b}^{(\ell)}
\end{array}\right]\right\},\qquad \ell \in [k-1],\\ 
\boldsymbol{\theta}^{(0)}_{{\rm S}, k} &= \left[\begin{array}{cc}
     \boldsymbol{\theta}_{\rm t}^{(0)}  \\ \hline
    \boldsymbol{\theta}_{\rm b}^{(0)}
\end{array}\right],
\end{split}
\end{equation}
and the corresponding loss is given by $L_N(\btheta_{{\rm S}, k})$. We now prove by induction that $\btheta$ is connected to $\btheta_{{\rm S}, k}$ via a piecewise linear path in parameter space, such that the loss along the path is upper bounded by  $L_N(\btheta)+\varepsilon$.

\ \\
\emph{Base step: from $\btheta$ to $\btheta_{{\rm S}, L}$}. As $\btheta$ is $\varepsilon$-dropout stable, we have that $L_N(\btheta_{{\rm S}, L})\le L_N(\btheta) + \varepsilon$. Note that if $\boldsymbol{a}^{(L)}_{\rm t}=\b0$, then the value of $\boldsymbol{w}^{(L)}_{\rm t}$ does not affect the loss. Hence, we can interpolate from $\{[\,\,2\boldsymbol{a}^{(L)}_{\rm t} \mid \boldsymbol{0}\,\,], [\,\,\boldsymbol{w}^{(L)}_{\rm t} \mid \boldsymbol{0}\,\,]\}$ to $\{[\,\,2\boldsymbol{a}^{(L)}_{\rm t} \mid \boldsymbol{0}\,\,], [\,\,\boldsymbol{w}^{(L)}_{\rm t} \mid \boldsymbol{w}^{(L)}_{\rm b}\,\,]\}$ with no change in loss. Furthermore, the loss is convex in $\boldsymbol{a}^{(L)}$. Thus, we can interpolate from $\{[\,\,\boldsymbol{a}^{(L)}_{\rm t} \mid \boldsymbol{a}^{(L)}_{\rm b}\,\,], [\,\,\boldsymbol{w}^{(L)}_{\rm t} \mid \boldsymbol{w}^{(L)}_{\rm b}\,\,]\}$ to $\{[\,\,2\boldsymbol{a}^{(L)}_{\rm t} \mid \b0\,\,], [\,\,\boldsymbol{w}^{(L)}_{\rm t} \mid \boldsymbol{w}^{(L)}_{\rm b}\,\,]\}$ while keeping the loss upper bounded by $L_N(\btheta)+\varepsilon$.

\ \\
\emph{Induction step: from $\btheta_{{\rm S}, k}$ to $\btheta_{{\rm S}, k-1}$}. We construct the path by passing through the following intermediate points in parameter space:
\begin{align*}
& \boldsymbol{\theta}^{(L)}_1 =\left\{ \left[\begin{array}{c|c}
 2\boldsymbol{a}^{(L)}_{\rm t} & \boldsymbol{0}
\end{array}\right],\left[\begin{array}{c|c}
 \boldsymbol{w}^{(L)}_{\rm t} & \boldsymbol{0}
\end{array}\right]\right\},\\
&\boldsymbol{\theta}^{(i)}_1 = \left\{\left[\begin{array}{c|c}
 2\boldsymbol{a}_{\rm t, t}^{(i)} &  \boldsymbol{0} \\ \hline
  \boldsymbol{0} &  \boldsymbol{0}
\end{array}\right], \left[\begin{array}{c|c}
 \boldsymbol{w}_{\rm t, t}^{(i)}&  \boldsymbol{0} \\ \hline
  \boldsymbol{0} &  \boldsymbol{0}
\end{array}\right] \right\},\qquad i\in \{k, \ldots, L-1\}, \\
& \boldsymbol{\theta}^{(k-1)}_1 =\left\{\left[\begin{array}{c|c}
 \boldsymbol{a}_{\rm t, t}^{(k-1)} &  \boldsymbol{a}_{\rm t, b}^{(k-1)} \\ \hline
  \boldsymbol{a}_{\rm b, t}^{(k-1)} &  \boldsymbol{a}_{\rm b, b}^{(k-1)}
\end{array}\right],\ \left[\begin{array}{c|c}
 \boldsymbol{w}_{\rm t, t}^{(k-1)} &  \boldsymbol{w}_{\rm t, b}^{(k-1)} \\ \hline
  \boldsymbol{w}_{\rm b, t}^{(k-1)} &  \boldsymbol{w}_{\rm b, b}^{(k-1)}
\end{array}\right]\right\}. 
\end{align*}

\begin{align*}
& \boldsymbol{\theta}^{(L)}_2 =\left\{ \left[\begin{array}{c|c}
 2\boldsymbol{a}^{(L)}_{\rm t} & \boldsymbol{0}
\end{array}\right],\left[\begin{array}{c|c}
 \boldsymbol{w}^{(L)}_{\rm t} & \boldsymbol{0}
\end{array}\right]\right\},\\
&\boldsymbol{\theta}^{(i)}_2 = \left\{\left[\begin{array}{c|c}
 2\boldsymbol{a}_{\rm t, t}^{(i)} &  \boldsymbol{0} \\ \hline
  \boldsymbol{0} &  2\boldsymbol{a}_{\rm t, t}^{(i)}
\end{array}\right], \left[\begin{array}{c|c}
 \boldsymbol{w}_{\rm t, t}^{(i)}&  \boldsymbol{0} \\ \hline
  \boldsymbol{0} &  \boldsymbol{w}_{\rm t, t}^{(i)}
\end{array}\right] \right\},\qquad i\in \{k, \ldots, L-1\}, \\
& \boldsymbol{\theta}^{(k-1)}_2 =\left\{\left[\begin{array}{c|c}
 \boldsymbol{a}_{\rm t, t}^{(k-1)} &  \boldsymbol{a}_{\rm t, b}^{(k-1)} \\ \hline
  2\boldsymbol{a}_{\rm t, t}^{(k-1)} &  \boldsymbol{0}
\end{array}\right],\ \left[\begin{array}{c|c}
 \boldsymbol{w}_{\rm t, t}^{(k-1)} &  \boldsymbol{w}_{\rm t, b}^{(k-1)} \\ \hline
  \boldsymbol{w}_{\rm t, t}^{(k-1)} &  \boldsymbol{0}
\end{array}\right]\right\}.
\end{align*}

\begin{align*}
& \boldsymbol{\theta}^{(L)}_3 =\left\{ \left[\begin{array}{c|c}
\b0 & 2\boldsymbol{a}^{(L)}_{\rm t} 
\end{array}\right],\left[\begin{array}{c|c}
 \b0 & \boldsymbol{w}^{(L)}_{\rm t} 
\end{array}\right]\right\},\\
&\boldsymbol{\theta}^{(i)}_3 = \left\{\left[\begin{array}{c|c}
 2\boldsymbol{a}_{\rm t, t}^{(i)} &  \boldsymbol{0} \\ \hline
  \boldsymbol{0} &  2\boldsymbol{a}_{\rm t, t}^{(i)}
\end{array}\right], \left[\begin{array}{c|c}
 \boldsymbol{w}_{\rm t, t}^{(i)}&  \boldsymbol{0} \\ \hline
  \boldsymbol{0} &  \boldsymbol{w}_{\rm t, t}^{(i)}
\end{array}\right] \right\},\qquad i\in \{k, \ldots, L-1\}, \\
& \boldsymbol{\theta}^{(k-1)}_3 =\left\{\left[\begin{array}{c|c}
 \boldsymbol{a}_{\rm t, t}^{(k-1)} &  \boldsymbol{a}_{\rm t, b}^{(k-1)} \\ \hline
  2\boldsymbol{a}_{\rm t, t}^{(k-1)} &  \boldsymbol{0}
\end{array}\right],\ \left[\begin{array}{c|c}
 \boldsymbol{w}_{\rm t, t}^{(k-1)} &  \boldsymbol{w}_{\rm t, b}^{(k-1)} \\ \hline
  \boldsymbol{w}_{\rm t, t}^{(k-1)} &  \boldsymbol{0}
\end{array}\right]\right\}.
\end{align*}

\begin{align*}
& \boldsymbol{\theta}^{(L)}_4 =\left\{ \left[\begin{array}{c|c}
 \b0 & 2\boldsymbol{a}^{(L)}_{\rm t}
\end{array}\right],\left[\begin{array}{c|c}
 \b0 & \boldsymbol{w}^{(L)}_{\rm t} 
\end{array}\right]\right\},\\
&\boldsymbol{\theta}^{(i)}_4 = \left\{\left[\begin{array}{c|c}
 2\boldsymbol{a}_{\rm t, t}^{(i)} &  \boldsymbol{0} \\ \hline
  \boldsymbol{0} &  2\boldsymbol{a}_{\rm t, t}^{(i)}
\end{array}\right], \left[\begin{array}{c|c}
 \boldsymbol{w}_{\rm t, t}^{(i)}&  \boldsymbol{0} \\ \hline
  \boldsymbol{0} &  \boldsymbol{w}_{\rm t, t}^{(i)}
\end{array}\right] \right\},\qquad i\in \{k, \ldots, L-1\}, \\
& \boldsymbol{\theta}^{(k-1)}_4 =\left\{\left[\begin{array}{c|c}
 2\boldsymbol{a}_{\rm t, t}^{(k-1)} &  \b0 \\ \hline
  2\boldsymbol{a}_{\rm t, t}^{(k-1)} &  \boldsymbol{0}
\end{array}\right],\ \left[\begin{array}{c|c}
 \boldsymbol{w}_{\rm t, t}^{(k-1)} &  \b0 \\ \hline
  \boldsymbol{w}_{\rm t, t}^{(k-1)} &  \boldsymbol{0}
\end{array}\right]\right\}. 
\end{align*}

\begin{align*}
& \boldsymbol{\theta}^{(L)}_5 =\left\{ \left[\begin{array}{c|c}
 2\boldsymbol{a}^{(L)}_{\rm t} & \boldsymbol{0}
\end{array}\right],\left[\begin{array}{c|c}
 \boldsymbol{w}^{(L)}_{\rm t} & \boldsymbol{0}
\end{array}\right]\right\},\\
&\boldsymbol{\theta}^{(i)}_5 = \left\{\left[\begin{array}{c|c}
 2\boldsymbol{a}_{\rm t, t}^{(i)} &  \boldsymbol{0} \\ \hline
  \boldsymbol{0} &  2\boldsymbol{a}_{\rm t, t}^{(i)}
\end{array}\right], \left[\begin{array}{c|c}
 \boldsymbol{w}_{\rm t, t}^{(i)}&  \boldsymbol{0} \\ \hline
  \boldsymbol{0} &  \boldsymbol{w}_{\rm t, t}^{(i)}
\end{array}\right] \right\},\qquad i\in \{k, \ldots, L-1\}, \\
& \boldsymbol{\theta}^{(k-1)}_5 =\left\{\left[\begin{array}{c|c}
 2\boldsymbol{a}_{\rm t, t}^{(k-1)} &  \b0 \\ \hline
  2\boldsymbol{a}_{\rm t, t}^{(k-1)} &  \boldsymbol{0}
\end{array}\right],\ \left[\begin{array}{c|c}
 \boldsymbol{w}_{\rm t, t}^{(k-1)} &  \b0 \\ \hline
  \boldsymbol{w}_{\rm t, t}^{(k-1)} &  \boldsymbol{0}
\end{array}\right]\right\}. 
\end{align*}

\begin{align*}
& \boldsymbol{\theta}^{(L)}_6 =\left\{ \left[\begin{array}{c|c}
 2\boldsymbol{a}^{(L)}_{\rm t} & \boldsymbol{0}
\end{array}\right],\left[\begin{array}{c|c}
 \boldsymbol{w}^{(L)}_{\rm t} & \boldsymbol{0}
\end{array}\right]\right\},\\
&\boldsymbol{\theta}^{(i)}_6 = \left\{\left[\begin{array}{c|c}
 2\boldsymbol{a}_{\rm t, t}^{(i)} &  \boldsymbol{0} \\ \hline
  \boldsymbol{0} &  \boldsymbol{0}
\end{array}\right], \left[\begin{array}{c|c}
 \boldsymbol{w}_{\rm t, t}^{(i)}&  \boldsymbol{0} \\ \hline
  \boldsymbol{0} &  \boldsymbol{0}
\end{array}\right] \right\},\qquad i\in \{k-1, \ldots, L-1\}.
\end{align*}
As we do not change the parameters in layer $\ell\in [k-2]$, we have omitted them in the definitions above.

\emph{From $\btheta_1$ to $\btheta_2$.} The loss is not affected by the values in the bottom right quadrant of $\btheta_1^{(k-1)}$, since the bottom neurons of layer $k$ are not active ($\ba^{(k)}_{\rm t, b}=\ba^{(k)}_{\rm b, b}=\b0$). Consequently, we can interpolate from $\ba_{\rm b, b}^{(k-1)}$ to $\b0$ and from $\bw_{\rm b, b}^{(k-1)}$ to $\b0$ with no change in loss. Similarly, the loss is not affected by the values in the bottom right quadrant of $\btheta_1^{(i)}$ for $i\in \{k, \ldots L-1\}$, since the bottom neurons of layer $i+1$ are not active ($\ba^{(i+1)}_{\rm t, b}=\ba^{(i+1)}_{\rm b, b}=\b0$ and $\boldsymbol{a}^{(L)}_{\rm b}=\b0$). Consequently, for $i\in \{k, \ldots L-1\}$, we can successively interpolate from $\b0$ to $2\ba_{\rm t, t}^{(i)}$ and from $\b0$ to $2\bw_{\rm t, t}^{(i)}$ with no change in loss.   

\emph{From $\btheta_5$ to $\btheta_6$.} We use the same reasoning as for $\btheta_1\to\btheta_2$ and go in reverse layer order (i.e., from layer $L-1$ to layer $k-1$). The loss is not affected by the values in the bottom right quadrant of $\btheta_5^{(i)}$, since the bottom neurons of layer $i+1$ are not active. Consequently, we can interpolate from $2\ba_{\rm t, t}^{(i)}$ to $\b0$ and from $\bw_{\rm t, t}^{(i)}$ to $\b0$ with no change in loss. Similarly, the loss is not affected by the values in the bottom left quadrant of $\btheta_5^{(k-1)}$, since the bottom neurons of layer $k$ are not active. Consequently, we can interpolate from $2\ba_{\rm t, t}^{(k-1)}$ to $\b0$ and from $\bw_{\rm t, t}^{(k-1)}$ to $\b0$ with no change in loss.

\emph{From $\btheta_4$ to $\btheta_5$.} Note that the parameters of $\btheta_4$ and $\btheta_5$ are the same except for layer $L$. Furthermore, the structure of these parameters implies that the output of layer $L - 1$ is obtained by stacking the output of two identical sub-networks. In formulas, let $\boldsymbol{z}^{(L-1)}$ be the output of layer $L-1$. Then, $\boldsymbol{z}^{(L-1)} = [\,\,\bar{\bz}\mid \bar{\bz}\,\,]$ for some $\bar{\boldsymbol{z}}$. Consequently, we can interpolate between $\btheta_4$ and $\btheta_5$ with no change in loss. 

\emph{From $\btheta_3$ to $\btheta_4$.} By using the same reasoning as for $\btheta_5\to\btheta_6$, we interpolate from $2\ba_{\rm t, t}^{(i)}$ to $\b0$ and from $\bw_{\rm t, t}^{(i)}$ to $\b0$ in the top left corner of $\btheta_3^{(i)}$ with no change in loss, for $i=L-1, \ldots, k$. Then, we interpolate from $\btheta_3^{(k-1)}$ to $\btheta_4^{(k-1)}$ with no change in loss, since the top neurons of layer $k$ are not active. Finally, we restore sequentially $2\ba_{\rm t, t}^{(i)}$ and $\bw_{\rm t, t}^{(i)}$  in the top left corner of the corresponding parameter matrices with no change in loss, by using the same reasoning as for $\btheta_1\to \btheta_2$. 

\emph{From $\btheta_2$ to $\btheta_3$.} From the previous arguments, we have that $L_N(\btheta_2)= L_N(\btheta_1)$ and $L_N(\btheta_3)= L_N(\btheta_6)$. Furthermore, $\btheta$ is $\varepsilon$-dropout stable, which implies that  $|L_N(\btheta_1)- L_N(\btheta_6)|\le \varepsilon$. Consequently, we have that $|L_N(\btheta_2)- L_N(\btheta_3)|\le \varepsilon$. Note that if $\boldsymbol{a}^{(L)}_{\rm t}=\b0$, then the value of $\boldsymbol{w}^{(L)}_{\rm t}$ does not affect the loss. Hence, we can interpolate from $\{[\,\,2\boldsymbol{a}^{(L)}_{\rm t} \mid \boldsymbol{0}\,\,], [\,\,\boldsymbol{w}^{(L)}_{\rm t} \mid \boldsymbol{0}\,\,]\}$ to $\{[\,\,2\boldsymbol{a}^{(L)}_{\rm t} \mid \boldsymbol{0}\,\,], [\,\,\boldsymbol{w}^{(L)}_{\rm t} \mid \boldsymbol{w}^{(L)}_{\rm t}\,\,]\}$ with no change in loss. Similarly, we can interpolate from $\{[\,\,\b0 \mid 2\boldsymbol{a}^{(L)}_{\rm t} \,\,], [\,\,\b0 \mid \boldsymbol{w}^{(L)}_{\rm t} \,\,]\}$ to $\{[\,\,\b0 \mid 2\boldsymbol{a}^{(L)}_{\rm t} \,\,], [\,\,\boldsymbol{w}^{(L)}_{\rm t} \mid \boldsymbol{w}^{(L)}_{\rm t}\,\,]\}$ with no change in loss. Furthermore, the loss is convex in $\boldsymbol{a}^{(L)}$. Thus, we can interpolate from $\{[\,\,2\boldsymbol{a}^{(L)}_{\rm t} \mid \boldsymbol{0}\,\,], [\,\,\boldsymbol{w}^{(L)}_{\rm t} \mid \boldsymbol{w}^{(L)}_{\rm t}\,\,]\}$ to $\{[\,\,\b0\mid 2\boldsymbol{a}^{(L)}_{\rm t}\,\,], [\,\,\boldsymbol{w}^{(L)}_{\rm t} \mid \boldsymbol{w}^{(L)}_{\rm t}\,\,]\}$ while keeping the loss upper bounded by $L_N(\btheta)+\varepsilon$.

\ \\
As a result, we are able to connect $\btheta$ with $\btheta_{{\rm S}, 1}$ via a piecewise linear path, where the loss is upper bounded by $L_N(\btheta)+\varepsilon$. Similarly, let $\bar{\btheta}_{{\rm S}, k}$ be obtained from $\bar{\btheta}$ by keeping only the top half neurons at layer $\ell\in \{k, \ldots, L\}$. Then, we can connect $\bar{\btheta}$ with $\bar{\btheta}_{{\rm S}, 1}$ via a piecewise linear path, where the loss is upper bounded by $L_N(\bar{\btheta})+\varepsilon$.

In order to complete the proof, it remains to connect $\btheta_{{\rm S}, 1}$  with $\bar{\btheta}_{{\rm S}, 1}$ via a piecewise linear path, where the loss is upper bounded by $\max(L_N(\btheta), L_N(\bar{\btheta}))+\varepsilon$. We construct the path by passing through the following intermediate points in parameter space:

\begin{align*}
& \tilde{\boldsymbol{\theta}}^{(L)}_1 =\left\{ \left[\begin{array}{c|c}
 2\boldsymbol{a}^{(L)}_{\rm t} & \boldsymbol{0}
\end{array}\right],\left[\begin{array}{c|c}
 \boldsymbol{w}^{(L)}_{\rm t} & \boldsymbol{0}
\end{array}\right]\right\},\\
&\tilde{\boldsymbol{\theta}}^{(i)}_1 = \left\{\left[\begin{array}{c|c}
 2\boldsymbol{a}_{\rm t, t}^{(i)} &  \boldsymbol{0} \\ \hline
  \boldsymbol{0} &  \boldsymbol{0}
\end{array}\right], \left[\begin{array}{c|c}
 \boldsymbol{w}_{\rm t, t}^{(i)}&  \boldsymbol{0} \\ \hline
  \boldsymbol{0} &  \boldsymbol{0}
\end{array}\right] \right\},\qquad i\in \{1, \ldots, L-1\}, \\
& \tilde{\boldsymbol{\theta}}^{(0)}_1 = \left[\begin{array}{cc}
     \boldsymbol{\theta}_{\rm t}^{(0)}  \\ \hline
     \boldsymbol{\theta}_{\rm b}^{(0)}
\end{array}\right]. 
\end{align*}

\begin{align*}
& \tilde{\boldsymbol{\theta}}^{(L)}_2 =\left\{ \left[\begin{array}{c|c}
 2\boldsymbol{a}^{(L)}_{\rm t} & \boldsymbol{0}
\end{array}\right],\left[\begin{array}{c|c}
 \boldsymbol{w}^{(L)}_{\rm t} & \boldsymbol{0}
\end{array}\right]\right\},\\
&\tilde{\boldsymbol{\theta}}^{(i)}_2 = \left\{\left[\begin{array}{c|c}
 2\boldsymbol{a}_{\rm t, t}^{(i)} &  \boldsymbol{0} \\ \hline
  \boldsymbol{0} &  2\bar{\boldsymbol{a}}_{\rm t, t}^{(i)}
\end{array}\right], \left[\begin{array}{c|c}
 \boldsymbol{w}_{\rm t, t}^{(i)}&  \boldsymbol{0} \\ \hline
  \boldsymbol{0} &  \bar{\boldsymbol{w}}_{\rm t, t}^{(i)}
\end{array}\right] \right\},\qquad i\in \{1, \ldots, L-1\}, \\
& \tilde{\boldsymbol{\theta}}^{(0)}_2 = \left[\begin{array}{cc}
     \boldsymbol{\theta}_{\rm t}^{(0)}  \\ \hline
     \bar{\boldsymbol{\theta}}_{\rm t}^{(0)}
\end{array}\right]. 
\end{align*}

\begin{align*}
& \tilde{\boldsymbol{\theta}}^{(L)}_3 =\left\{ \left[\begin{array}{c|c}
 \boldsymbol{0} & 2\bar{\boldsymbol{a}}^{(L)}_{\rm t}
\end{array}\right],\left[\begin{array}{c|c}
 \b0 & \bar{\boldsymbol{w}}^{(L)}_{\rm t}
\end{array}\right]\right\},\\
&\tilde{\boldsymbol{\theta}}^{(i)}_3 = \left\{\left[\begin{array}{c|c}
 2\boldsymbol{a}_{\rm t, t}^{(i)} &  \boldsymbol{0} \\ \hline
  \boldsymbol{0} &  2\bar{\boldsymbol{a}}_{\rm t, t}^{(i)}
\end{array}\right], \left[\begin{array}{c|c}
 \boldsymbol{w}_{\rm t, t}^{(i)}&  \boldsymbol{0} \\ \hline
  \boldsymbol{0} &  \bar{\boldsymbol{w}}_{\rm t, t}^{(i)}
\end{array}\right] \right\},\qquad i\in \{1, \ldots, L-1\}, \\
& \tilde{\boldsymbol{\theta}}^{(0)}_3 = \left[\begin{array}{cc}
     \boldsymbol{\theta}_{\rm t}^{(0)}  \\ \hline
     \bar{\boldsymbol{\theta}}_{\rm t}^{(0)}
\end{array}\right]. 
\end{align*}

\begin{align*}
& \tilde{\boldsymbol{\theta}}^{(L)}_4 =\left\{ \left[\begin{array}{c|c}
 \boldsymbol{0} & 2\bar{\boldsymbol{a}}^{(L)}_{\rm t}
\end{array}\right],\left[\begin{array}{c|c}
 \b0 & \bar{\boldsymbol{w}}^{(L)}_{\rm t}
\end{array}\right]\right\},\\
&\tilde{\boldsymbol{\theta}}^{(i)}_4 = \left\{\left[\begin{array}{c|c}
 2\bar{\boldsymbol{a}}_{\rm t, t}^{(i)} &  \boldsymbol{0} \\ \hline
  \boldsymbol{0} &  2\bar{\boldsymbol{a}}_{\rm t, t}^{(i)}
\end{array}\right], \left[\begin{array}{c|c}
 \bar{\boldsymbol{w}}_{\rm t, t}^{(i)}&  \boldsymbol{0} \\ \hline
  \boldsymbol{0} &  \bar{\boldsymbol{w}}_{\rm t, t}^{(i)}
\end{array}\right] \right\},\qquad i\in \{1, \ldots, L-1\}, \\
& \tilde{\boldsymbol{\theta}}^{(0)}_4 = \left[\begin{array}{cc}
     \bar{\boldsymbol{\theta}}_{\rm t}^{(0)}  \\ \hline
     \bar{\boldsymbol{\theta}}_{\rm t}^{(0)}
\end{array}\right]. 
\end{align*}

\begin{align*}
& \tilde{\boldsymbol{\theta}}^{(L)}_5 =\left\{ \left[\begin{array}{c|c}
2\bar{\boldsymbol{a}}^{(L)}_{\rm t} & \boldsymbol{0}
\end{array}\right],\left[\begin{array}{c|c}
 \bar{\boldsymbol{w}}^{(L)}_{\rm t} & \b0 
\end{array}\right]\right\},\\
&\tilde{\boldsymbol{\theta}}^{(i)}_5 = \left\{\left[\begin{array}{c|c}
 2\bar{\boldsymbol{a}}_{\rm t, t}^{(i)} &  \boldsymbol{0} \\ \hline
  \boldsymbol{0} &  2\bar{\boldsymbol{a}}_{\rm t, t}^{(i)}
\end{array}\right], \left[\begin{array}{c|c}
 \bar{\boldsymbol{w}}_{\rm t, t}^{(i)}&  \boldsymbol{0} \\ \hline
  \boldsymbol{0} &  \bar{\boldsymbol{w}}_{\rm t, t}^{(i)}
\end{array}\right] \right\},\qquad i\in \{1, \ldots, L-1\}, \\
& \tilde{\boldsymbol{\theta}}^{(0)}_5 = \left[\begin{array}{cc}
     \bar{\boldsymbol{\theta}}_{\rm t}^{(0)}  \\ \hline
     \bar{\boldsymbol{\theta}}_{\rm t}^{(0)}
\end{array}\right]. 
\end{align*}

\begin{align*}
& \tilde{\boldsymbol{\theta}}^{(L)}_6 =\left\{ \left[\begin{array}{c|c}
2\bar{\boldsymbol{a}}^{(L)}_{\rm t} & \boldsymbol{0}
\end{array}\right],\left[\begin{array}{c|c}
 \bar{\boldsymbol{w}}^{(L)}_{\rm t} & \b0 
\end{array}\right]\right\},\\
&\tilde{\boldsymbol{\theta}}^{(i)}_6 = \left\{\left[\begin{array}{c|c}
 2\bar{\boldsymbol{a}}_{\rm t, t}^{(i)} &  \boldsymbol{0} \\ \hline
  \boldsymbol{0} &  \b0
\end{array}\right], \left[\begin{array}{c|c}
 \bar{\boldsymbol{w}}_{\rm t, t}^{(i)}&  \boldsymbol{0} \\ \hline
  \boldsymbol{0} &  \b0
\end{array}\right] \right\},\qquad i\in \{1, \ldots, L-1\}, \\
& \tilde{\boldsymbol{\theta}}^{(0)}_6 = \left[\begin{array}{cc}
     \bar{\boldsymbol{\theta}}_{\rm t}^{(0)}  \\ \hline
     \bar{\boldsymbol{\theta}}_{\rm b}^{(0)}
\end{array}\right]. 
\end{align*}
The arguments to connect $\tilde{\btheta}_j$ with $\tilde{\btheta}_{j+1}$ are analogous to those previously used to connect $\btheta_j$ with $\btheta_{j+1}$. We briefly outline them below for completeness.

\emph{From $\tilde{\btheta}_1$ to $\tilde{\btheta}_2$.} First, we interpolate from $\btheta_{\rm b}^{(0)}$ to $\bar{\btheta}_{\rm t}^{(0)}$ with no loss change. Then, for $i=1, \ldots, L-1$, we successively interpolate from $\b0$ to $\bar{\bw}_{\rm t, t}^{(i)}$ and from $\b0$ to $2\bar{\ba}_{\rm t, t}^{(i)}$ with no loss change. 

\emph{From $\tilde{\btheta}_5$ to $\tilde{\btheta}_6$.} For $i=L-1, \ldots, 1$, we successively interpolate from $2\bar{\ba}_{\rm t, t}^{(i)}$ to $\b0$ and from $\bar{\bw}_{\rm t, t}^{(i)}$ to $\b0$ with no loss change. Finally, we interpolate from $\bar{\btheta}_{\rm t}^{(0)}$ to $\bar{\btheta}_{\rm b}^{(0)}$ with no loss change.

\emph{From $\tilde{\btheta}_4$ to $\tilde{\btheta}_5$.} The output of layer $L-1$ is obtained by stacking the output of two identical sub-networks. Thus, we can interpolate between $\tilde{\btheta}_4$ and $\tilde{\btheta}_5$ with no change in loss. 

\emph{From $\tilde{\btheta}_3$ to $\tilde{\btheta}_4$.} For $i=L-1, \ldots, 1$, we interpolate from $2\ba_{\rm t, t}^{(i)}$ to $\b0$ and from $\bw_{\rm t, t}^{(i)}$ to $\b0$ with no change in loss. Then, we interpolate from $\btheta_{\rm t}^{(0)}$ to $\bar{\btheta}_{\rm t}^{(0)}$ with no change in loss. Finally, for $i=1, \ldots, L-1$, we restore sequentially $2\bar{\ba}_{\rm t, t}^{(i)}$ and $\bar{\bw}_{\rm t, t}^{(i)}$ in the top left corner of the corresponding parameter matrices with no change in loss. 

\emph{From $\tilde{\btheta}_2$ to $\tilde{\btheta}_3$.} From the previous arguments, we have that $L_N(\tilde{\btheta}_2)= L_N(\tilde{\btheta}_1)\le L_N(\btheta)+\varepsilon$ and $L_N(\tilde{\btheta}_3)= L_N(\tilde{\btheta}_6)\le L_N(\bar{\btheta})+\varepsilon$. First, we interpolate from $\{[\,\,2\boldsymbol{a}^{(L)}_{\rm t} \mid \boldsymbol{0}\,\,], [\,\,\boldsymbol{w}^{(L)}_{\rm t} \mid \boldsymbol{0}\,\,]\}$ to $\{[\,\,2\boldsymbol{a}^{(L)}_{\rm t} \mid \boldsymbol{0}\,\,], [\,\,\boldsymbol{w}^{(L)}_{\rm t} \mid \bar{\boldsymbol{w}}^{(L)}_{\rm t}\,\,]\}$ with no change in loss. Similarly, we interpolate from $\{[\,\,\b0 \mid 2\bar{\boldsymbol{a}}^{(L)}_{\rm t} \,\,], [\,\,\b0 \mid \bar{\boldsymbol{w}}^{(L)}_{\rm t} \,\,]\}$ to $\{[\,\,\b0 \mid 2\bar{\boldsymbol{a}}^{(L)}_{\rm t} \,\,], [\,\,\boldsymbol{w}^{(L)}_{\rm t} \mid \bar{\boldsymbol{w}}^{(L)}_{\rm t}\,\,]\}$ with no change in loss. Furthermore, as the loss is convex in $\boldsymbol{a}^{(L)}$, we interpolate from $\{[\,\,2\boldsymbol{a}^{(L)}_{\rm t} \mid \boldsymbol{0}\,\,], [\,\,\boldsymbol{w}^{(L)}_{\rm t} \mid \bar{\boldsymbol{w}}^{(L)}_{\rm t}\,\,]\}$ to $\{[\,\,\b0\mid 2\bar{\boldsymbol{a}}^{(L)}_{\rm t}\,\,], [\,\,\boldsymbol{w}^{(L)}_{\rm t} \mid \bar{\boldsymbol{w}}^{(L)}_{\rm t}\,\,]\}$ while keeping the loss upper bounded by $\max(L_N(\btheta), L_N(\bar{\btheta}))+\varepsilon$.

\end{proof}

\section{Additional Numerical Results}\label{app:addexp}

\begin{figure}[p]
\begin{minipage}{0.5\linewidth}
\includegraphics[width=\linewidth]{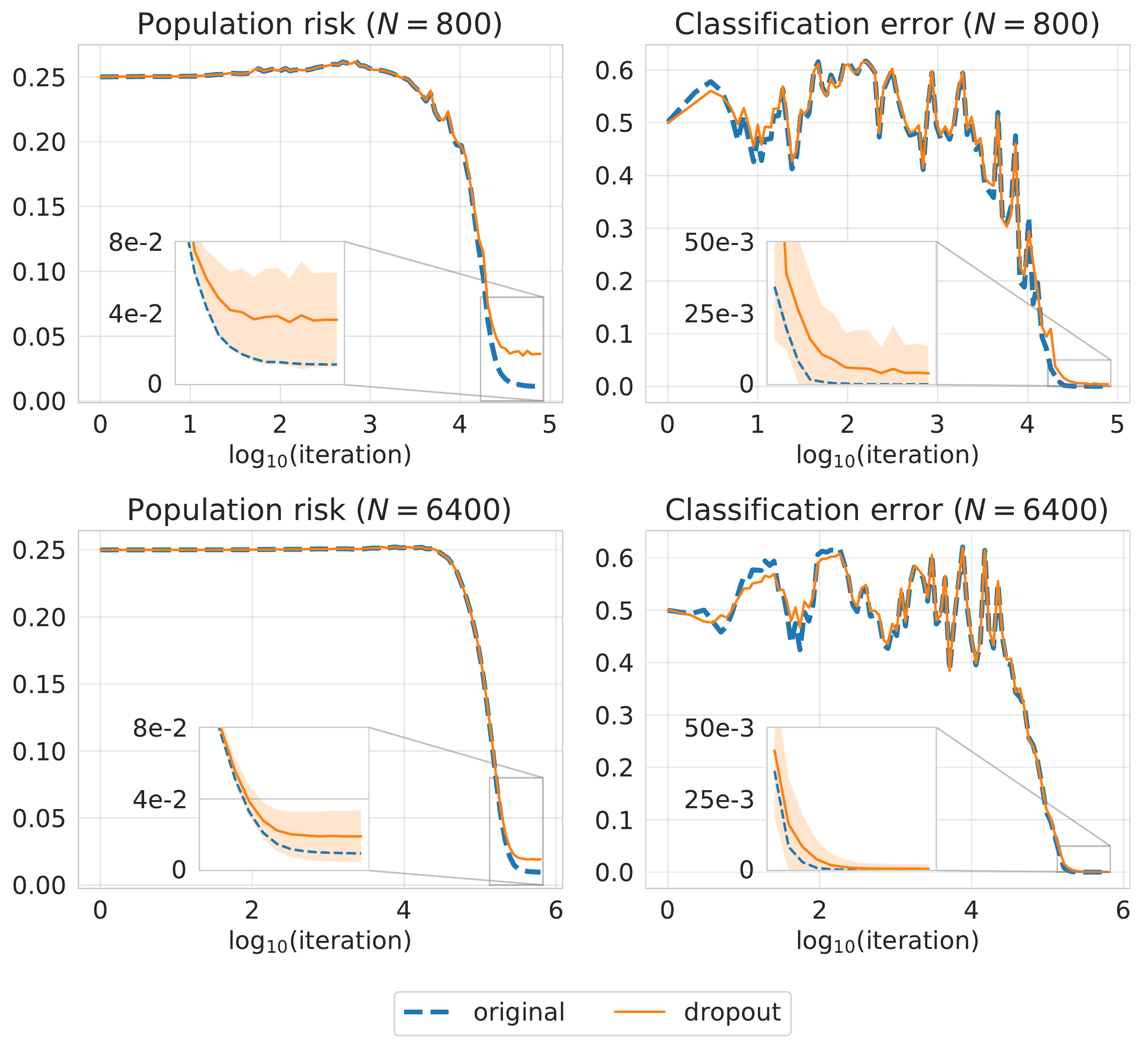}
\captionof{figure}{Comparison of population risk and classification error between the trained network (blue dashed curve) and the dropout network (orange curve) for the classification of isotropic Gaussians.}\label{fig:iso1}
\end{minipage}
\hspace{8mm}\begin{minipage}{0.5\linewidth}
\includegraphics[width=\linewidth]{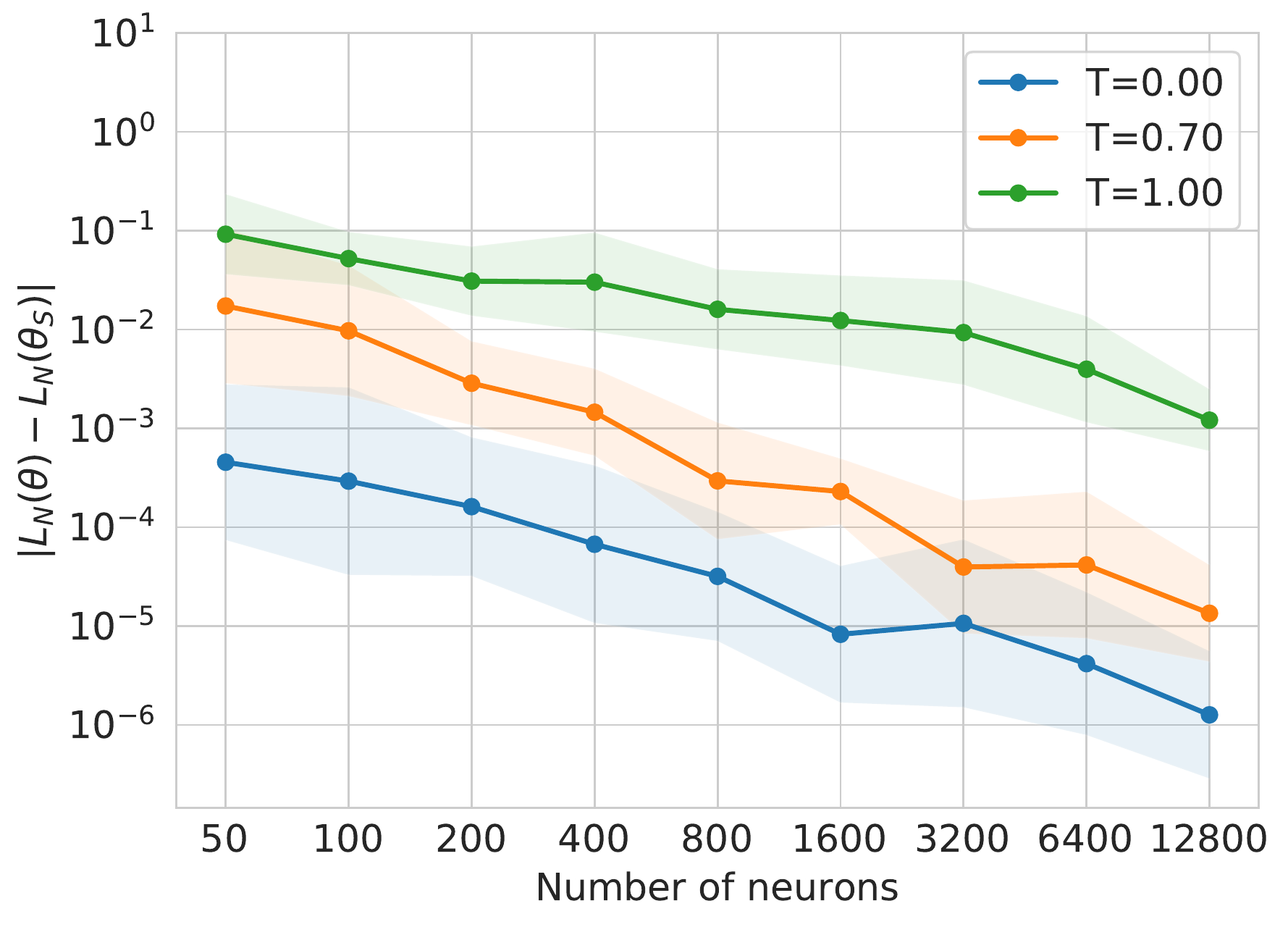}
\captionof{figure}{Change in loss between the full network and the dropout network for the classification of isotropic Gaussians, as a function of the number of neurons $N$ of the full network.}\label{fig:iso2}
\end{minipage}
\end{figure}

\begin{figure*}[p]
    \centering
    \includegraphics[width=0.5\linewidth]{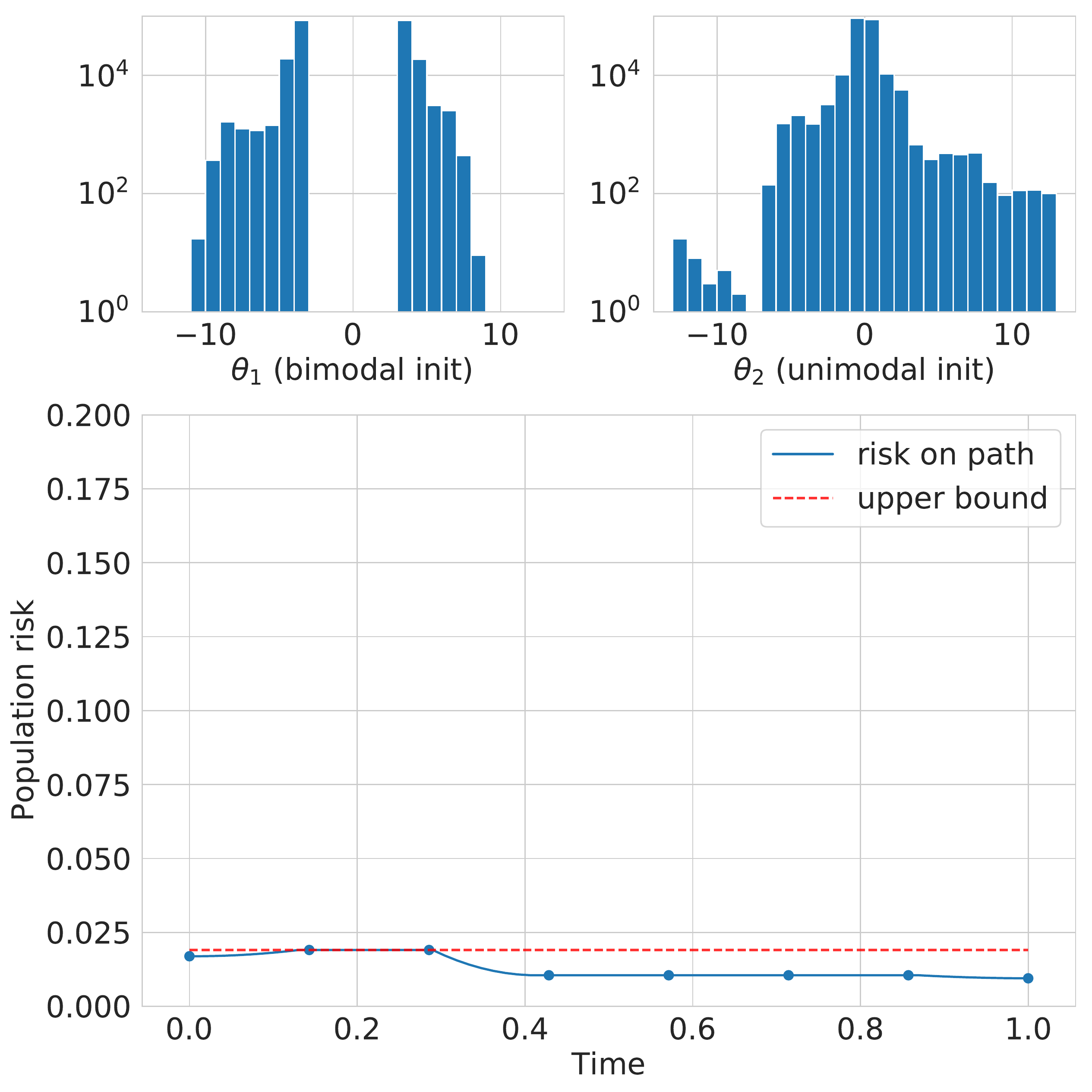}
    \caption{Classification error along a piecewise linear path that connects two SGD solutions $\btheta_1$ and $\btheta_2$ for the classification of isotropic Gaussians with $N=3200$. The two SGD solutions are initialized with different distributions, and we show their histograms to highlight that $\btheta_1$ cannot be obtained by permuting $\btheta_2$.}\label{fig:iso3}

\end{figure*}

In Figures \ref{fig:iso1}, \ref{fig:iso2} and \ref{fig:iso3}, we consider the problem of classifying isotropic Gaussians. This is an artificial dataset considered in \cite{mei2018mean}. The label $y$ is chosen uniformly at random between $-1$ and $1$, i.e., $y\sim\operatorname{Unif}(\{-1, 1\})$. Given $y$, the feature vector $\bx$ is a $d$-dimensional isotropic Gaussian with covariance matrix $(1+y\Delta)^2\bI_d$, i.e., $\bx\sim \mathcal N(\b0, (1+y\Delta)^2\bI_d)$. We set $d = 32$ and $\Delta = 0.5$, and we run the one-pass (or online) SGD algorithm \eqref{eq:SGD} on the two-layer neural network \eqref{eq:2layer} with sigmoid activation function ($\sigma(x)=1/(1+e^{-x})$). We estimate the population risk and the classification error on $10^4$ independent samples. Figure \ref{fig:iso1} compares the performance of the trained network (blue dashed curve) and of the dropout network (orange curve) obtained by removing half of the neurons. We plot the population risk and the classification error for $N=800$ and $N=6400$. As expected, the performance of the dropout network improves with $N$, and it is very close to that of the trained network already for $N=800$. In fact, for $N=800$ the classification error of the dropout network is $<0.4\%$. Figure~\ref{fig:iso2} plots the change in loss between the full and the dropout network, as a function of the number of neurons of the full network $N$. The change in loss decreases steadily with $N$ for all the values of $T$ taken into account. Finally, Figure \ref{fig:iso3} shows that the optimization landscape is approximately connected when $N=3200$.

\begin{figure}[p]
    \centering
    \subfloat[MNIST, three-layer]{\includegraphics[width=.5\columnwidth]{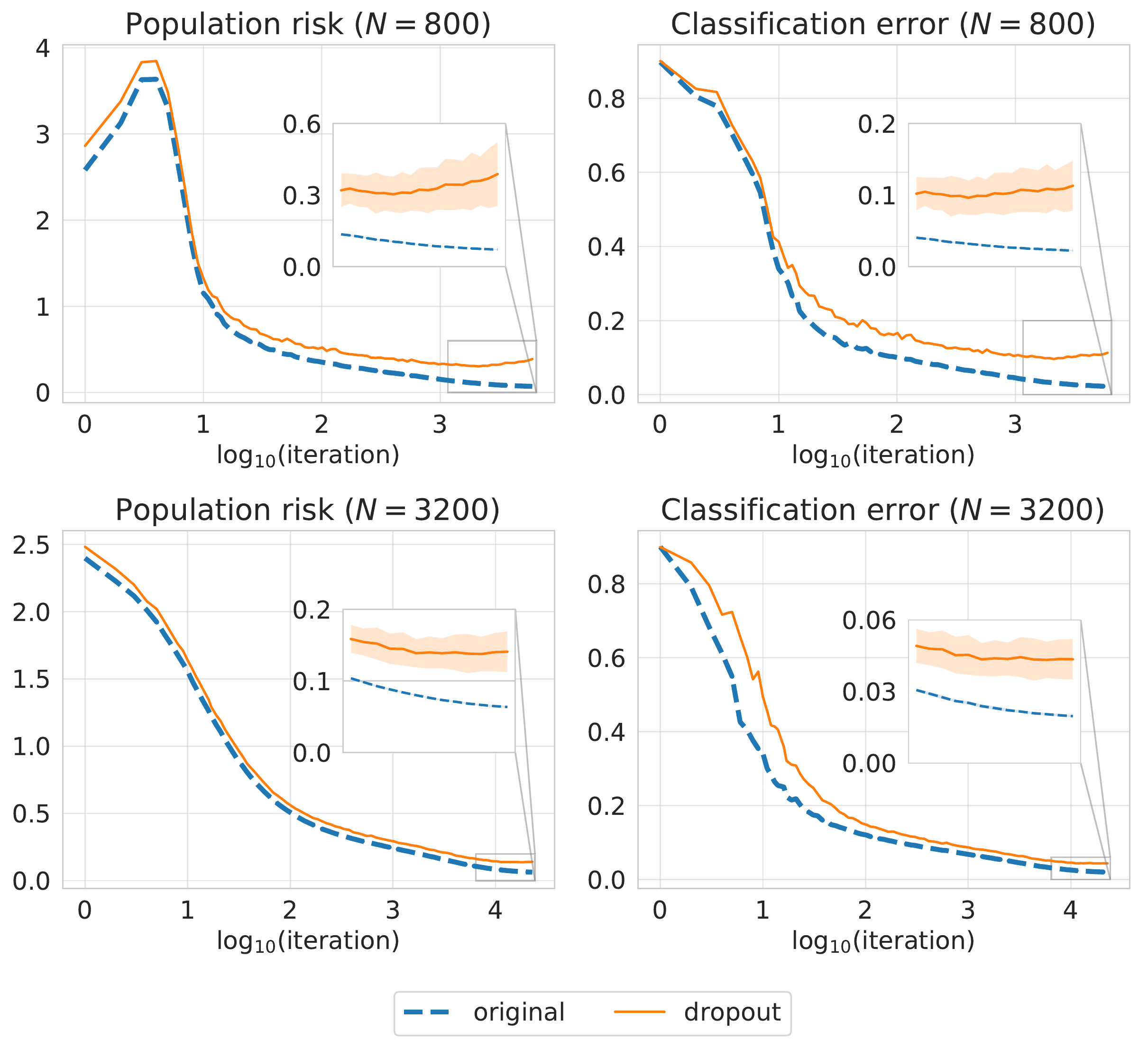}}
    \subfloat[CIFAR-10, four-layer]{\includegraphics[width=.5\columnwidth]{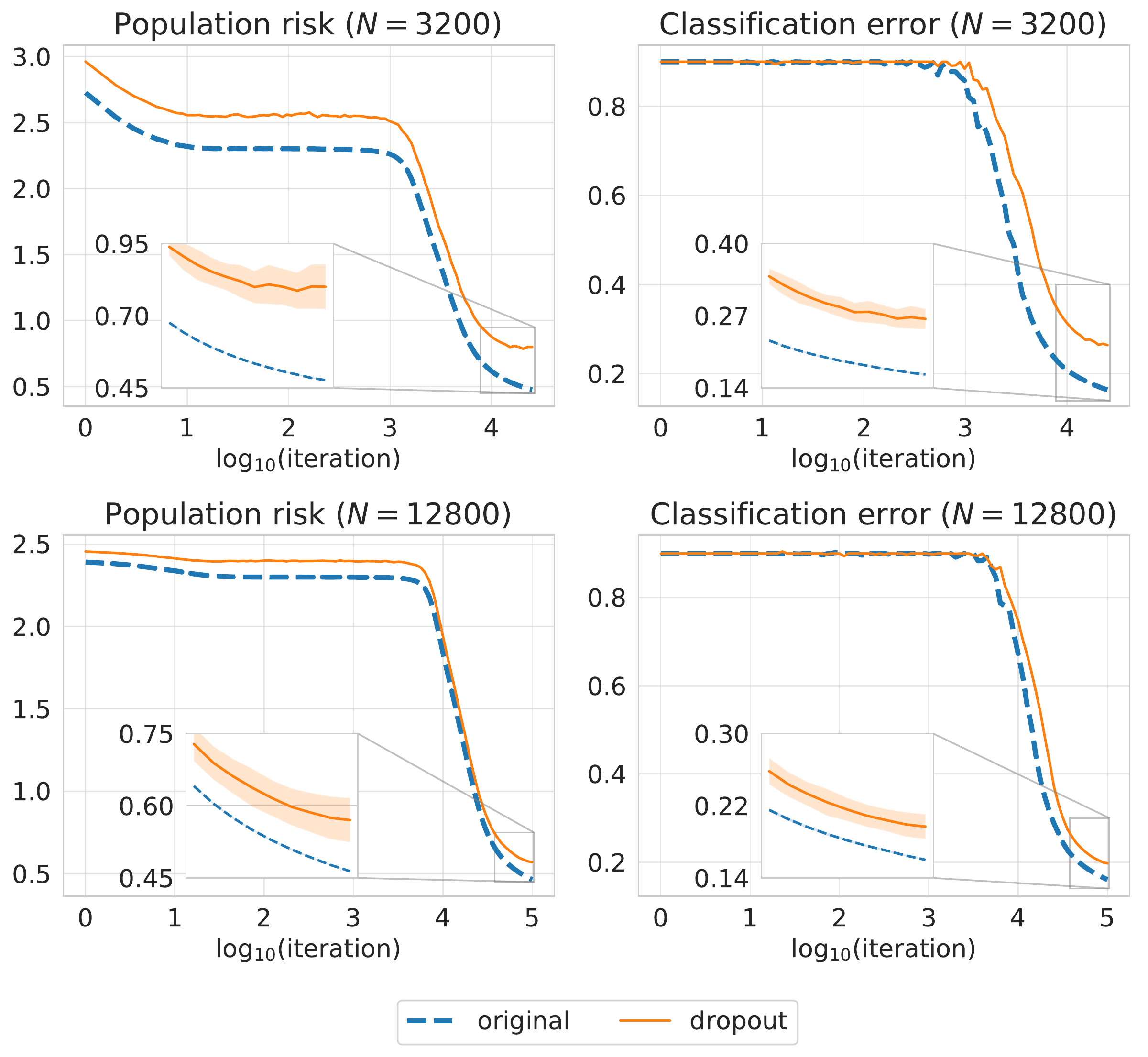}}
    \caption{Comparison of population risk and classification error between the trained network (blue dashed curve) and the dropout network (orange curve).}\label{fig:multi1}
\end{figure}

\begin{figure}[p]
    \centering
    \subfloat[MNIST, three-layer]{\includegraphics[width=.5\columnwidth]{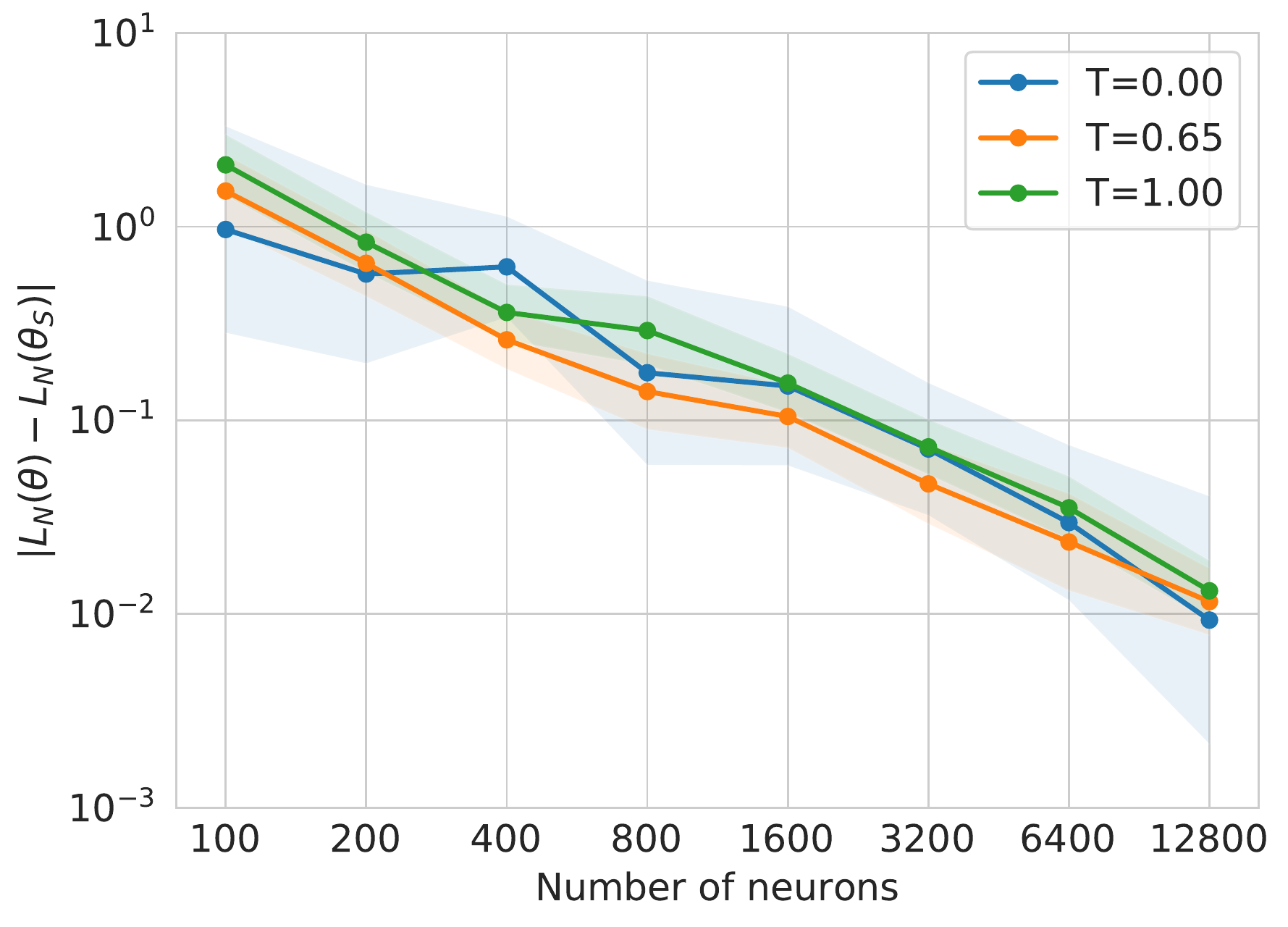}}
    \subfloat[CIFAR-10, four-layer]{\includegraphics[width=.5\columnwidth]{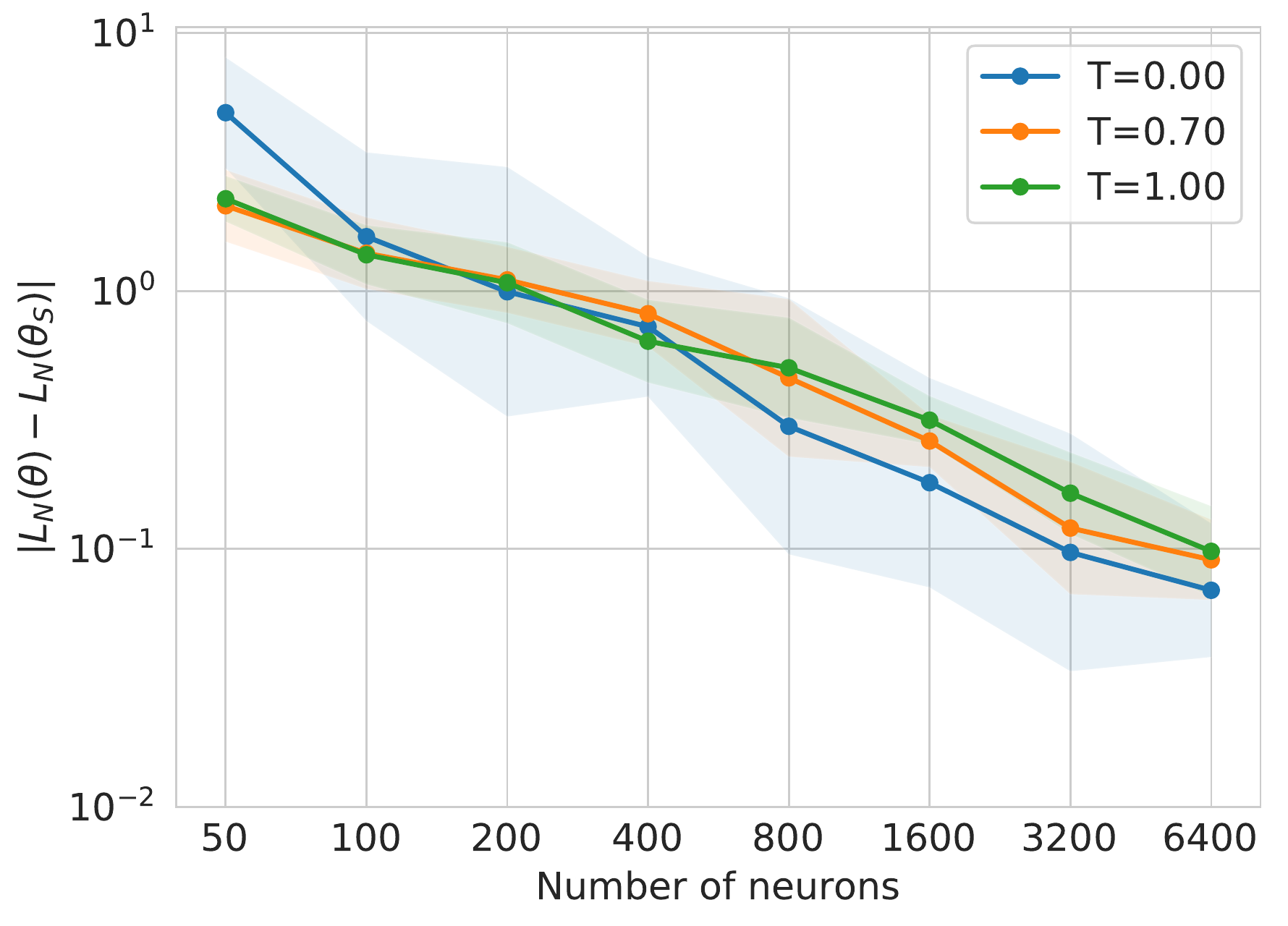}}
    \caption{Change in loss after removing half of the neurons from each layer, as a function of the number of neurons $N$ of the full network.}\label{fig:multi2}
\end{figure}

\begin{figure}[t!]
    \centering
    \includegraphics[width=0.55\linewidth]{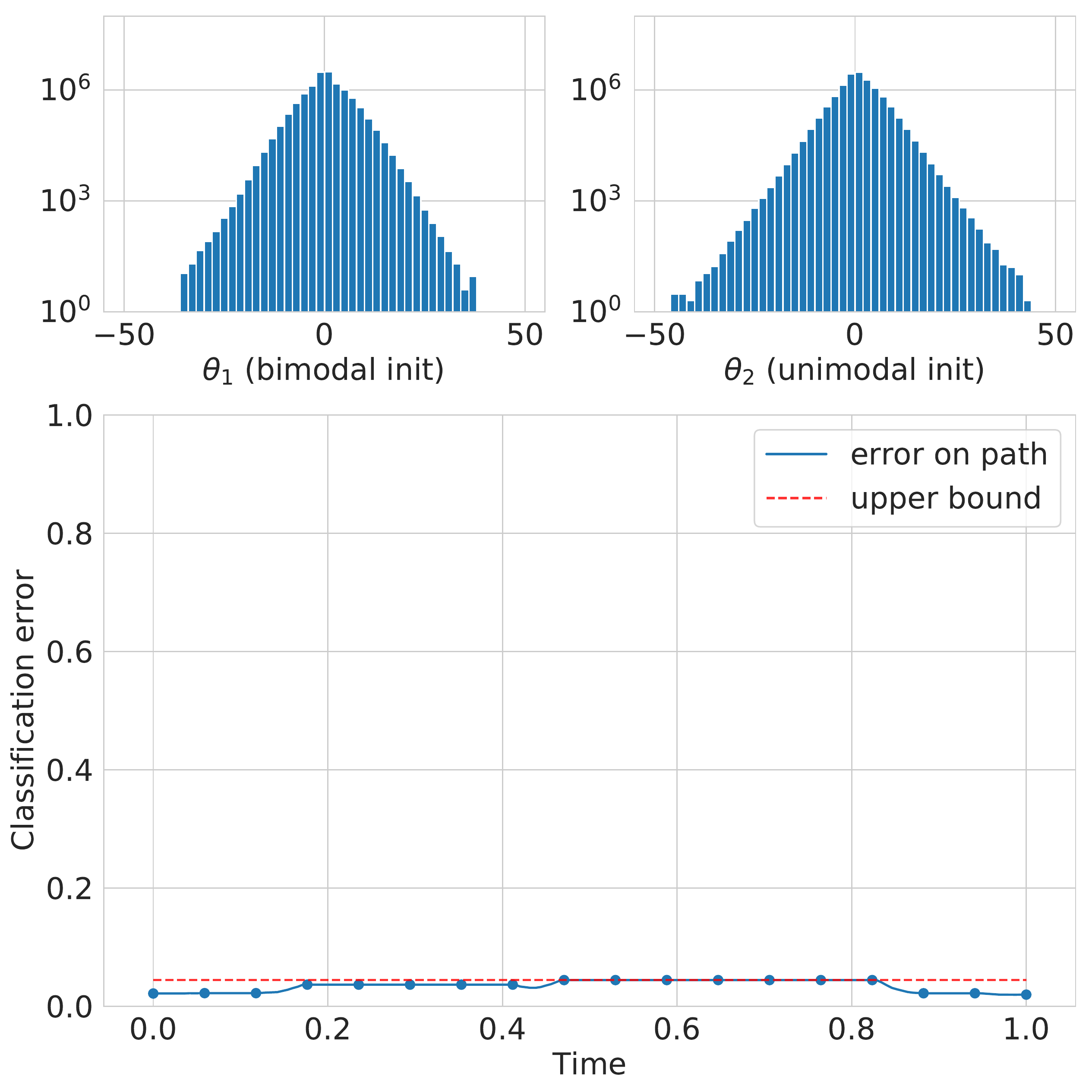}
    \caption{Classification error along a piecewise linear path that connects two SGD solutions $\btheta_1$ and $\btheta_2$ for MNIST classification with a three-layer neural network with $N=3200$.}    \label{fig:multi3}
\end{figure}

In Figures \ref{fig:multi1}, \ref{fig:multi2} and \ref{fig:multi3}, we consider MNIST classification with a three-layer neural network and CIFAR-10 classification with a four-layer neural network. The results are qualitatively similar to those of Figures \ref{fig:1}, \ref{fig:2} and \ref{fig:3} in Section \ref{sec:num}.

\end{document}